\documentclass[preprint, review, 12pt]{elsarticle}
\usepackage{graphicx}
\usepackage{fullpage}
\usepackage{mathrsfs}
\usepackage{setspace}
\usepackage{bm}
\usepackage{amssymb}
\usepackage{graphicx,subcaption}
\usepackage[normalem]{ulem}
\usepackage[dvipsnames]{xcolor}
\usepackage{comment}
\definecolor{mypink}{rgb}{0.858, 0.188, 0.478}
\usepackage{lineno}
\usepackage{verbatim}
\usepackage{xcolor}
\usepackage{eqnarray}
\usepackage{siunitx}
\usepackage[T1]{fontenc}
\usepackage{mathtools}
\RequirePackage{ifthen}
\RequirePackage{graphicx}
\usepackage{algpseudocode}
\RequirePackage{amsmath}
\RequirePackage{fancyhdr}
\usepackage{algorithm,algpseudocode}
\usepackage{booktabs}
\usepackage{siunitx}
\usepackage{fancyhdr,graphicx,amsmath,amssymb,microtype}
\usepackage{alltt}%
\include{pythonlisting}
\usepackage{setspace}
\usepackage{booktabs}
\usepackage{multirow}
\usepackage{glossaries}
\newacronym{bc}{BC}{boundary condition}
\usepackage[utf8]{inputenc}
\usepackage[english]{babel}
\definecolor{purp}{rgb}{0.4,0.2,0.8}

\DeclareMathOperator*{\argmax}{arg\,max}
\DeclareMathOperator*{\argmin}{arg\,min}

\usepackage{caption}
\usepackage{subcaption}
\usepackage{hyperref}
\definecolor{custom-blue}{RGB}{3,69,173}
\hypersetup{
    colorlinks = true,
    urlcolor   = blue,
    citecolor  = custom-blue,
}

\journal{Journal of Computational Physics}

\begin{document}

\begin{frontmatter}


% \title{Non-intrusive polynomial chaos on diffusion manifolds for fast uncertainty quantification in complex stochastic systems}

%\title{Local polynomial chaos expansions on principal geodesic submanifolds for uncertainty quantification in high-dimensional models}

%%% Alternate Titles
\title{Polynomial Chaos Expansions on Principal Geodesic Grassmannian Submanifolds for Surrogate Modeling and Uncertainty Quantification}

%\title{Learning on Principal Geodesic Submanifolds for High-Dimensional Surrogate Modeling and Uncertainty Quantification}

% Polynomial Chaos Expansions on Subspace Manifolds

% Fast UQ in high dimensions: A novel encoder-decoder framework based on polynomial chaos expansions

% A novel polynomial chaos-based encoder-decoder framework for uncertainty quantification in high-dimensional models

% Surrogate modeling in high dimensions with an encoder-decoder framework based on polynomial chaos expansions and manifold learning

% Surrogates for high-dimensional models with polynomial chaos expansions on latent spaces


\author[1,3]{Dimitris G. Giovanis\corref{cor1}}\ead{dgiovan1@jhu.edu@jhu.edu}
\author[4]{Dimitrios Loukrezis}\ead{dimitrios.loukrezis@tu-darmstadt.de}
\author[2, 3]{Ioannis G Kevrekidis}\ead{yannisk@jhu.edu}
\author[1,3]{Michael D. Shields}\ead{michael.shields@jhu.edu}
\address[1]{Department of Civil \& Systems Engineering, Johns Hopkins University, USA}
\address[2]{Department of Chemical and Biomolecular Engineering, Johns Hopkins University, USA}
\address[3]{Hopkins Extreme Materials Institute, Johns Hopkins University, USA}
\address[4]{Institute for Accelerator Science and
Electromagnetic Fields, Technische Universität Darmstadt, Germany}

\cortext[cor1]{Corresponding author.}

\begin{abstract}
In this work we introduce a manifold learning-based surrogate modeling framework for uncertainty quantification in high-dimensional stochastic systems. Our first goal is to perform data mining on the available simulation data to identify a set of low-dimensional (latent) descriptors that efficiently parameterize the response of the high-dimensional computational model. To this end, we employ Principal Geodesic Analysis  on the Grassmann manifold of the response to identify a set of disjoint principal geodesic submanifolds, of possibly different dimension, that captures the variation in the data.  Since operations on the Grassmann require the data to be concentrated,  we propose an adaptive algorithm based on Riemanniann K-means and the minimization of the sample Fr\'{e}chet variance on the Grassmann manifold to identify ``local'' principal geodesic submanifolds that represent different system behavior across the parameter space. Polynomial chaos expansion is then used to construct a mapping between the random input parameters and the projection of the response on these local principal geodesic submanifolds. The method is demonstrated on four  test cases, a toy-example that involves points on a hypersphere, a Lotka-Volterra dynamical system, a continuous-flow stirred-tank chemical reactor system, and a two-dimensional Rayleigh-B\'enard convection problem. %In all test cases, the proposed method produces highly accurate surrogate models with a very small number of available data. %, which can ultimately lead to the significant acceleration of uncertainty quantification tasks.

\end{abstract}


\end{frontmatter}

%%\begin{comment}

%\tableofcontents


%\linenumbers

\section{Introduction}
\label{S:Background}

Computer simulations play a vital role in the field of scientific computing and its engineering applications. The principle of the simulation is to numerically solve a set of mathematical equations (e.g., nonlinear ordinary or partial differential equations) that describe the behavior of the real physical and/or engineering system (e.g., materials, climate systems, and civil infrastructure), often on a complex spatio-temporal domain. 
In real-world applications, such systems are complex and multiscale (and/or multiphysics), requiring very high-fidelity models to accurately capture their response. Despite the aggregation of computational power into supercomputers and the development of sophisticated algorithms to solve massive-scale problems, running a single simulation of a very high-dimensional model can be cumbersome, even with exascale computing.  On top of that, since the model is an approximation of the true physics (typically based on the modeler’s knowledge and experience) with parameters calibrated to existing data and may have components (such as initial conditions or excitation) that are inherently stochastic, these systems can be highly uncertain. 
% For example, the model parameters are not known exactly. 
To increase our confidence in the predictions of the model, we need to propagate these uncertainties through the system and understand their influence on the response quantities of interest (QoIs). To this end, uncertainty quantification (UQ)  provides a framework for the mathematical treatment of uncertainty in computational models \cite{SullivanUQ}.  However, a major limitation in performing UQ with high-fidelity models is the numerous repeated simulations of the computational model required for Monte Carlo-based methods \cite{Metropolis, Liu}. Therefore, there is a continuous effort to overcome these computational obstacles.

To reduce computational effort, it is common to employ a surrogate model (aka response surface, metamodel, or emulator) \cite{Sudret2017, Bhosekar}, which is a fast-running alternative mathematical function that maps model inputs to outputs (e.g., via supervised learning), to propagate uncertainty at a significantly reduced computational cost. 
% Moreover, surrogate models can be used for optimization, parameter estimation, and control in physical systems.
To construct a surrogate that can approximate the response of a system at new points in the input parameter space, a finite number of high-fidelity model evaluations are required to provide input-output training data. The construction of accurate surrogates typically requires smooth input-output functional relations that predict low-dimensional QoIs, especially when the available training data are scarce, e.g., due to a limited simulation budget. 
However, these assumptions may not be realistic in real-world applications, as many system exhibit strongly nonlinear behavior and produce high-dimensional responses for which scalar (or low-dimensional) descriptors are inadequate.
% it often oversimplifies the complexity of the solution. 

Over the last 20+ years, various surrogate modeling techniques have been proposed to accelerate UQ tasks in computational models, such as polynomial chaos expansions (PCEs) \cite{wu2017inverse, moustapha2019surrogate, sun2020global}, support vector machines (SVMs) \cite{trinchero2018machine, he2019data},  Gaussian processes (GPs) \cite{bilionis2012multi, tripathy2016gaussian}, and artificial neural networks \cite{tripathy2018deep, mo2019deep}. Among them, the non-intrusive, regression-based PCE is an attractive choice when only limited data are available \cite{loukrezis2020robust, blatman2011adaptive, hadigol2018least}. The PCE learns the input-output relationship in terms of a polynomial expansion, the basis polynomials of which are orthonormal with respect to the probability density function that characterizes the input variables. One of the main advantages of PCE methods is that UQ tasks such as moment estimation \cite{knio2006uncertainty} and sensitivity analysis \cite{sudret2008global,crestaux2009polynomial} can be performed by simply post-processing the PCE coefficients. 
Furthermore, recent advances have enabled physical constraints to be placed on PCE models \cite{Sharma_et_al_ICASP_2023, novak2023physics}, thus making PCE surrogates even more attractive for UQ on physics-based models.  %On the other hand, Gaussian process regression incorporates prior knowledge about the input-output relationship through the mean and covariance function of a Gaussian process. 

% However, in the construction of the surrogate, there are certain factors that may reduce its accuracy and even its applicability. For example, 
A major challenge in UQ is the \textit{curse of dimensionality}. 
For example, fast growing polynomial bases due to high-dimensional inputs pose a major obstacle to constructing PCE surrogates. To overcome this limitation, numerous sparse and/or basis-adaptive PCE methods have been proposed in the literature \cite{luethen2021sparse, luthen2022automatic}.  
The curse of dimensionality also arises for models with high-dimensional outputs, for example, in the form of time series or spatial fields quantities. 
This is the case of interest in this work. 
%In order to understand and eventually predict the functional relationship between the input parameter space and the corresponding high-dimensional response, the latter must not be treated as a scalar where information (that may be important) is discarded and the behavior of the high-dimensional model is represented by a single quantity (e.g., the maximum value of the stress field). 

One way to mitigate the curse of dimensionality in high-dimensional models is to use dimension reduction. Dimension reduction is an unsupervised learning task that refers to the process of identifying and extracting \textit{important} latent features (attributes) while preserving most of the information contained in the original dataset \cite{MaatenDR}.  To identify the essential features, a variety of linear and nonlinear dimension reduction techniques can be used to map data onto lower-dimensional manifolds (aka embeddings or latent spaces).  Linear dimension reduction methods such as the widely-used principal component analysis (PCA) (and the closely related singular value decomposition, SVD, and proper orthogonal decomposition, POD) apply a linear operation to the high-dimensional data to map it onto a new rotated and/or stretched basis in Euclidean space that has a lower dimension. Nonlinear dimension reduction methods, also known as manifold learning methods, assume that high-dimensional data reside on some low-dimensional nonlinear manifold. These include methods such as diffusion maps (DMaps) \cite{coifman2006diffusion, dos2022grassmannian}, Laplacian eigenmaps \cite{belkin2003laplacian}, and Isomap \cite{balasubramanian2002isomap}. For UQ purposes, surrogate model training for high-dimensional physics-based models can become tractable when dimension reduction methods are applied to high-dimensional model outputs, e.g., due to smooth functional relations between inputs and reduced outputs.
%Moreover, recent advances in computer resources such as graphics and tensor processors (GPUs, TPUs) have enabled supervised machine learning algorithms such as deep neural networks (DNN) (e.g., multilayer perceptrons and autoencoders) as an alternative to perform dimensionality reduction. 
As a result, dimension reduction methods are now being fully integrated into surrogate-based UQ efforts for high-dimensional models. For example, recent works of the authors have considered the construction of surrogate models using nonlinear manifold learning on the Grassmannian \cite{giovanis2020data, giovanis2018uncertainty, giovanis2019variance}. Another class of methods leverages DMaps to either draw samples from a distribution on the diffusion manifold \cite{soize2016data, soize2017polynomial, soize2021probabilistic} or construct surrogate models on the diffusion manifold \cite{kalogeris2020diffusion, koronaki2020data, dos2022grassmannian_b}. Lataniotis et al.~\cite{lataniotis2020extending} coupled kernel PCA with Kriging and PCE to extend surrogates to high-dimensional models. Recently, a systematic survey of 14 different manifold learning methods was undertaken for high-dimensional surrogate modeling by Kontolati et al.~\cite{kontolati2022survey}, who devised the manifold PCE (m-PCE).

An alternative means of constructing surrogates for high dimensional models uses deep learning. Recent advances have allowed the identification of latent representations of data using multi-layer perceptrons, autoencoders \cite{wang2016auto}, and convolutional neural networks \cite{rawat2017deep} for building surrogate models with low-dimensional input and output parameters \cite{tripathy2018deep, hesthaven2018non, mo2019deep, mo2019deep2, thuerey2020deep, wang2021efficient}. Although deep learning-based surrogates are capable of capturing complex nonlinear relations between high-dimensional inputs and outputs, they are very costly to train, rely on the heuristic choice of the network architecture and the calibration of a huge number of hyperparameters. Very recently, the class of neural operators -- specifically the Deep Operator Network (DeepONet)~\cite{lu2021learning} and Fourier Neural Operator (FNO)~\cite{li2020fourier} and their variants -- have become very attractive options to build continuous machine learned surrogate solution operators. These operators have been systematically compared (under certain conditions \cite{kontolati2023influence}) with the general class of manifold learning-based finite dimensional regressors (since we do not learn solution operators) of the type we study here (specifically the manifold PCE, m-PCE \cite{kontolati2022survey}) to compare model generalization and robustness to noise \cite{kontolati2023influence}. Another particularly noteworthy candidate for surrogate model construction in UQ are the general class of physics informed neural networks (PINNs) \cite{raissi2019physics} and in particular their extension for physics informed neural operators \cite{goswami2022physics}. 
A detailed discussion of these methods is out of the scope of this work though.

In this work we introduce a novel framework based on Principal Geodesic Analysis (PGA) on the Grassmann manifold, which combines manifold learning principles with PCE surrogate model construction for the prediction of dimension-reduced solutions that can be used to reconstruct approximate full model solutions from on a limited number of model evaluations.  We are particularly interested in complex models that generate high-dimensional outputs (e.g., time series or spatial field quantities) that may be computationally expensive to run. Similar to our previous works \cite{giovanis2018uncertainty, giovanis2020data, kontolati2021manifold}, each data point corresponds to a vector of the full-field solution.  The projection onto the Grassmann manifold is performed by recasting the solution vector into a matrix and then via singular value decomposition (SVD) the representative on the Grassmann manifold is obtained. The reader is referred to  \cite{giovanis2018uncertainty} for a detailed discussion on the motivation that underpins such representation.

The remainder of the paper is organized as follows. Theoretical concepts of the Grassmann manifold and PGA are briefly introduced in Section 2. We also present the important ingredients of the proposed clustering framework of points on the Grassmann manifold which is based on the minimization of the Fr\'{e}chet variance. This step is necessary for developing local surrogate models. Section three discuss in detail the steps required for (i) performing local geodesic submanifold interpolation using polynomial chaos expansion within each submanifold and, (ii) obtaining the full solutions from the reduced-order predictions. The performance of the proposed approach is assessed by three illustrative applications given in Section 4. The first application involves a toy-example that involves data on the sphere. In the second application, the method is applied to predict time-evolution on the classic Lotka-Volterra (predator-prey) dynamical system. The third application deals with a continuous  stirred-tank reactor system. The fourth example involves predicting Rayleigh-B\'enard convection in a two-dimensional rectangular domain.  Finally, Section 5 presents the conclusions. 

The distinct components of the proposed method have been implemented\footnote{The code to reproduce the results for all numerical examples will be available in \url{https://github.com/dgiovanis} after the publication of the paper.} using UQpy  (Uncertainty Quantification with python \cite{olivier2020uqpy, tsapetis2023uqpy}), a general-purpose open-source software for modeling uncertainty in physical and mathematical systems, and Geomstats \cite{JMLR:v21:19-027, geomstats_paper}. 

\section{Principal Geodesic Analysis on the Grassmann Manifold}
\label{prem}

In this section, we briefly review the geometry of the Grassmann manifold and the basic concepts of Principal Geodesic analysis (PGA). 

\subsection{Grassmann manifold}

\begin{figure}[t!]
\begin{center}
\includegraphics[width=0.7\textwidth]{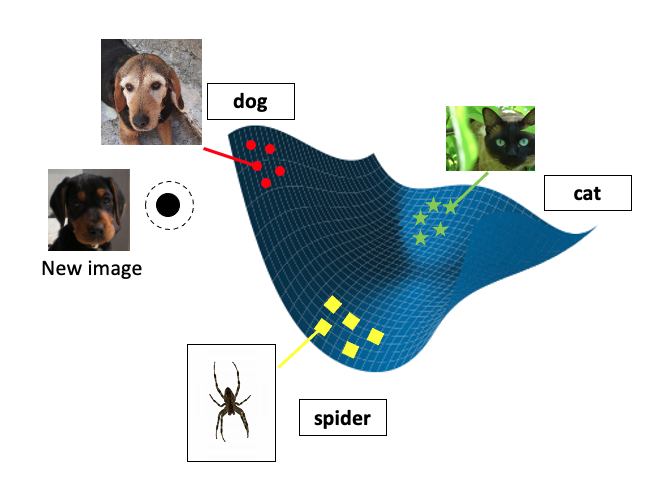}
\caption{ For  a set of training images that correspond to three different classes, namely cats, dogs, and spiders, we can classify these images based on their respective subspaces on the Grassmannian and when a new image (e.g., an image of a dog) is introduced we can assign it to the appropriate class.}
\label{fig:Grassmann}
\end{center}
\end{figure}

The Grassmann manifold $\mathcal{G}(p, n)$ is a (smooth, compact) matrix manifold defined as the set of all $p$-dimensional linear subspaces in $\mathbb{R}^n$. The Grassmann manifold can be viewed as a quotient manifold of the Stiefel manifold $\mathcal{S}t(p, n)$ \cite{edelman1998geometry}, which is the manifold of all $p$-dimensional orthonormal matrices in $\mathbb{R}^n$.  In this setting, a matrix $\textbf{U} \in \mathcal{S}t(p, n)$ can be considered representative of subspace $\mathcal{U} \in \mathcal{G}_{p, n}$ if its columns span the corresponding subspace, i.e., $\mathcal{U}\equiv \text{span}{(\textbf{U}})$.  Therefore, the Stiefel representation of the Grassmannian provides a natural and intuitive representation for subspace analysis \cite{absil2004riemannian, Edelman1998}. We specifically aim to use a subspace representation of a system due to the difficulty in assessing similarity between high-dimensional data, i.e., data generated by models with high-dimensional QoIs. The Grassmann manifold is an important mathematical modeling construct for several applications, ranging from problems in machine learning \cite{huang2018building} and computer vision and image processing (see e.g. Figure \ref{fig:Grassmann}) \cite{lui2012advances}, to low-rank matrix optimization problems \cite{boumal2015low}, dynamic low-rank decomposition \cite{wang2014low} and model reduction \cite{amsallem2008interpolation, dos2022grassmannian, upadhyay2022data, giovanis2020data, giovanis2018uncertainty}. 

On the Grassmann, the curve  with the minimum length between two points $\textbf{U}_1, \textbf{U}_2$  is called the geodesic path and it can be parameterized by a function:  $\gamma:t \in [0, 1] \rightarrow \mathcal{G}_{p, n}$, where $\gamma(0)=\textbf{U}_1$ and $\gamma(1)=\textbf{U}_2$ are the initial conditions. The geodesic is determined by solving a second-order differential equation on $\mathcal{G}_{p, n}$ that corresponds to motion along the curve with constant tangential velocity (i.e., all acceleration is normal to the manifold). The length of the geodesic path is called the geodesic distance and is defined by a nonlinear function of the principal angles between subspaces, $\theta_i=\cos^{-1}\Sigma_i$ calculated from the full SVD given by $\textbf{U}_1^{\intercal}\textbf{U}_2=\boldsymbol{\Phi\Sigma\Psi}^{\intercal}$.  Several such distance measures on the Grassmann manifold have been proposed in the literature \cite{Hamm:2008}. Regardless of the distance measure, the notion of similarity is maximized when the distance between two subspaces is zero. An issue that often arises in computing distances between subspaces representing physical models is that the subspaces may have different dimensions \cite{giovanis2018uncertainty}. In such cases, the infinite $\mathcal{G}_{\infty, n}$ and the doubly infinite $\mathcal{G}_{\infty, \infty}$ Grassmannian distances proposed by Ye and Lim \cite{YeLim2014} can be used. For a detailed discussion on handling different Grassmann dimensions for surrogate construction the reader is refered to \cite{giovanis2018uncertainty}.

% parametrize subspaces of all dimensions regardless of the ambient space .  

% Given the smoothness of the Grassmannian, one can define 
The tangent space $\mathcal{T}_{\textbf{U}_1}$ of the Grassmannian with origin at $\textbf{U}_1$ is a flat inner-product space composed of subspaces on the plane tangential to $\mathcal{G}_{p, n}$ at $\textbf{U}_1$. In the neighborhood of $\textbf{U}_1$, a point $\textbf{U}_2$ can be mapped onto the tangent plan $\mathcal{T}_{\textbf{U}_1}$ using the logarithmic mapping $\log_{\textbf{U}_1}(\textbf{U}_2)=\boldsymbol{\Gamma}$,
% mapping between the Grassmannian and the tangent space can be performed. The logarithmic mapping of $\textbf{U}_2$ onto this tangent space is denoted by $\log_{\textbf{U}_1}(\textbf{U}_2)=\boldsymbol{\Gamma}$, 
while the inverse mapping from the tangent space to the manifold can be performed with the associated exponential map $\exp_{\textbf{U}_1}(\boldsymbol{\Gamma})=\textbf{U}_2$. An analytical formulation of the mappings is presented in Appendix A.  
% Due to its intrinsic geometric properties, the Grassmann manifold 
Points on the manifold can be mapped to a reproducing kernel Hilbert space by using appropriate kernels $\mathcal{K}:\mathcal{G}_{p, n} \times \mathcal{G}_{p, n} \rightarrow \mathbb{R}$, where $\mathcal{K}$ is positive semi-definite and invariant to the choice of basis.  Several families of Grassmannian kernels exist in the literature \cite{NIPS2008_e7f8a7fb}. 

Given a distance function $d\left(\cdot, \cdot\right)$ and a set of $\mathcal{N}$ points on $\mathcal{G}_{p, n}$, the sample Karcher mean ($\hat{\boldsymbol{\mu}}$) and the sample Fr\'{e}chet variance ($\sigma^2$) can be defined as  \cite{Dubey2017FrchetAO}:
\begin{equation}
\label{eq:KarcherMean}
\hat{\boldsymbol{\mu}}= \argmin_{\omega \in \mathcal{G}_{p, n}} \frac{1}{\mathcal{N}}\sum_{i=1}^{\mathcal{N}} d^{2}(\textbf{U}_i, \omega)\in \mathbb{R}^{n \times p}
\end{equation}
and
\begin{equation}
\sigma^2 = \frac{1}{\mathcal{N}}\sum_{i=1}^{\mathcal{N}} d^2(\textbf{U}_i, \hat{\boldsymbol{\mu}}) \in \mathbb{R},
\label{eq:FrechetVariance}
\end{equation}
respectively. Various iterative algorithm such as Newton's method or the first-order gradient descent \cite{absil2004riemannian,Pennec1999, Begelfor2006} be used in order to estimate the sample Karcher mean.  However, a unique optimal solution is not always guaranteed \cite{Begelfor2006, Karcher1977}. If the ensemble of Grassmann points has a radius greater than $\pi/4$, then the exponential and logarithmic maps are no longer bijective, and the Karcher mean is no longer unique \cite{Begelfor2006}.

\subsection{Principal Geodesic Analysis (PGA)}

In Euclidean space, Principal Component Analysis (PCA) is arguably the leading linear method for identifying a low-dimensional subspace that best represents the variability of the data \cite{abdi2010principal}. However, PCA cannot be directly applied to data on the Grassmann manifold due to its nonlinear structure. PGA generalizes the PCA concept to data on manifolds by seeking a sequence of nested geodesic submanifolds that maximizes the sample Fr\'{e}chet variance of the projected data\cite{fletcher2004principal}. %Note that other variants  of PGA such as the probabilistic PGA \cite{NIPS2013_eb6fdc36} can be used if necessary. 

Before introducing PGA, let us briefly recall the basic concepts of PCA in Euclidean space. Consider the vector $\bm{X}=[\bm{x}_1, \bm{x}_2,\dots, \bm{x}_{\mathcal{N}}]^\intercal$ of $\mathcal{N}$ (zero-mean) points $\bm{x}_i \in \mathbb{R}^{d}$.  PCA seeks the orthonormal basis $[\bm{v}_1, \bm{v}_2, \ldots, \bm{v}_k] \in \mathbb{R}^k$ such that the basis vectors, $\bm{v}_i$, define linear subspaces that align with the principal directions of the variance. That is, $\bm{v}_1$ aligns with the direction of maximum variance, $\bm{v}_2$ aligns with the direction of second highest variance, and so on. The basis can be computed as the set of ordered eigenvectors of the sample covariance matrix of the data through a recursive relationship or through SVD of the matrix $\bm{X}^{T}\bm{X}$.
%\begin{equation}
%v_1=\argmax_{||v||=1} \sum_{j=1}^N (v\cdot x_i)^2, \quad v_k=\argmax_{||v||=1} \sum_{j=1}^N \sum_{l=1}^{k-1} [(v_l\cdot x_i)^2 + (v\cdot x_i)^2]
%\end{equation}
% In other words, the subspace $V_k=\text{span}(\{v_1, \ldots, v_k\})$ is the $k-$dimensional subspace that maximizes the variance of the data projected to that subspace.  

% Working with data on manifolds requires an extension of concepts such as variance, subspace, and projection to the manifold. Let us discuss these concepts in more detail. It is well-known that in Euclidean space the variance $\sigma^2$ of a scalar random variable  $x \in \mathbb{R}$ with mean $\mu$ is defined as $\sigma^2 = \mathbb{E}[(x-\mu)^2]$,  while for a vector-valued random variable $\textbf{x}\in\mathbb{R}^d$ with mean $\boldsymbol{\mu}$ the covariance $\Sigma$ is defined as $\Sigma = \mathbb{E}[(\textbf{x}-\boldsymbol{\mu})(\textbf{x}-\boldsymbol{\mu})^\intercal]$. However, this definition does not hold for general manifolds. Therefore, 
PGA generalizes PCA by first extending the notion of a linear subspace. For general manifolds, the generalization of a linear subspace is referred to as a \textit{geodesic submanifold}. Recall that a geodesic is a curve with the shortest length between points on the manifold; hence it is the generalization of a straight line. It is therefore natural to represent one-dimensional subspaces on a manifold, $\mathcal{M}$, using geodesics.  A submanifold, $\mathcal{H}$, is called geodesic if all geodesics of $\mathcal{H}$ passing through $x$ are also geodesics of the manifold $\mathcal{M}$. However, the geodesics of a submanifold $\mathcal{H}$ of a manifold $\mathcal{M}$ are not necessarily geodesics of $\mathcal{M}$ itself. For example, the sphere $\mathcal{S}^2$ is a submanifold of $\mathbb{R}^3$, but its geodesics are great circles, which are not geodesics of $\mathbb{R}^3$ -- whose geodesics are straight lines. Importantly, geodesic submanifolds defined at a point $\bm{x}$ preserve distances to $\bm{x}$.

PGA then defines  the ``principal directions'' by identifying the geodesic submanifolds that best represent the Fr\'{e}chet variance -- referred to as \textit{principal geodesic submanifolds}. Because geodesic submanifolds preserve the distance to a point $\bm{x}$ and the  Fr\'{e}chet variance is defined in terms of a mean square distance to the Karcher mean, the principal geodesic submanifolds defined at the Karcher mean are are natural generalization of the principal linear subspaces in PCA.  
% In order to generalize PCA in manifolds we need to define the the analog of 
% The principal direction is a one-dimensional subspace on the tangent space whose image under the exponential map is referred to as the \textit{principal geodesic submanifold}. A submanifold $\mathcal{H}$  is called geodesic if all geodesics of $\mathcal{H}$ passing through $x$ are also geodesics of the manifold . However, the geodesics of a submanifold $\mathcal{H}$ of a manifold $\mathcal{G}_{p, n}$ are not necessarily geodesics of the manifold itself. In order for this to be true, geodesic submanifolds at the Karcher mean $\hat{\boldsymbol{\mu}}$ on $\mathcal{G}_{p, n}$ (since the Fr\`{e}chet  variance is defined as the average squared distance to the mean $\hat{\boldsymbol{\mu}}$) are utilized as the generalization of the PCA's linear subspaces on manifolds. 
 
The projection of a point $\textbf{U}_i \in \mathcal{G}_{p, n}$ onto a geodesic submanifold $\mathcal{H}$ is defined as the point  $y \in \mathcal{H}$ that is nearest to $\textbf{U}_i$ in Riemannian distance:
% Therefore, PGA seeks for a sequence of nested geodesic subspaces (submanifolds) that maximizes the sample Fr\`{e}chet variance of the projected data. These submanifolds are called the principal geodesic submanifolds. Since the geodesic is a curve with the shortest length between points on the manifold, it is natural to consider it as the generalization of a straight line, i.e., as the one-dimensional subspace that is the analog to the first principal direction in PCA. However, the geodesics of a submanifold $\mathcal{H}$ of $\mathcal{G}_{p, n}$ are not necessarily geodesics of $\mathcal{G}_{p, n}$. Given that PGA utilizes the sample Fr\`{e}chet variance, a basic property of PGA is that submanifold geodesics at the Karcher mean $\hat{\boldsymbol{\mu}}$ on  $\mathcal{G}_{p, n}$ preserve distances to $\hat{\boldsymbol{\mu}}$. Thus,  geodesic submanifolds  at $\hat{\boldsymbol{\mu}}$ are to be considered as generalizations of the linear subspaces of PCA. The projection of a point $\textbf{U}_i \in \mathcal{G}_{p, n}$ onto a geodesic submanifold $\mathcal{H}$ is defined as the point on $\mathcal{H}$ that is nearest to $\textbf{U}_i$ in Riemannian distance, i.e., 
 \begin{equation}
 \pi_{\mathcal{H}}(\textbf{U}_i) = \argmin_{y\in \mathcal{H}} d^2(\textbf{U}_i, y).
 \end{equation}
However, existence and uniqueness of the point $y$ is only guaranteed by restricting the optimization to a small enough neighborhood around the mean. Projection of the point $\textbf{U}_i$ onto a geodesic submanifold can be approximated linearly in the tangent space  $\mathcal{T}_{\hat{\boldsymbol{\mu}}}$ of $\mathcal{G}_{p, n}$. 
% at the mean $\hat{\boldsymbol{\mu}}$. % of $[\textbf{U}_1, \textbf{U}_2, \ldots, \textbf{U}_N]$. 
Let $\mathcal{H} \subset \mathcal{G}_{p, n}$ be a geodesic submanifold at a point $\omega \in\mathcal{G}_{p, n}$, then the 
% and $\textbf{U}_i \in \mathcal{G}_{p, n}$ be a point to be projected on $\mathcal{H}$. The  
projection $\textbf{U}_i$ is approximated by
 \begin{equation}
 \pi_{\mathcal{H}}(\textbf{U}_i) = \argmin_{y\in \mathcal{H}}||\text{Log}_{\textbf{U}_i}(y)||^2 \approx \argmin_{\bm{y}\in \mathcal{H}}||\text{Log}_{\omega}(\textbf{U}_i) - \text{Log}_{\omega}(\bm{y})||^2
 \label{eq.project}
 \end{equation}
where $\text{Log}_{\omega}(\bm{y})$ is simply a vector in $\mathcal{T}_{\omega}\mathcal{H}$. We can rewrite Eq.~\eqref{eq.project} in terms of tangent vectors as 
  \begin{equation}
 \text{Log}_{\omega}(\pi_{\mathcal{H}}(\textbf{U}_i))  \approx \argmin_{\bm{v}\in \mathcal{T}_{\omega}\mathcal{H}}||\text{Log}_{\omega}(\textbf{U}_i) - \bm{v}||^2
 \end{equation}
where $\text{Log}_{\omega}(\pi_{\mathcal{H}}(\textbf{U}_i))$ implies that the projection operation in the tangent space, transforming the minimization on the manifold into to one in the 
% stands for the minimization formula for the linear projection of $ \text{Log}_{\omega}(\textbf{U}_i)$ onto 
the linear subspace $\mathcal{T}_{\omega}\mathcal{H}$. Therefore, if $\bm{v}=[\bm{v}_1, \bm{v}_2, \ldots, \bm{v}_k]^\intercal$ is an orthonormal basis for  $\mathcal{T}_{\omega}\mathcal{H}$, the projection operator can be approximated by 
 \begin{equation}
  \text{Log}_{\omega}(\pi_{\mathcal{H}}(\textbf{U}_i)) \approx \sum_{i=1}^{k} \langle \bm{v}_i, \text{Log}_{\omega}(\textbf{U}_i) \rangle
 \end{equation}

In the same way that PCA seeks to find the set of linear subspaces that maximize the projected variance, PGA aims to identify the set of nested geodesic submanifolds that maximize the projected Fr\'{e}chet variance of the data, i.e.\ the principal geodesic submanifolds. 
 % For a set of points on the Grassmann manifold, in order 
To define principal geodesic submanifolds from a set of points on the Grassmann manifold, we construct an orthonormal basis of vectors $\bm{v}_1, \bm{v}_2, \ldots, \bm{v}_k$ spanning the tangent space $\mathcal{T}_{\hat{\boldsymbol{\mu}}}$ defined at the Karcher mean $\hat{\boldsymbol{\mu}}$. These vectors form a sequence of nested subspaces $V_k=\text{span}(\bm{v}_1, \bm{v}_2, \ldots, \bm{v}_k) \cap U$, where $U \subset \mathcal{T}_{\hat{\boldsymbol{\mu}}}$ is a neighborhood around $\hat{\boldsymbol{\mu}}$ in which the projection operation is well defined for all geodesic submanifolds of $\text{Exp}_{\bm{\hat{\mu}}}(U)$. The principal geodesic submanifolds are the images of $V_k$ under the exponential map $\mathcal{H}_k = \text{Exp}_{\hat{\boldsymbol{\mu}}}(V_k)$. We start by defining the first principal direction to maximize the projected Fr\'{e}chet variance along the corresponding geodesic:
 \begin{equation}
 v_1 = \argmax_{||\bm{v}||=1} \sum_{i=1}^{\mathcal{N}} ||\text{log}_{\hat{\boldsymbol{\mu}}}( \pi_{\mathcal{H}_1}(\textbf{U}_i))||^2
 \end{equation}
 where $\mathcal{H}_1 =\text{Exp}_{\hat{\boldsymbol{\mu}}}(\text{span}(\{\bm{v}\} \cap U))$. The remaining principal directions are defined recursively as
  \begin{equation}
  v_k = \argmax_{||v||=1} \sum_{i=1}^{\mathcal{N}} ||\text{log}_{\hat{\boldsymbol{\mu}}}( \pi_{\mathcal{H}_k}(\textbf{U}_i))||^2
 \end{equation}
where $\mathcal{H}_k =\text{Exp}_{\hat{\boldsymbol{\mu}}}(\text{span}(\{v_1, \ldots, v_{k-1}, v\} \cap U))$. 

Practically speaking, PGA is performed by simply applying standard PCA in  the tangent space $\mathcal{T}_{\hat{\boldsymbol{\mu}}}$. The algorithm for PGA on the Grassmann manifold is given in Algorithm~\ref{alg-PGA}.

  \begin{algorithm}[H]
\caption{Principal Geodesic Analysis on the Grassmann}\label{alg:cap1}
\begin{algorithmic}
\Require Points $\textbf{U}_1, \textbf{U}_2, \ldots, \textbf{U}_{\mathcal{N}} \in \mathcal{G}_{p, n}$
%\Ensure $y = x^n$
\State Calculate $\hat{\boldsymbol{\mu}}= \argmin_{\omega \in \mathcal{G}_{p, n}} \frac{1}{N}\sum_{i=1}^{\mathcal{N}} d^{2}(\textbf{U}_i, \omega)$  \Comment{Karcher mean}
\State Find $\bm{u}_i=\text{log}_{\hat{\boldsymbol{\mu}}}(\textbf{U}_i)$  \Comment{Projection on the tangent space $\mathcal{T}_{\hat{\boldsymbol{\mu}}}$}
\State $\textbf{S}=\frac{1}{N} \sum_{i=1}^N \bm{u}_i \bm{u}_i^\intercal$ \Comment{Covariance matrix of tangent vectors}
\State $\text{Eig}(\textbf{S})\to\{\bm{v}_k, \lambda_k\}$ \Comment{Eigenvectors and eigenvalues of $\textbf{S}$}
\end{algorithmic}
\label{alg-PGA}
\end{algorithm}
  
\section{Dimension reduction on the Grassmann based using local PGAs}

% Then, we discuss a method for dimensionality reduction based on local PGAs on the Grassmann manifold. We also propose an iterative method for unsupervised learning on the Grassmannian based on the minimization of the sample Fr\`{e}chet variance and present some technical details regarding the implementation.

Consider a model $\mathcal{M}(\boldsymbol{\theta})$ (e.g.,  a stochastic partial differential equation) 
% solved for realizations $\boldsymbol{\theta}_i = [\theta_1^{(i)}, \ldots, \theta_d^{(i)}] 
where $\boldsymbol{\theta} \in \mathbb{R}^{d}$ is a random vector having joint probability distribution function (PDF) $\varrho_{\boldsymbol{\theta}}\left(\boldsymbol{\theta}\right)$ defined on the probability space $\left(\Omega, \Sigma, \mathbb{P}\right)$, where $\Omega$ is the sample space, $\Sigma$ the set of events, and $\mathbb{P}$ the probability measure. Given an experimental design $\boldsymbol{\Theta}=\{\boldsymbol{\theta}_1 , \boldsymbol{\theta}_2 , \ldots, \boldsymbol{\theta}_{\mathcal{N}}\}$, let us denote the corresponding full-field solutions (responses) as $\textbf{y}_i=\mathcal{M}(\boldsymbol{\theta}_i) \in \mathbb{R}^n$, which are assumed to have very high dimensionality, e.g., $n \sim \mathcal{O}(10^{4-6})$.  
% with random samples and the corresponding responses $\mathcal{Y}=\{\textbf{y}_1, \textbf{y}_2, \ldots, \textbf{y}_N\}$, 
We first aim to reduce the dimension of each response $\textbf{y}_i$ in such a way that its salient information, such as the intrinsic geometry of the solution space, is retained. Note that for most real-world applications, the size of the experimental design is usually small to medium, e.g., $\mathcal{N} \sim\mathcal{O}(10^{1-3})$.

\begin{figure}[t!]
\begin{center}
\includegraphics[width=0.7\textwidth]{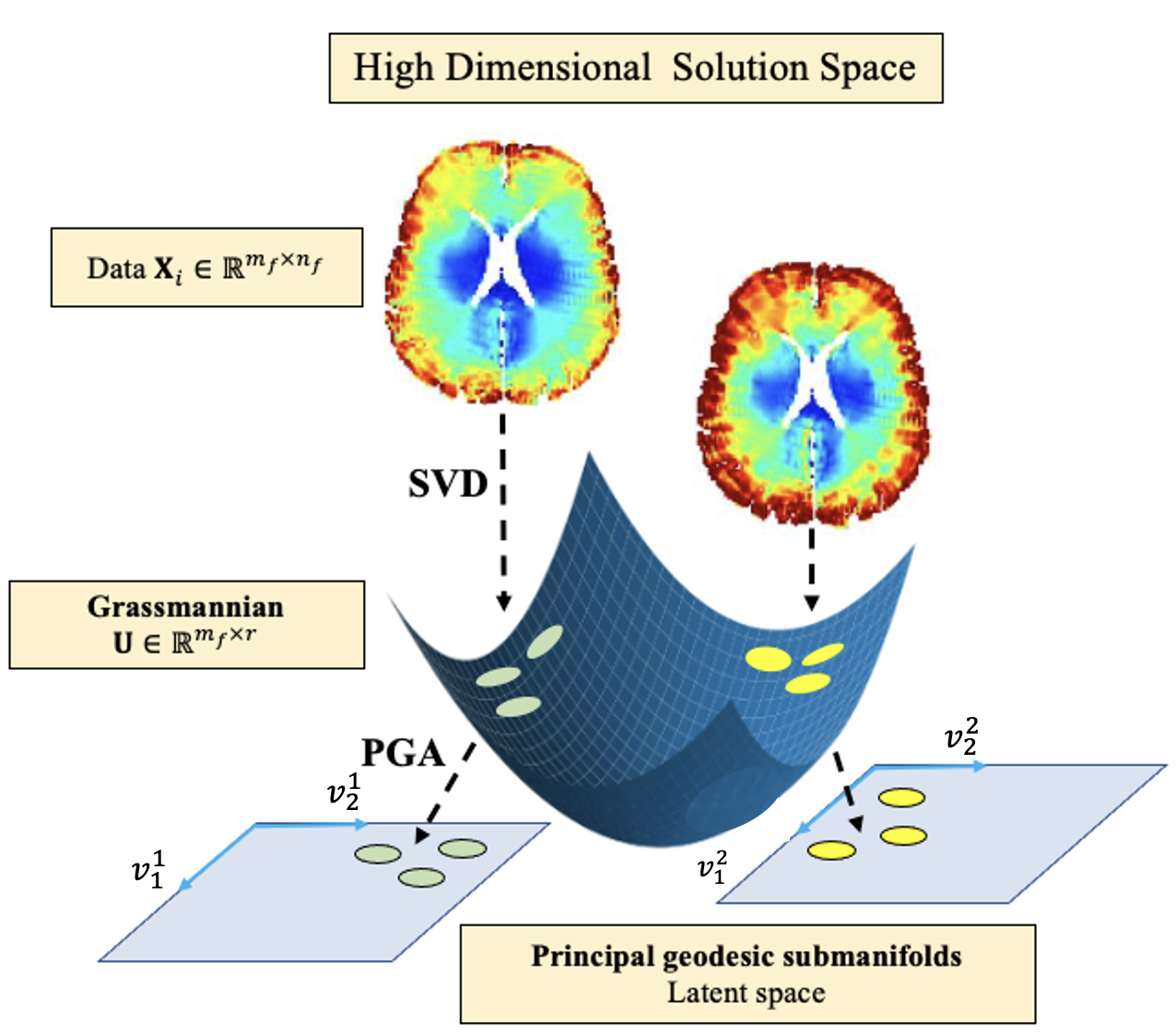}
\caption{Illustration of the proposed two-phase dimension reduction scheme using Grassmann manifold projection and principal geodesic analysis. The strain field images of computational head models are adopted from \cite{upadhyay2022data}. Each circle on the manifold corresponds to a realization of the strain field. Distinct colors represent different clusters on the Grassmann.}
\label{fig:method}
\end{center}
\end{figure}

The proposed dimension reduction method is performed in two stages: First, each response $\textbf{y}_i$ is projected onto a Grassmann manifold using thin SVD \cite{giovanis2020data, Kontolati2021ManifoldLP, dos2022grassmannian, dos2022grassmannian_b}. In the second stage, we perform unsupervised learning on the Grassmann using an extension of K-means for clustering on the manifold \cite{10.1007/978-3-030-32520-6_42}. Since the optimal data partition (number of clusters) is not know \textit{a priori}, we propose a method to identify the minimum necessary number of clusters automatically through minimization of the sample Fr\'{e}chet variance. Then, for each cluster, we perform PGA to further reduce the dimension of the data. An illustration of the proposed two-stage dimension reduction method is shown in Figure~\ref{fig:method}, while a detailed description follows.

\subsection{Projection on the Grassmann manifold}

To exploit the subspace structure of the high-dimensional model response, we first project each response $\textbf{y}_i \in \mathbb{R}^{n(=m_f\times n_f)}$  onto the Grassmannian by means of thin-SVD, such that $\textbf{y}_i = \textbf{U}_i \boldsymbol{\Sigma}_i\textbf{V}_i^\intercal$, where $\textbf{U}_i \in \mathbb{R}^{m_f \times p_i}$ and $\textbf{V}_i \in \mathbb{R}^{n_f \times p_i}$ are orthonormal matrices, i.e., $\textbf{U}_i^\intercal\textbf{U}_i=\textbf{I}_{p_i}$, $\textbf{V}_i\textbf{V}_i^\intercal=\textbf{I}_{p_i}$, and $\boldsymbol{\Sigma}_i \in \mathbb{R}^{p_i \times p_i}$ is a diagonal matrix whose non-zero elements are the singular values ordered by magnitude. Note that $p_i$ is the rank of $\textbf{y}_i$, which is determined during the SVD based on a prescribed tolerance. By performing SVD, each response is now represented by two matrices i.e., points on two Grassmann manifolds, and a vector i.e., a point in Euclidean space. In general, two responses $\textbf{y}_i$ and $\textbf{y}_j$ might have different ranks ($p_i \neq p_j$), thus also a different number of columns. This means that $\textbf{U}_i$ and $\textbf{U}_j$  lie on different Grassmann manifolds, i.e., $\textbf{U}_i \in \mathcal{G}_{p_i, m_f}$ and $\textbf{U}_j \in \mathcal{G}_{p_j, m_f}$. In this case, we can embed the points into the infinite Grassmannian $\mathcal{G}_{\infty, m_f} \equiv \mathcal{G}_{p_{\infty, m_f}}$, with $p_{\infty}=\max(p_1, p_2, \ldots,)$, and perform the calculations there.

\subsection{Unsupervised learning on the Grassmann}

Given a set of points on the Grassmann manifold, we next perform a modified K-means clustering \cite{MacQueen1967SomeMF}.  For a prescribed number of clusters, $n_c$, K-means partitions data $(x_1, x_2, \ldots, x_{\mathcal{N} })$ by minimizing the objective function:
\begin{equation}
    J = \sum_{j=1}^\mathcal{N}  \sum_{k=1}^{n_c} r_{jk}||x_j - \mu_k||^2
\end{equation}
where $\mu_k$ are the centroids, and $r_{jk}$ is equal to 1 if $x_j \in S_k$ and 0 otherwise, where $S_k=\{x_j: ||x_j-\mu_k||^2 \leq ||x_j-\mu_m||^2, \forall m \neq k\}$. The algorithm starts by randomly choosing $n_c$ points from the data set to serve as initial guesses of the cluster centroids. The distance from each data point to each centroid is computed, and the data is assigned to its closest centroid. The centroid then is replaced by the mean of the partitioned data and the steps are repeated until all $n_c$ centroids are identified. 

K-means is effective for data in Euclidean space where the notion of similarity is defined through the Euclidean distance. However, when the data have intrinsic geometric features constrained on some manifold, K-means performed in Euclidean space may fail to capture similarity. A variant of K-means for data that lie on Riemmannian manifolds (such as the Grassmannian) was introduced in \cite{10.1007/978-3-030-32520-6_42}.  In this variation, the objective function to minimize takes the form:
 \begin{equation}
 \label{kmeans_Gr}
    J = \sum_{j=1}^\mathcal{N}  \sum_{k=1}^{n_c} r_{jk}d_{\mathcal{G}}(\textbf{U}_j, \boldsymbol{\mu}_k)^2 
\end{equation}
where %$\log_{x_n}(\mu_k)$ is the logarithmic map that projects point $\mu_k$ onto the tangent space with origin $x_n$, and 
$d_{\mathcal{G}}(\cdot)$ is the Riemmannian distance (i.e.\ Grassmann distance) and $r_{jk}$ is equal to 1 if $\textbf{U}_j \in S_k$ and 0 otherwise, where $S_k=\{\textbf{U}_j: d_{\mathcal{G}}(\textbf{U}_j, \boldsymbol{\mu}_k)^2 \leq d_{\mathcal{G}}(\textbf{U}_j, \boldsymbol{\mu}_m), \forall m \neq k\}$. Again, the algorithm starts by randomly selecting $n_c$ points from the data as the centroids $\boldsymbol{\mu}_k$ and computing the Riemmannian distance (using the logarithmic map) between each point $\textbf{U}_j,$ and the centroids. Then, each point is assigned to its closest centroid which is then replaced by the Karcher mean of the partitioned data (computed by the stochastic gradient descent method \cite{Begelfor2006}). The process is repeated until all $n_c$ centroids are identifed. 

K-means requires the number of clusters, $n_c$, to be specified a priori. However, this number is not known beforehand for arbitrary models where variations in performance are not already known.
% it is difficult to assess the similarity of high-dimensional models. 
We propose a novel iterative algorithm to identify the optimal number of clusters for points on the Grassmann that begins with a small number of clusters, i.e., $n_{c}=2$. At each iteration ($iter$) we increase the number of clusters by one, partition the data using Riemmannian K-means, and compute the Karcher mean $\boldsymbol{\mu}_c$ and the sample Fr\'{e}chet variance $\sigma_i^2$  for each cluster. 
Next, we compute the sample Fr\'{e}chet variance of all Karcher means $\boldsymbol{\sigma}^2_{\hat{\boldsymbol{\mu}}}$ and we compute the following score:
\begin{equation}
    \lambda_{S_h}^{n_c} = \frac{\boldsymbol{\sigma}^2_{\hat{\boldsymbol{\mu}}}}{\sum_{i=2}^{n_c} \sigma^{2}_i}.
\end{equation}
which is quantifies the goodness of the clustering. The definition of this coefficient is similar to the definition of the silhouette Coefficient used to validate the clustering when dealing with data in Euclidean spaces. 
% To compute the Karcher mean and minimize the loss function, we use the stochastic gradient descent method \cite{Begelfor2006}. 
%Then, the mean sample Fr\`{e}chet variance, for all clusters is computed according to
%\begin{equation}
%    \overline{f}_{iter} = \frac{1}{n_{c}} \sum_{i=1}^{n_{c}} \sigma_i^2.
%\end{equation}
%\quad \text{std}_{\overline{f}, iter}=\sqrt{\frac{1}{n_c} \sum_{i=1}^{n_c} (\sigma_i -\overline{f}_{iter})^2}
We repeat this process until the number of points $N_h$ in at least one of the clusters is less or equal to a minimum value, e.g., $N_h \leq 5$. The optimal number of clusters corresponds to the largest value of the silhouette coefficient $\lambda_{S_h}^{n_c}$.

%(1) we have achieved a uniform distribution of the sample Fr\`{e}chet variances $\sigma_i^2$, (with $h=1, \ldots, n_c$), i.e., the variance $\text{var}_{\overline{f}_{iter}}$ of $\overline{f}_{iter}$ is zero or for practical purposes below  a prescribed threshold, e.g.,  $\text{var}_{\overline{f}_{iter}}\leq 0.1$; %(2) the mean sample Fr\`{e}chet variance is below a predefined threshold, e.g., $\overline{f}_{iter} \leq 0.1$; 
%(2) a maximum number of iterations has been reached; (3) the number of points $N_h$   in at least one of the clusters is less or equal to a minimum value, e.g., $N_h \leq 5$. The steps of the iterative approach are presented in Algorithm~\ref{alg:cap}. Note that the clustering is performed on the manifold of the left eigenvectors. This could also be performed with the right eigenvectors or using a composite distance -- depending on the user's preferences and the dimension of the matrices.
% This is justified by to the fact that each model response is stored as a tall and skinny matrix, as a result of thin-SVD. Hence, the important information regarding the geometry of the response is mostly encapsulated in the left eigenvectors.

\begin{algorithm}[t!]
\caption{Optimization based on Riemmannian K-means on the Grassmann}\label{alg:cap}
\begin{algorithmic}
\Require minimum number of points in a cluster $N_h \geq 5$
%\Ensure $y = x^n$
\State $iter = 1$  \Comment{iterations number}
%\State $n_{\max} = 50$ \Comment{maximum number of iterations (depends on the number of data)}
\State $\lambda_{S_h}[0] = 0$
\State $l = 2$      \Comment{total number of clusters}
%\State $\alpha = 0.1$ \Comment{threshold for mean sample Fr\`{e}chet variance}
\State $n_c = 2$  \Comment{minimum number of clusters}
\State $\overline{f}_{iter} = \infty$
\While{$\min({N_1, \ldots, N_{n_c}}) \leq 5$}
\State Perform Riemmannian K-means with $n_c$ clusters
\For{$h=1:n_c$}
    \State $\sigma_h^2 =1/{N_h}\sum_{i=1}^{N_h} d_{\mathcal{G}}^2(\textbf{U}_i, \hat{\boldsymbol{\mu}}_{\textbf{U}, h})$ \Comment{Compute the sample Fr\'{e}chet variance}
\EndFor
    \State $\lambda_{S_h}=\frac{1}{n_c} \sum_{c=2}^{n_c} \lambda_{S_h}^{(c)}$     \Comment{Silhouette coefficient}
    %\State $\overline{f}_{iter} = 1/n_c\cdot\sum_{i=1}^{n_c} \sigma_i^2$ 
    \State $iter=iter+ 1$ \Comment{next iteration}
    \State $n_c = n_c + 1$ \Comment{increase number of clusters by one}
\EndWhile
\State Perform Riemmannian K-means with $n_c$ clusters
\end{algorithmic}
\end{algorithm}

\subsection{Local PGA on the Grassmann}

Once the optimal number of clusters $K$ has been identified and the data on the manifold have been appropriately partitioned into clusters  $C_h$, $h=1,\ldots, n_c$, a second dimension reduction is performed using PGA on each cluster. The goal is to embed the points in latent geodesic submanifolds that best describes the geometric variability of the data on the manifold. PGA is performed in three steps, for each cluster $C_i$ \cite{fletcher2004principal}: (1) lift the points from the manifold to a well-defined tangent space using the logarithmic map; (2) perform PCA in the tangent space; (3) map the results back to the manifold by an exponential map. These steps are presented in detail next.\\

\underline{Step 1}: Project points $\{\textbf{U}_j, \textbf{V}_j\}_{j=1}^{N_h} \in C_h$, where $N_h$ is the total number of data in cluster $h$, onto the tangent spaces with origins at the corresponding Karcher means $\hat{\boldsymbol{\mu}}_{\textbf{U},h}$ and $\hat{\boldsymbol{\mu}}_{\textbf{V},h}$, i.e.,
\begin{subequations}
\begin{align}
\left\{\textbf{U}_j\in \mathcal{G}_{p_{\infty}, m_f}\right\}_{j=1}^{N_h} & \xrightarrow{ \text{Log}_{\hat{\mu}} }  \{\boldsymbol{\Gamma}^{u}_{j}\in \mathcal{T}_{\hat{\boldsymbol{\mu}}_{\textbf{U},h}}\}_{j=1}^{N_h}, \\
\left\{\textbf{V}_j\in \mathcal{G}_{p_{\infty}, N_f}\right\}_{j=1}^{N_h} & \xrightarrow{ \text{Log}_{\hat{\mu}} }  \{\boldsymbol{\Gamma}^{v}_{j}\in \mathcal{T}_{\hat{\boldsymbol{\mu}}_{\textbf{V},h}}\}_{j=1}^{N_h},
\end{align}
\end{subequations}
where the indices $u, v$ of $\boldsymbol{\Gamma}_{j}$ represent actions on the  left eigenvector $\textbf{U}_j$ and the right eigenvector $\textbf{V}_j$ in the tangent space, respectively (see \ref{appendix_1} for details on computation of matrices $\boldsymbol{\Gamma}$).  Ensure $\boldsymbol{\Gamma}^{u}_{j}\in U_u^h, \forall j$ where $U_u^h \subset \mathcal{T}_{\hat{\boldsymbol{\mu}}_{\textbf{U},h}}$ is a neighborhood such that the projection $\text{Exp}_{\hat{\boldsymbol{\mu}}_{\textbf{V},h}}(U_u^h)$ is well-defined. 
% for all geodesic submanifolds . 
Similarly ensure $\boldsymbol{\Gamma}^{u}_{j}\in U_v^h \subset \mathcal{T}_{\hat{\boldsymbol{\mu}}_{\textbf{V},h}}$.\\

\underline{Step 2}: For each cluster $C_h$, construct an orthonormal basis of tangent vectors. The first principal direction is chosen to maximize the projected variance along the corresponding geodesic, while the remaining principal directions are defined recursively.

\begin{subequations}
\small
\begin{align}
\boldsymbol{\xi}_1=\argmax_{||\boldsymbol{\xi}||=1} \sum_{i=1}^{N_h} ||\log_{\hat{\boldsymbol{\mu}}_{\textbf{U},h}}(\pi_{\mathcal{H}}(\textbf{U}_i)||^2, \quad & \mathcal{H}=\exp_{\hat{\boldsymbol{\mu}}_{\textbf{U},h}}(\text{span}(\{\boldsymbol{\xi}\})\cap U_u^h) \\
\boldsymbol{\xi}_k=\argmax_{||\boldsymbol{\xi}||=1} \sum_{i=1}^{N_h}  ||\log_{\hat{\boldsymbol{\mu}}_{\textbf{U},h}}(\pi_{\mathcal{H}}(\textbf{U}_i)||^2, \quad & \mathcal{H}=\exp_{\hat{\boldsymbol{\mu}}_{\textbf{U},h}}(\text{span}(\{\boldsymbol{\xi}_1, \ldots, \boldsymbol{\xi}_{d_u-1}, \boldsymbol{\xi}\})\cap U_u^h)\\
\boldsymbol{\phi}_1=\argmax_{||\boldsymbol{\phi}||=1} \sum_{i=1}^{N_h} ||\log_{\hat{\boldsymbol{\mu}}_{\textbf{V},h}}(\pi_{\mathcal{H}}(\textbf{V}_i)||^2, \quad & \mathcal{H}=\exp_{\hat{\boldsymbol{\mu}}_{\textbf{V},h}}(\text{span}(\{\boldsymbol{\phi}\})\cap U_v^h)\\
\boldsymbol{\phi}_k=\argmax_{||\boldsymbol{\phi}||=1} \sum_{i=1}^{N_h}  ||\log_{\hat{\boldsymbol{\mu}}_{\textbf{V},h}}(\pi_{\mathcal{H}}(\textbf{V}_i)||^2, \quad & \mathcal{H}=\exp_{\hat{\boldsymbol{\mu}}_{\textbf{V},h}}(\text{span}(\{\boldsymbol{\phi}_1, \ldots, \boldsymbol{\phi}_{d_v-1}, \boldsymbol{\phi}\})\cap U_v^h)
\end{align}
\end{subequations}
In practice, this is performed by simply applying standard PCA on the tangent vectors \cite{geomstats_paper}. 

\underline{Step 3}: Form a sequence of nested subspaces. 
\begin{subequations}
\label{eq.subspaces}
\begin{align}
\mathcal{U}_k^h& =\text{span}(\boldsymbol{\xi}_1, \boldsymbol{\xi}_2, \ldots, \boldsymbol{\xi}_{d_u}; \in \mathcal{T}_{\hat{\boldsymbol{\mu}}_{\textbf{U},h}}) \\
\mathcal{V}_k^h& =\text{span}(\boldsymbol{\phi}_1, \boldsymbol{\phi}_2, \ldots, \boldsymbol{\phi}_{d_v}; \in \mathcal{T}_{\hat{\boldsymbol{\mu}}_{\textbf{V},h}}) 
\end{align}
\end{subequations}
where $d_u$ and $d_v$ are the dimensions of the corresponding subspaces. The principal geodesic submanifolds are the images of the $\mathcal{U}_k^h$ and $\mathcal{V}_k^h$  under the exponential maps $\text{Exp}_{\hat{\boldsymbol{\mu}}_{\textbf{U},h}}(\mathcal{U}_k^h)$ and $\text{Exp}_{\hat{\boldsymbol{\mu}}_{\textbf{V},h}}(\mathcal{V}_k^h)$, respectively. 

In general, each cluster may require a different number of principal geodesic submanifolds in order to maximize the projected variance of the data. This number will depend on variability within each cluster. Therefore, by comparing the number of principal geodesic submanifolds across the clusters, we can identify regions of the space over which sharp changes in system behavior occur.

%To this end, we  propose to exploit the coherent structure of each data point by linearly projecting it onto the Grassmann manifold and then to further reduce the dimension of the data using Principal Geodesic Analysis (PGA). The second dimension reduction will reveal the underlying geometry of the Grassmannian. Herein, the basic elements of the Grassmann manifold (drawn from \cite{giovanis2020data, Kontolati2021ManifoldLP, giovanis2018uncertainty}) and PGA are briefly presented.

\section{Geodesic submanifold interpolation via polynomial chaos expansions}

In this section, we propose a method to construct polynomial chaos surrogate models on the lower-dimensional geodesic submanifolds. First, the basic concepts of polynomial chaos expansion (PCE) are discussed. This is followed by a PCE formulation to map from the input space to the low-dimensional latent space. Finally, the exponential mapping and inverse SVD are used to reconstruct the full-field response.

\subsection{Polynomial chaos expansion}

PCE is one of the most widely-used surrogate modeling techniques to accelerate UQ tasks in computational physics-based modeling \cite{feinberg2018multivariate, jakeman2019polynomial, rahman2018polynomial}.  PCE approximates the response of the model $\mathcal{M}\left(\boldsymbol{\theta}\right)$ by an expansion of multi-dimensional polynomials that are orthogonal with respect to the PDF of the input random variables $\boldsymbol{\theta}$. By the Doob-Dynkin lemma \cite{bobrowski2005functional}, the response of the model is also a random variable parametrized by $\boldsymbol{\theta}$. Note that in our case we consider a vector-valued response, i.e., $\mathcal{M}(\boldsymbol{\theta}) \equiv \textbf{y} \in \mathbb{R}^n$.

Provided that the quantity $\mathbb{E}[||\textbf{y}||^2]$ (where $||\cdot||$ is the Euclidean norm) is finite, PCE is a spectral approximation of the form
\begin{equation}
\label{eq:spectral_exact}
\mathcal{M}(\boldsymbol{\theta}) \approx \widetilde{\mathcal{M}}(\boldsymbol{\theta}) = \sum_{\boldsymbol{\alpha} \in \mathbb{N}^d} \textbf{a}_{\boldsymbol{\alpha}} \Psi_{\boldsymbol{\alpha}}(\boldsymbol{\theta}),
% \quad \text{where} ~ \textbf{a}_{\boldsymbol{\alpha}} = \{a_{\boldsymbol{\alpha}}^{(1)}, \ldots, a_{\boldsymbol{\alpha}}^{(n)}\}.
\end{equation}
where the coefficients $\textbf{a}_{\boldsymbol{\alpha}}=[a_{\boldsymbol{\alpha}}^{(1)}, \ldots, a_{\boldsymbol{\alpha}}^{(n)}]^\intercal$ are deterministic vectors and $\Psi_{\boldsymbol{\alpha}}$ are multivariate polynomials that are orthonormal with respect to the joint PDF $\varrho_{\boldsymbol{\theta}}$, such that $\mathbb{E}\left[\Psi_{\boldsymbol{\alpha}} \Psi_{\boldsymbol{\beta}}\right] = 1$ if $\boldsymbol{\alpha} = \boldsymbol{\beta}$ and zero otherwise. Depending on the PDF $\varrho_{\boldsymbol{\theta}}$, the orthonormal polynomials are chosen according to the Wiener-Askey scheme \cite{xiu2002wiener} or are constructed numerically \cite{wan2006multi, soize2004physical}.

For practical purposes, the PCE is truncated after a finite number of terms $P$. Typically, one retains the polynomials $\Psi_{\boldsymbol{\alpha}}$ such that a given $q$-norm of $\boldsymbol{\alpha}$ in not greater than a positive integer $s$, i.e., for $q \in(0, 1]$
\begin{equation}
    ||\boldsymbol{\alpha}||_q= \bigg(\sum_{i=1}^d \alpha_i^q\bigg)^{1/q} \leq s, \quad  s \in \mathbb{Z}_{\geq 0},
\end{equation}
which yields the following approximation of Eq.~\eqref{eq:spectral_exact}
\begin{equation}
\label{eq:spectral_approx}
\widetilde{\mathcal{M}}(\boldsymbol{\theta}) = \sum_{||\boldsymbol{\alpha}||_q\leq s} \textbf{a}_{\boldsymbol{\alpha}} \Psi_{\boldsymbol{\alpha}}(\boldsymbol{\theta})
\end{equation}
The choice of $q$ plays a central role in the construction of the PCE, as it defines which polynomials and corresponding coefficients form the PCE.
The most common choice, as well as the one employed in this work, is that of a total-degree polynomials, i.e., $\left\| \boldsymbol{\alpha} \right\|_1 \leq s$. In this case, the number of PCE terms is $P = \frac{\left(s + d\right)!}{s! d!}$. 
Since the size of the PCE  increases rapidly with both $d$ and $s$, for high-dimensional input random variables $\boldsymbol{\theta}$, several sparse PCE algorithms have been proposed in the literature to mitigate the impact of the curse of dimensionality \cite{loukrezis2020robust, blatman2011adaptive, jakeman2015enhancing, hampton2018basis, he2020adaptive, diaz2018sparse}. 

The PCE coefficients in Eq.~(\ref{eq:spectral_approx}) can be computed using several approaches such as the pseudo-spectral projection \cite{knio2006uncertainty, constantine2012sparse, conrad2013adaptive, winokur2016sparse}, interpolation \cite{buzzard2013efficient, loukrezis2019adaptive}, and, most commonly, regression \cite{blatman2011adaptive, loukrezis2020robust, hampton2018basis, diaz2018sparse, hadigol2018least, he2020adaptive, doostan2011non, tsilifis2019compressive} where the PCE coefficients are computed by solving a least-squares problem \cite{rifkin2007notes}. Consider the indices  $\{\boldsymbol{\alpha}_0, \ldots, \boldsymbol{\alpha}_{P-1}\}$ and the $P \times n$ coefficient matrix $\textbf{C}=[\textbf{a}_{\boldsymbol{\alpha}_0}, \ldots, \textbf{a}_{\boldsymbol{\alpha}_{P-1}}]^\intercal$. For a set of realizations $\boldsymbol{\Theta} = \left\{\boldsymbol{\theta}_i\right\}_{i=1}^{N}$ with corresponding model outputs $\bf{Y} = \left\{\mathcal{M}(\boldsymbol{\theta}_i)\right\}_{i=1}^{\mathcal{N}}$, the coefficients are estimated by the solution of:
\begin{align}
\label{eq:regression}
\underset {\mathbf{C} \in \mathbb{R}^{P \times n}}{\min} \left\{\sum_{i=1}^{\mathcal{N} } \left( \mathcal{M}(\boldsymbol{\theta}_i) - \sum_{||\boldsymbol{\alpha}||_q\leq s} \textbf{a}_{\boldsymbol{\alpha}} \Psi_{\boldsymbol{\alpha}}(\boldsymbol{\theta}_i) \right)^2 \right\}
\end{align}
which admits the closed form solution
\begin{equation}
    \hat{\textbf{C}}=\bigg(\boldsymbol{\Psi}^\intercal\boldsymbol{\Psi} \bigg)^{-1}\boldsymbol{\Psi}^\intercal \bf{Y}
\end{equation}
where $\boldsymbol{\Psi}$ is the matrix containing evaluations of the basis polynomials, that is $\boldsymbol{\Psi} = \{\psi_{\boldsymbol{\alpha}_j}(\boldsymbol{\theta}_i),\quad  1\leq i \leq \mathcal{N}, \quad 0 \leq j \leq P-1$\}. However, the sample size $N$ must be greater than $P$ to make this problem well-posed. To overcome this limitation, estimation of the PCE coefficients can be performed by solving a regularized least squares problem, referred to as the LASSO (least absolute shrinkage and selection operator). The LASSO method applies to scalar response quantities. Hence, in our case, it must be applied successively to every response component individually.

\subsection{PCE in local latent spaces}
Next, we train $n_c$ PCEs (one for each cluster) to approximate the reduced solution for any new realization of the random vector $\boldsymbol{\theta}$.  In cluster $C_h$, with $h=1, \ldots, n_c$, (which contains $N_h$ points), the truncated PCA of matrix $\boldsymbol{\Gamma}^{u, v}$ is given by
% containing the (vectorized) tangent vectors $\{\boldsymbol{\gamma}^{u, v}_i\}_{i=1}^{N_h}$ is
\begin{subequations}
\begin{align}
 \hat{\boldsymbol{\Gamma}}^u &= \overline{\boldsymbol{\Gamma}}^u + \sum_{r=1}^{d_{u}} \textbf{B}^u_r\boldsymbol{\xi}_r \\
 \hat{\boldsymbol{\Gamma}}^v &= \overline{\boldsymbol{\Gamma}}^v + \sum_{r=1}^{d_{v}} \textbf{B}^v_r\boldsymbol{\phi}_r 
\end{align}
\end{subequations}
where $\overline{\boldsymbol{\Gamma}}^{u,v}=\mathbb{E}[\boldsymbol{\Gamma}^{u}_r ]$ is the matrix containing the sample mean-vector 
% (replicated $N_h$ times) 
and $\textbf{B}^{u,v}_r$ is the projection of each vector $\boldsymbol{\Gamma}^{u, v}_r$ onto the corresponding basis given by
\begin{subequations}
\begin{align}
\textbf{B}^u_r &= \boldsymbol{\xi}_r^\intercal \bigg(\boldsymbol{\Gamma}^{u}_r -\mathbb{E}[\boldsymbol{\Gamma}^{u}_r ]\bigg) \\
\textbf{B}^v_r &= \boldsymbol{\phi}_r^\intercal \bigg(\boldsymbol{\Gamma}^{v}_r -\mathbb{E}[\boldsymbol{\Gamma}^{v}_r ]\bigg)
\end{align}
\end{subequations}
However, as the response of the model $\mathcal{M}(\boldsymbol{\theta})$ is uncertain, $\textbf{B}_{r}^{u, v}$ is  function of the input random vector $\boldsymbol{\theta}$ and is therefore a random variable. We seek an estimator $\hat{\textbf{B}}_{r}^{u, v}$ using a truncated  PCE as
% of the unknown $\textbf{B}^{u, v}_r$ as:
\begin{equation}\label{eq:PCEB}
\hat{\textbf{B}}^{u, v}_r(\boldsymbol{\theta}) = \sum_{||\boldsymbol{\alpha}_{u,v}||_q\leq s} \textbf{a}_{\boldsymbol{\alpha}_{u,v}} \Psi_{\boldsymbol{\alpha}_{u,v}}.(\boldsymbol{\theta})
\end{equation}

\subsection{Out-of-sample extension}

% As discussed in the previous section, for each cluster, the training data consist of $N_h$ random variable realizations $\{\boldsymbol{\theta}_1, \ldots,\boldsymbol{\theta}_{N_h}\}$, with $\boldsymbol{\theta} \in \mathbb{R}^d$, and the corresponding reduced solutions $\{\textbf{B}_{1}^{u, v}, \ldots, \textbf{B}_{N_h}^{u, v}\}$.
Next, consider that we aim to predict the solution for a new realization of the input random variable, $\boldsymbol{\theta}^\star$. In the previous step, we constructed $n_c$ PCEs to estimate the coefficients $\hat{\textbf{B}}^{u, v}$. 
% for an additional realization of the input random parameters $\boldsymbol{\theta}^\star$. 
For the given new sample, we first must determine which PCE model to use. 
% Among all the training samples, $\{\boldsymbol{\theta}_i\}_{i=1}^N$, 
To do so, we identify the nearest\footnote{Note, that we could identify more than one nearest training samples and use an ensemble of the corresponding trained PCEs to predict the solution. However, in this case consistency issues arise due to the different dimension across clusters. We will investigate the issue of consistency in future work.}  training sample from $\{\boldsymbol{\theta}_i\}_{i=1}^\mathcal{N}$ to $\boldsymbol{\theta}^\star$ based on the Euclidean distance as
% ( ${\|\cdot\|}_2$):
\begin{equation}
\label{eq.nn}
\boldsymbol{\theta}_{\text{index}}= \arg\min\limits_{\boldsymbol{\theta}_i} \big(\|\boldsymbol{\theta}^\star -\boldsymbol{\theta}_i)\|_2\big), \quad i=1, \ldots, \mathcal{N}.
%d(\boldsymbol{\theta}^\star, \boldsymbol{\theta}_i), \quad i=1, \ldots, N
\end{equation}
% This step is necessary for identifying the cluster $h$ in which $\boldsymbol{\theta}_{j_{\min}}$ belongs to and 
We then use the PCE for the cluster containing $\boldsymbol{\theta}_{\text{index}}$ to predict the PGA coefficients $\hat{\textbf{B}}^{u, v}(\boldsymbol{\theta}^\star)$ using Eq.~(\ref{eq:PCEB}) and the tangent vectors are approximated by
% . The inversion of PGA will give as an approximation of the tangent vectors as:
\begin{subequations}
\label{eq.pce_predict1}
\begin{align}
    \tilde{\boldsymbol{\Gamma}}^{u}(\boldsymbol{\theta}^\star) &= \overline{\boldsymbol{\Gamma}}^u + \sum_{r=1}^{d_{u}} \hat{\textbf{B}}^u_r(\boldsymbol{\theta}^\star)\boldsymbol{\xi}_r, \\
    \tilde{\boldsymbol{\Gamma}}^{v}(\boldsymbol{\theta}^\star) &= \overline{\boldsymbol{\Gamma}}^v + \sum_{r=1}^{d_{v}} \hat{\textbf{B}}^v_r(\boldsymbol{\theta}^\star)\boldsymbol{\phi}_r. 
\end{align}
\end{subequations}
To construct the approximate high-dimensional solution $\tilde{\textbf{y}}^{\star} = \tilde{\mathcal{M}}(\boldsymbol{\theta}^\star)$ and map to the original (physically interpretable) space ($ \mathbb{R}^{m_f \times n_f}$), we first perform the exponential mappings:
\begin{subequations}
\begin{align}
    \tilde{\textbf{U}}(\boldsymbol{\theta}^\star) &= \text{Exp}_{\hat{\boldsymbol{\mu}}_{\textbf{U},h}}( \tilde{\boldsymbol{\Gamma}}^{u}(\boldsymbol{\theta}^\star))\\
    \tilde{\textbf{V}}(\boldsymbol{\theta}^\star) &= \text{Exp}_{\hat{\boldsymbol{\mu}}_{\textbf{V},h}}( \tilde{\boldsymbol{\Gamma}}^{v}(\boldsymbol{\theta}^\star))
\end{align}
\end{subequations}
that project the points from the tangent space onto their corresponding Grassmann manifolds. We further construct a PCE approximation of the matrix containing the singular values as:
\begin{equation}
\label{eq.predict_sigma}
 \tilde{\boldsymbol{\Sigma}}(\boldsymbol{\theta}^\star) = \sum_{||\boldsymbol{\alpha}_{\Sigma}||_q\leq s} \textbf{a}_{\boldsymbol{\alpha}_{\Sigma}} \Psi_{\boldsymbol{\alpha}_{\Sigma}}(\boldsymbol{\theta}^\star)
\end{equation}
Finally, we use the inverse SVD to obtain a prediction of the full-field solution as:
\begin{equation}
\label{eq.invert_svd}
    \tilde{\textbf{y}}^{\star} = \tilde{\textbf{U}}(\boldsymbol{\theta}^\star)\tilde{\boldsymbol{\Sigma}}(\boldsymbol{\theta}^\star)\tilde{\textbf{V}}^\intercal(\boldsymbol{\theta}^\star)
\end{equation}

To assess the accuracy of predictions from the proposed method, we will use the following two scalar metrics \cite{Kontolati2021ManifoldLP}. The first scalar metric is the relative $L_2$ error in the original space given by
\begin{equation}
\label{eq:L2-error}
    L_2(\tilde{\textbf{y}}^{\star}, \textbf{y}^\star) = \frac{\|\tilde{\textbf{y}}^{\star}- \textbf{y}^\star\|_2}{\|\textbf{y}^\star\|_2},
\end{equation}
where $\textbf{y}^\star$ is the reference response. The second scalar metric is the coefficient of determination (i.e., the $R^2$ score) given by:
\begin{equation}
\label{eq:R2-score}
    R^2 = 1 - \frac{\sum_{i=1}^{n_{dof}} \Big(\tilde{\textbf{y}}^\star_i - \textbf{y}^\star_i\Big)^2}{\sum_{i=1}^{n_{dof}} \Big(\textbf{y}^\star_i - \overline{\textbf{y}}^\star\Big)^2},
\end{equation}
where $n_{dof} = m_f \times n_f$ is the size of the model solution and $\overline{\textbf{y}}^\star$ is the mean reference response. A summary of the steps of the proposed method is given in \ref{appendix_2}.

\section{Applications}
\label{examples}

In this section, we apply the proposed method to a diverse set of problems of increasing complexity. The first example aims to provide an intuitive understanding of the complex low-dimensional representations through a simple low-dimensional problem that can be interpreted graphically. The subsequent examples aim to demonstrate the breadth of high-dimensional problems that can be solved using the proposed approach. 

\subsection{Points on the hypersphere}
\label{subsec:sphere}
\begin{figure}
\centering
\begin{subfigure}[b]{0.30\textwidth}
 \centering
 \includegraphics[clip=true, trim=0cm 3cm 0cm 4cm, width=\textwidth]{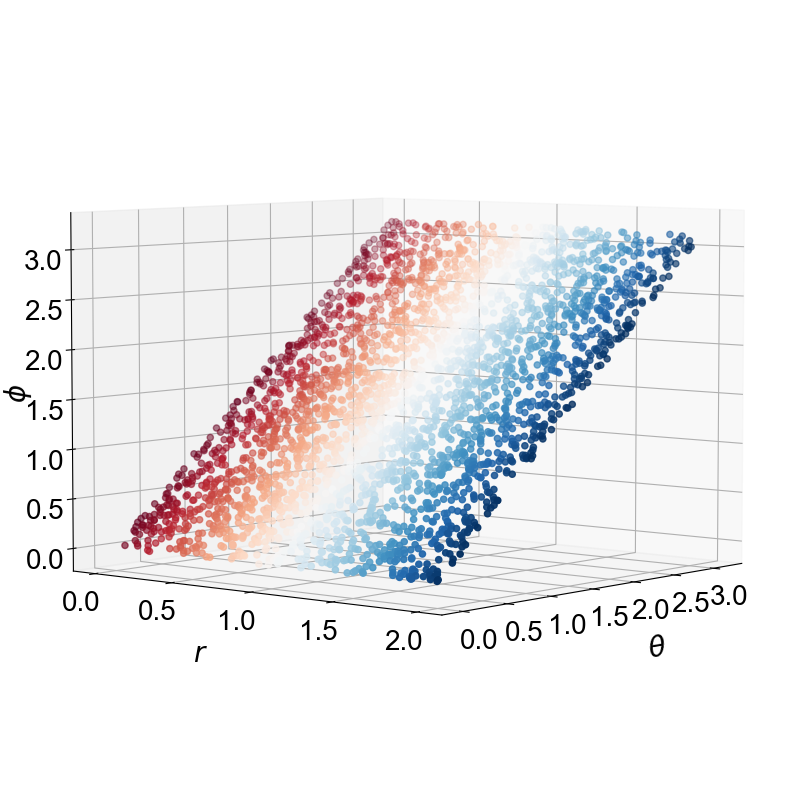}
 \caption{}
 \label{fig:frechet_Sphere_n50}
\end{subfigure}
\hfill
\begin{subfigure}[b]{0.32\textwidth}
 \centering
 \includegraphics[clip=true, trim=0.0cm 3cm 0cm 4cm, width=\textwidth]{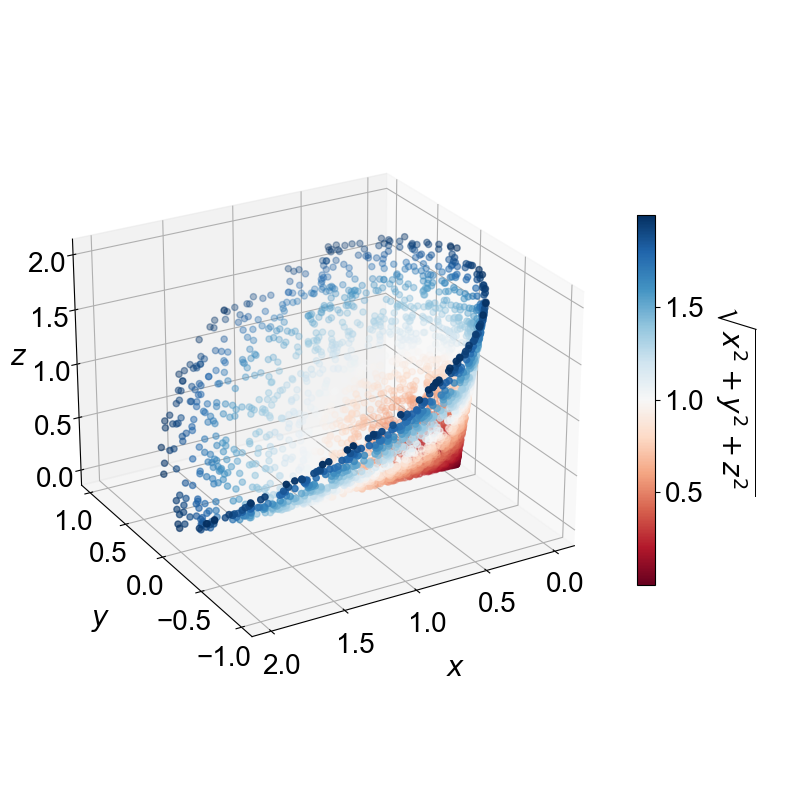}
 \caption{}
 \label{fig:frechet_Sphere_n100}
\end{subfigure}
\begin{subfigure}[b]{0.30\textwidth}
 \centering
 \includegraphics[clip=true, trim=2cm 5cm 2cm 4cm, width=\textwidth]{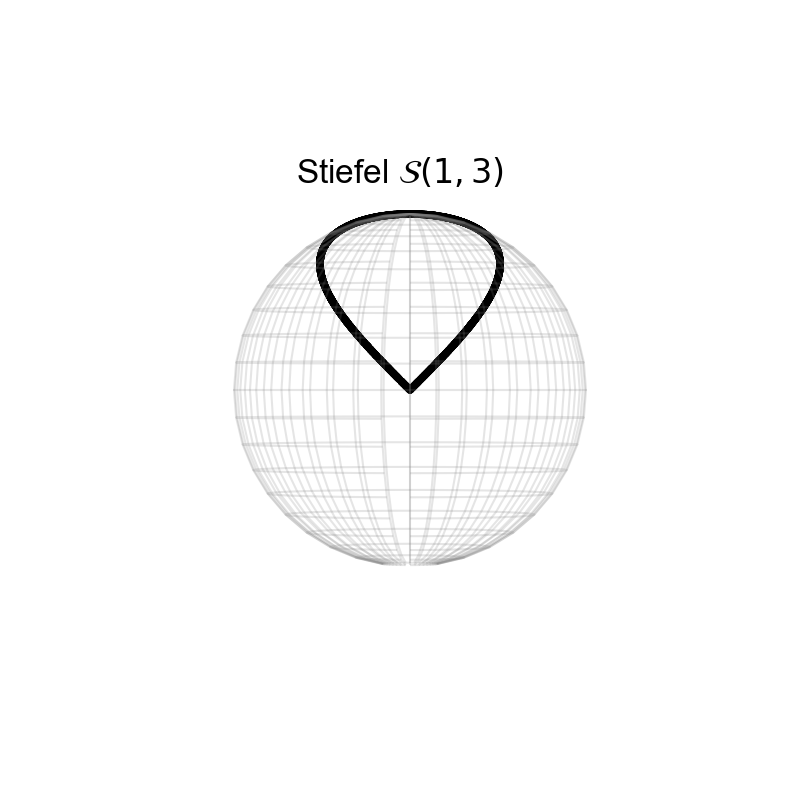}
 \caption{}
 \label{fig:frechet_Sphere_n150}
\end{subfigure}
\hfill 
\caption{Points on the hypersphere: (a) Realizations of $\boldsymbol{\theta}$. (b) Corresponding realizations of the responses $\mathbf{y}$. (c) Representation of the points on the Stiefel manifold $\mathbb{S}(1, 3)$}
\label{fig:sphere-data}
\end{figure}

In the first example, we use the proposed approach to learn the mapping between points ($\phi, \theta, r$) defined in polar coordinates and their Cartesian counterparts, that is
\begin{equation} 
\label{eq:cartesian-coordinates}
\begin{split}
    x & = r\cos(\phi)\cos(\theta) \\
    y & = r\cos(\phi)\sin(\theta) \\
    z & = r\sin(\phi) \\
\end{split}
\end{equation}
Vector $\boldsymbol{\theta} =[r, \theta, \phi]$ is considered to be stochastic with each variable following a uniform probability distribution within a specific range, such that
\begin{eqnarray}
\theta = U[0,  \pi], \quad
\phi = \theta, \quad
r = U[0, 2]
\end{eqnarray}
The response, i.e., the corresponding Cartesian point $\mathbf{y} = [x, y, z]^\intercal$, lies in the three-dimensional Euclidean space, $\mathbb{R}^3$.
The SVD $\mathbf{y}=\textbf{U} \boldsymbol{\Sigma} \textbf{V}^\intercal$ gives $\textbf{U}\in \mathbb{S}(1, 3)$ which is a point in three-dimensional space that lies on the surface of a unit sphere (Stiefel manifold $\mathcal{S}t(1, 3)$), while $\boldsymbol{\Sigma}, \textbf{V} \in \mathbb{R}$. Figure \ref{fig:sphere-data}(a-c) shows 3000 realizations of the parameters vector $\boldsymbol{\theta} =[r, \theta, \phi]$, the corresponding realizations of $\mathbf{y}$, and their representations $\textbf{U}$ on the Stiefel manifold $\mathcal{S}t(1, 3)$, respectively.

%\begin{figure}
%\centering
%\begin{subfigure}[b]{0.49\textwidth}
% \centering
% \includegraphics[width=\textwidth]{Figures/Results/Sphere/Sphere_FrechetVarianceConvergence_Ntrain_.png}
% \caption{.}
% \label{fig:frechet_Sphere_n}
%\end{subfigure}
%\hfill
%\begin{subfigure}[b]{0.49\textwidth}
% \centering
% \includegraphics[width=\textwidth]{Figures/Results/Sphere/Sphere_FrechetVarianceConvergence_Ntrain_200.png}
% \caption{}
% \label{fig:frechet_Sphere_n200}
%\end{subfigure}
%\caption{(a). Convergence of the minimum mean Fr\'{e}chet variance for an increasing number of clusters and for training data sets of increasing size $\mathcal{N}$. (b) Convergence of the minimum mean Fr\'{e}chet variance (solid blue line) and the corresponding 96\% confidence interval (light blue area) for $\mathcal{N}=200$.}
%\label{fig:frechet_sphere}
%\end{figure}

\begin{figure}
\centering
\includegraphics[width=0.6\textwidth]{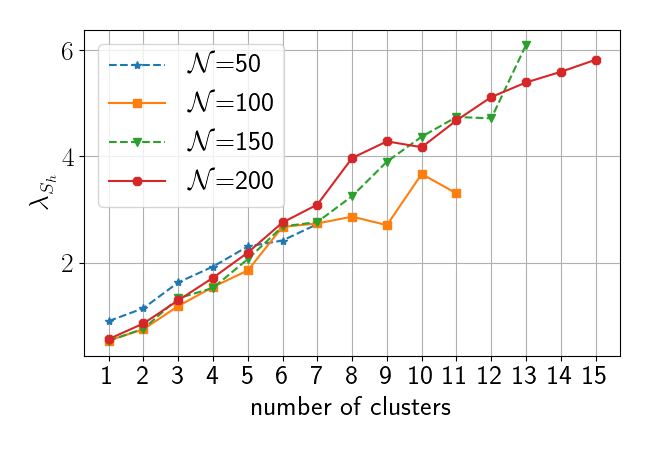}
\caption{Convergence of the Silhouette coefficient for an increasing number of clusters and for training data sets of increasing size $\mathcal{N}$. The optimum number of clusters is the one corresponding to the maximum peak of the each plot.}
\label{fig:frechet_Sphere_n}
\end{figure}

\begin{figure}
\centering
\begin{subfigure}[b]{0.34\textwidth}
 \centering
 \includegraphics[clip=true, trim=6cm 6cm 6cm 6cm, width=\textwidth]{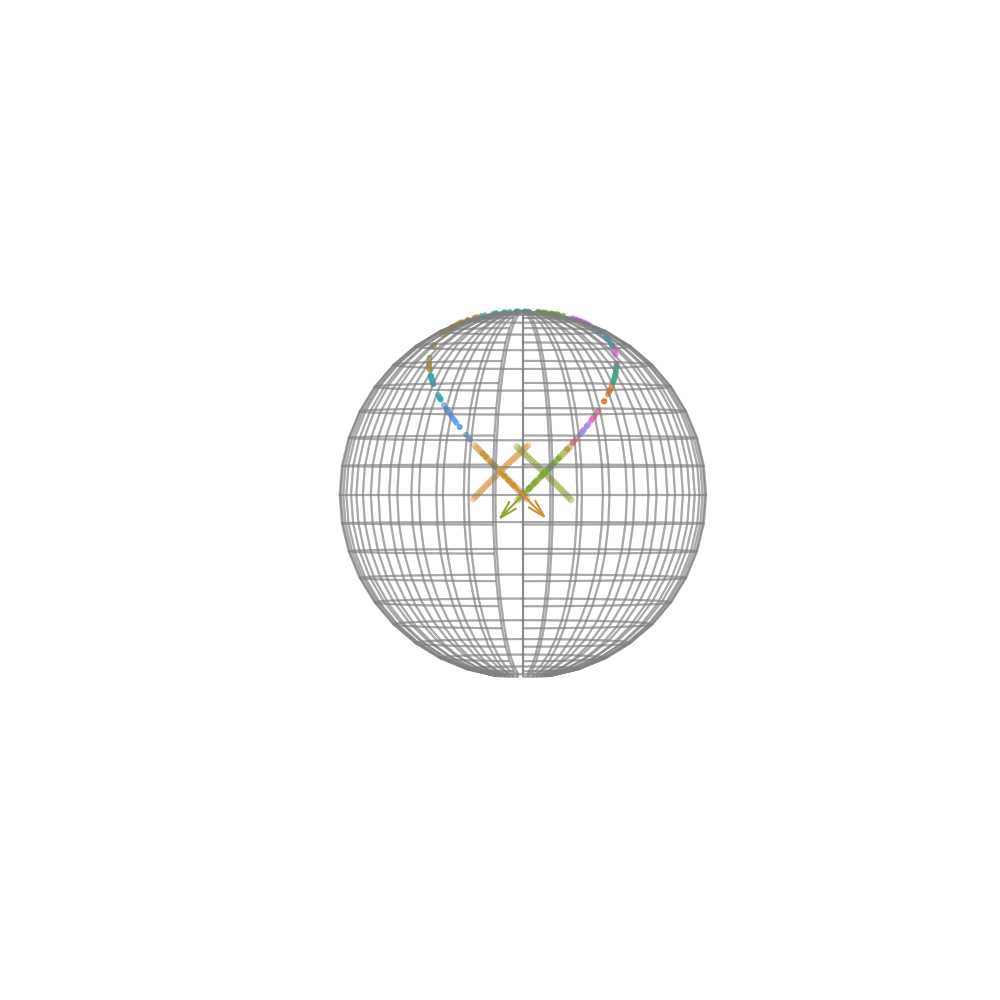}
 \caption{}
 \label{fig:frechet_Sphere_n50}
\end{subfigure}
\hfill
\begin{subfigure}[b]{0.31\textwidth}
 \centering
 \includegraphics[clip=true, trim=5cm 5cm 5cm 5cm, width=\textwidth]{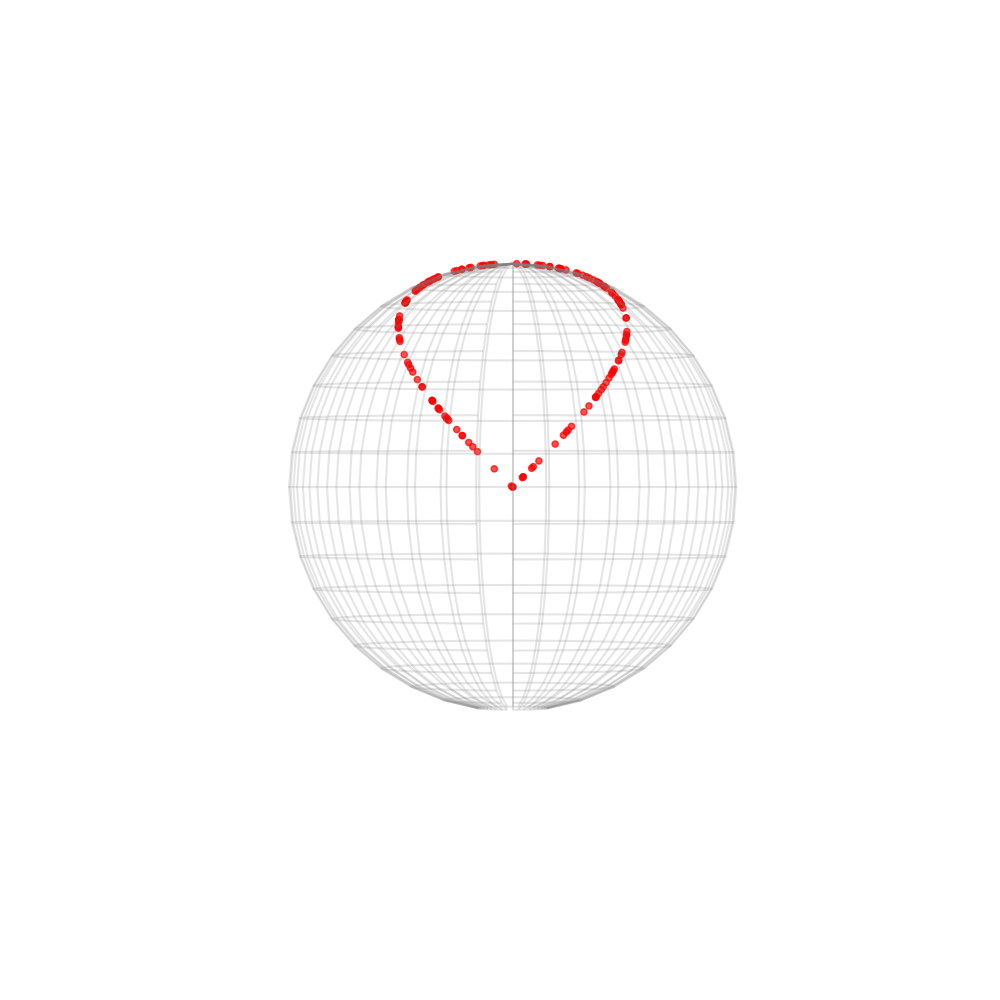}
 \caption{}
 \label{fig:frechet_Sphere_n100}
\end{subfigure}
\begin{subfigure}[b]{0.31\textwidth}
 \centering
 \includegraphics[clip=true, trim=0cm 1cm 0cm 1cm, width=\textwidth]{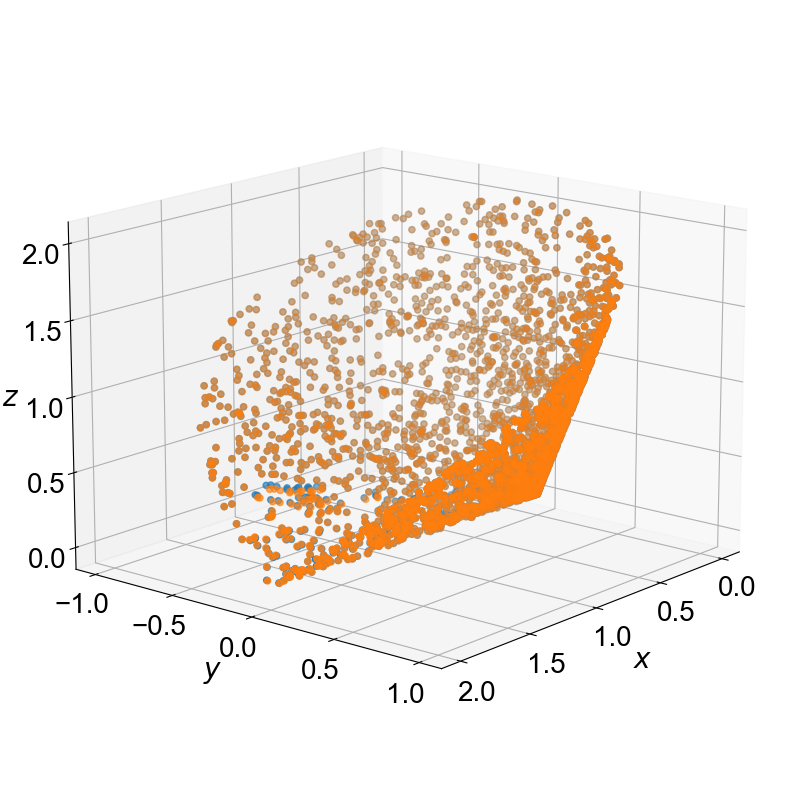}
 \caption{}
 \label{fig:frechet_Sphere_n150}
\end{subfigure}
\hfill 
\caption{Points on the hypersphere: (a) Cluster of points on $\mathbb{S}(1, 3)$; for two clusters we plot the geodesics on the sphere corresponding to the first  and second principal components (b) 150 PCE predicted points on $\mathbb{S}(1, 3)$, (c) and the corresponding solutions in Cartesian space. }
\label{fig:sphere-vectors}
\end{figure}

We consider training data sets consisting of $\mathscr{N}=50, 100, 150$, and $200$ training points, i.e., pairs $\{\boldsymbol{\theta}_i, \mathbf{y}_i\}_{i=1}^{\mathscr{N}}$. 
To evaluate the performance of the surrogate, we generate a test data set of size $\mathscr{N}_*=3000$.
We perform Riemannian K-means to identify groups of points $\{\textbf{U}_i\}_{i=1}^{N_h}
$ that are close on the Stiefel manifold 
$\mathcal{S}t(1, 3)$. 
Figures \ref{fig:frechet_Sphere_n} shows the convergence of the Silhouette coefficient as a function of the number of clusters for increasing number of training points. The optimum number of clusters is the one corresponding to the maximum peak of the each plot.  A set requirement is that each cluster contains at least five points. 
In Figure \ref{fig:sphere-vectors}(a) we show $\mathcal{N}=200$ train points on the sphere divided into 15 clusters, where each color represents a cluster. For two of the clusters we show the tangent vectors that align  with the first and second principal geodesic components. %Figure \ref{fig:sphere-vectors}(b) then 
% isolates one of the clusters identified with the proposed Fr\'{e}chet variance-based approach. As evidence that the clustering on the sphere was successful, Figure~\ref{fig:sphere-vectors}(b) shows that, as expected, within a cluster 
%shows the tangent vectors for each point in one cluster showing a dominant principal direction.
% on the tangent space defined at the Karcher mean are  pointing towards the same direction (i.e., move along the same geodesic).
 %shown in Figure~\ref{fig:sphere-vectors}(c) with smaller variation along the second principal geodesic submanifold.
% depicts the geodesics corresponding to the first two principal components.

%\begin{figure}[t!]
%\begin{center}
%\includegraphics[width=1.0\textwidth]{PGA-PCE/Figures/Results/Sphere/Sphere_vectors_new.png}
%\caption{Points on the hypersphere: (a) Cluster of points on $\mathbb{S}(1, 3)$. (b) Tangent vectors with the same direction. (c) Geodesics on the sphere corresponding to the first (magenta) and second (green) principal components..}
%\label{fig:sphere-vectors}
%\end{center}
%\end{figure}

%\begin{figure}[t!]
%\begin{center}
%\includegraphics[width=0.8\textwidth]{PGA-PCE/Figures/Results/Sphere/Sphere_pga.png}
%\caption{(a) Cluster points (blue marker) along with their corresponding Fr\'{e}chet mean. (b) Geodesics on the sphere corresponding to the first (magenta) and second (green) principal components.}
%\label{fig:sphere-pga}
%\end{center}
%\end{figure}

PCE surrogates that map from $\mathbf{x}$ to $\mathbf{U}$ are then constructed, with a maximum polynomial degree $p_{\text{max}}$ that varies from 1 to 3. Figures \ref{fig:sphere-vectors}(b-c) visualizes the performance of the PCE surrogate for $p_{\max}=2$ and $\mathcal{N}=200$ training points: More specifically, Fig.~\ref{fig:sphere-vectors}(b) depicts 150 predicted values on the sphere $\mathcal{S}t(1, 3)$ while Fig.~\ref{fig:sphere-vectors}(c) shows the predicted full solutions in the Cartesian space. We see that no discrepancies can be observed between the surrogate prediction and the exact solution.
% , which are used to predict the QoI. 
%Figures \ref{fig:sphere-prediction} visualizes the performance of the PCE surrogate for $p_{\max}=2$ and $\mathcal{N}=100$ training points: More specifically, Fig.~\ref{fig:sphere-prediction}(a) depicts the predicted values on the sphere $\mathcal{S}t(1, 3)$. We see that the surrogate is able to predict points that are constrained along the geodesic. Figure~\ref{fig:sphere-prediction}(b) shows the corresponding reconstructed solutions in the Cartesian space where no discrepancies can be observed. 
This performance of the surrogate is reflected on very low $L_2$ errors and high $R^2$ scores for $p_{\max}=2$ and $p_{\max}=3$, even for the smallest training data sets.  As would be expected, the approximation accuracy in general increases if larger training data sets are employed. This can be attributed to an overfitted PCE and shows clearly that $p_{\max}$ must be chosen in accordance to the available training data.

\subsection{Lotka-Volterra Dynamical System}
\label{subsec:LV}
In this example, we consider the Lotka-Volterra system of equations \cite{mao2003asymptotic}, a system of first-order non-linear ordinary differential equations (ODEs) used to model the interactions between predators and prey. The model is defined as 
\begin{equation} 
\label{eq:lotka-volterra}
\begin{split}
    \frac{du}{dt} & = \alpha u - \beta uv, \\
    \frac{dv}{dt} & = \delta uv - \gamma v,
\end{split}
\end{equation}
where $u$ and $v$ are the prey and predator populations, respectively. The model parameters $\alpha$, $\beta$, $\gamma$, and $\delta$ are described in Table \ref{table:lotka}. In our setting, we consider the parameters $\alpha$ and $\beta$ to be stochastic with values sampled from uniform probability distributions that lie between certain bounds, while parameters $\gamma$ and $\delta$ have fixed values. The initial conditions for the state variables are set to $u_0=10$ prey and $v_0=5$ predators.

\begin{table}[!ht]
\small
\caption{Description of the parameters of the Lotka-Volterra equations.}
\vspace{-7pt}
\centering
\begin{tabular}{l c c l}
\toprule
Parameters & \hspace{15pt} & Notation & \hspace{2pt} Uncertainty/value \\ [0.5ex]
\toprule
Population of prey species   &  &  $u$  &    $u_0=10$  \vspace{2pt}  \\
Population of predator species  &   & $v$ & $v_0=5$ \vspace{2pt}  \\ 
Natural growing rate of prey when no predator exists   &   &   $\alpha$ & $\it{U}(0.90, 1)$ \vspace{2pt} \\
Natural dying rate of prey due to predation   &   &  $\beta$ &    $\it{U}(0.10, 0.15)$  \vspace{2pt}  \\  
Natural dying rate of predator when no prey exists   &   &  $\gamma$ &    $1.50$  \vspace{2pt} \\ 
Reproduction rate of predators per prey eaten  &   &  $\delta$ &    $0.75$  \vspace{1pt}  \\ 
\bottomrule
\end{tabular}
\label{table:lotka}
\end{table}

Our goal is to predict the trajectory of both predator and prey species over time using the PGA-PCE surrogate. For each realization $\boldsymbol{\theta}_i$ of the vector $[\alpha, \beta]\in \mathbb{R}^{2}$, we solve the Lotka-Volterra system using a fourth-order Runge-Kutta method with period $T=25$, discretized in $512$ points. Therefore, the response of the system $\bf{y}_i=[\bf{u}_i, \bf{v}_i]$ is a point lying in $\mathbb{R}^{2\times 512}$. The objective is to learn the mapping  $\boldsymbol{\theta} \rightarrow \bf{y}$ using a small number of training data $\{\boldsymbol{\theta}_i, \bf{y}_i\}$. Each response is factorized into three  matrices $\bf{y}_i \rightarrow U_i, \Sigma_i,V_i^\intercal$ using thin SVD. Matrices $U_i$ and $V_i$ correspond to points on the Stiefel manifolds $\mathcal{S}t_U(n_1, p) \equiv\mathcal{S}t_U(2, 512)$ and $\mathcal{S}t_V(n_2, p)\equiv \mathcal{S}t(2, 2)$, respectively. Matrix $\Sigma_i$ is a vector in the Euclidean space $\mathbb{R}^{p}\equiv \mathbb{R}^{2}$.

\begin{figure}[!ht]
\centering
\includegraphics[width=0.6\textwidth]{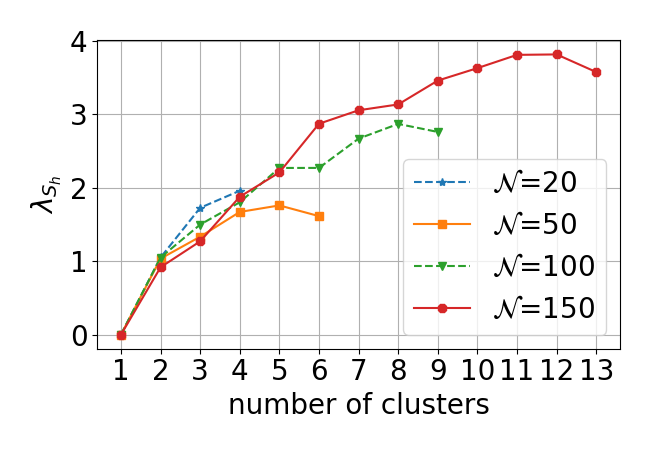}
\caption{Convergence of the Silhouette coefficient for an increasing number of clusters and for training data sets of increasing size $\mathcal{N}$. The optimum number of clusters is the one corresponding to the maximum peak of the each plot.}
\label{fig:frechet_lv}
\end{figure}

% Frechet variances figure
%\begin{figure}
%\centering
%\begin{subfigure}[b]{0.49\textwidth}
% \centering
% \includegraphics[width=\textwidth]{Figures/Results/Lotka-Volterra/LV_FrechetVarianceConvergence_Ntrain_20.png}
% \caption{$\mathcal{N}=20$.}
% \label{fig:frechet_lv_n20}
%\end{subfigure}
%\hfill
%\begin{subfigure}[b]{0.49\textwidth}
% \centering
% \includegraphics[width=\textwidth]{Figures/Results/Lotka-Volterra/LV_FrechetVarianceConvergence_Ntrain_50.png}
% \caption{$\mathcal{N}=50$.}
% \label{fig:frechet_lv_n50}
%\end{subfigure}
%\\
%\begin{subfigure}[b]{0.49\textwidth}
% \centering
% \includegraphics[width=\textwidth]{Figures/Results/Lotka-Volterra/LV_FrechetVarianceConvergence_Ntrain_100.png}
% \caption{$\mathcal{N}=100$.}
% \label{fig:frechet_lv_n100}
%\end{subfigure}
%\hfill 
%\begin{subfigure}[b]{0.49\textwidth}
% \centering
%\includegraphics[width=\textwidth]{Figures/Results/Lotka-Volterra/LV_FrechetVarianceConvergence_Ntrain_150.png}
% \caption{$\mathcal{N}=150$.}
% \label{fig:frechet_lv_n150}
%\end{subfigure}
%\caption{Lotka-Volterra: Convergence of the minimum mean Fr\'{e}chet variance (solid blue line) and the corresponding 96\% confidence interval (light blue area) for an increasing number of clusters and for training data sets of increasing size $\mathcal{N}$.}
%\label{fig:frechet_lv}
%\end{figure}

Figure~\ref{fig:frechet_lv} depicts the convergence of the Silhouette coefficient for an increasing number of clusters and for training data sets of increasing size $\mathcal{N}$. The optimum number of clusters is the one corresponding to the maximum peak of the each plot., i.e.,  $\mathcal{N} \in \left\{20,50,100,150\right\}$.
For each training data set size $\mathcal{N}$, the number of clusters increases progressively, up to the point where the optimal number of clusters is identified, as described in Algorithm~\ref{alg:cap}. For $\mathcal{N}=100$, the optimal number of clusters is 12. In this example, the set requirement is that all clusters contain at least 5 data points.
%The results are omitted for $\mathcal{N}=10$, since the Riemannian K-means results in a partition of the $\{U_i\}_{i=1}^{\mathcal{N}}$ data on the Stiefel manifold into two clusters, as expected.  
In each cluster, PGA  provides a set of principal components -the optimal number of principal components explains $99\%$ of the  variance in the data. As expected, different number of principal components is required to achieve a uniformly explained variance across all clusters. For $\mathcal{N}=150$, the required number of principal components identified in each cluster is 1. 

\begin{figure}
 \centering
 \includegraphics[width=0.6\textwidth]{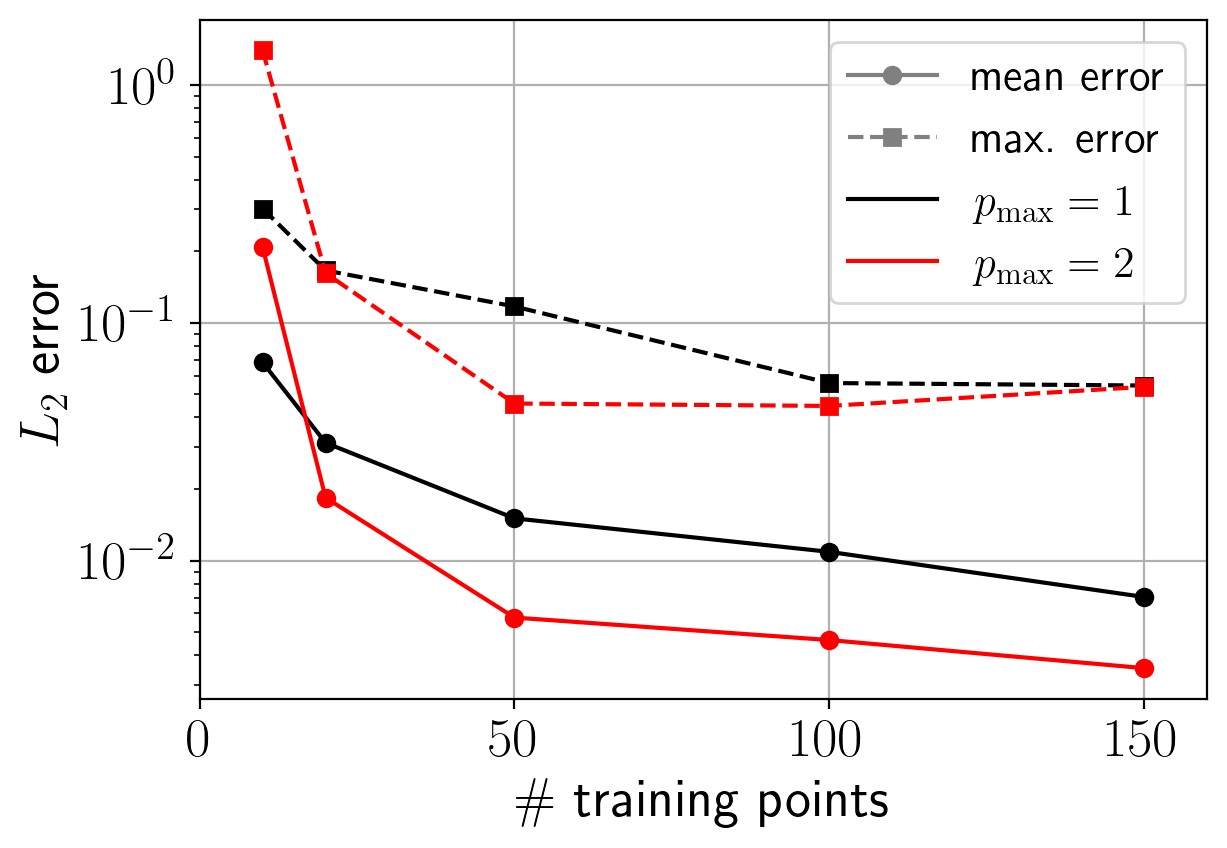}
 \caption{$\mathscr{Y}$ prediction.}
 \label{fig:l2y_lv}
\caption{Lotka-Volterra: $L_2$ errors of the surrogate model's predictions for the QoI $\mathscr{Y}$, for training data sets of increasing size. The $L_2$ errors have been computed using a validation data set with $\mathcal{N}_*=5000$ data points.}
\label{fig:l2_lv}
\end{figure}

Next, a PCE is constructed for each cluster, %where the number of clusters depends on the size of the training data set. 
and the maximum polynomial degree is  set $p_{\text{max}}=1$ (linear regression) or $p_{\text{max}}=2$ across the clusters. By identifying the optimal number of clusters in a way that minimizes the variance within each cluster, in the case of small data, a small polynomial degree (i.e., $p_{\text{max}}\leq 3$) will be sufficient. Figure \ref{fig:l2_lv} depicts the performance of the surrogate models in terms of the average and maximum $L_2$ errors, where the latter are computed using a validation data set of size $\mathcal{N}_*=5000$.
%A similar performance comparison is presented in Table~\ref{tab:r2_lv}, this time concerning the average and minimum $R^2$ score of the different surrogate models.
More specifically, the surrogate's predictions of the 
%manifolds $\mathbb{S}_U(n_1, p)$ and $\mathbb{S}_V(n_2, p)$, the Euclidean space $\mathbb{R}^2$, and the 
solutions $\bf{\tilde{y}}= \tilde{U}\hat{\Sigma}\tilde{V}^\intercal$ are compared to the true ones, for training data sets of increasing size. 
It can be observed that even with 50 training data, the performance of the surrogate can already be considered very good, while additional training data improve the surrogate's accuracy further. 
%Additionally, it can be seen that PCEs with $p_{\max}=1$ outperform PCEs with $p_{\max}=2$ for the smaller training data sets, i.e., for $\mathcal{N}=10$ and $\mathcal{N}=20$. However, the total degree $p_{\max}=2$ results in significant accuracy gains for training data sets of size $\mathcal{N}=50$ or larger.

\begin{figure}[t!]
\centering
\begin{subfigure}[b]{0.49\textwidth}
 \centering
 \includegraphics[width=\textwidth]{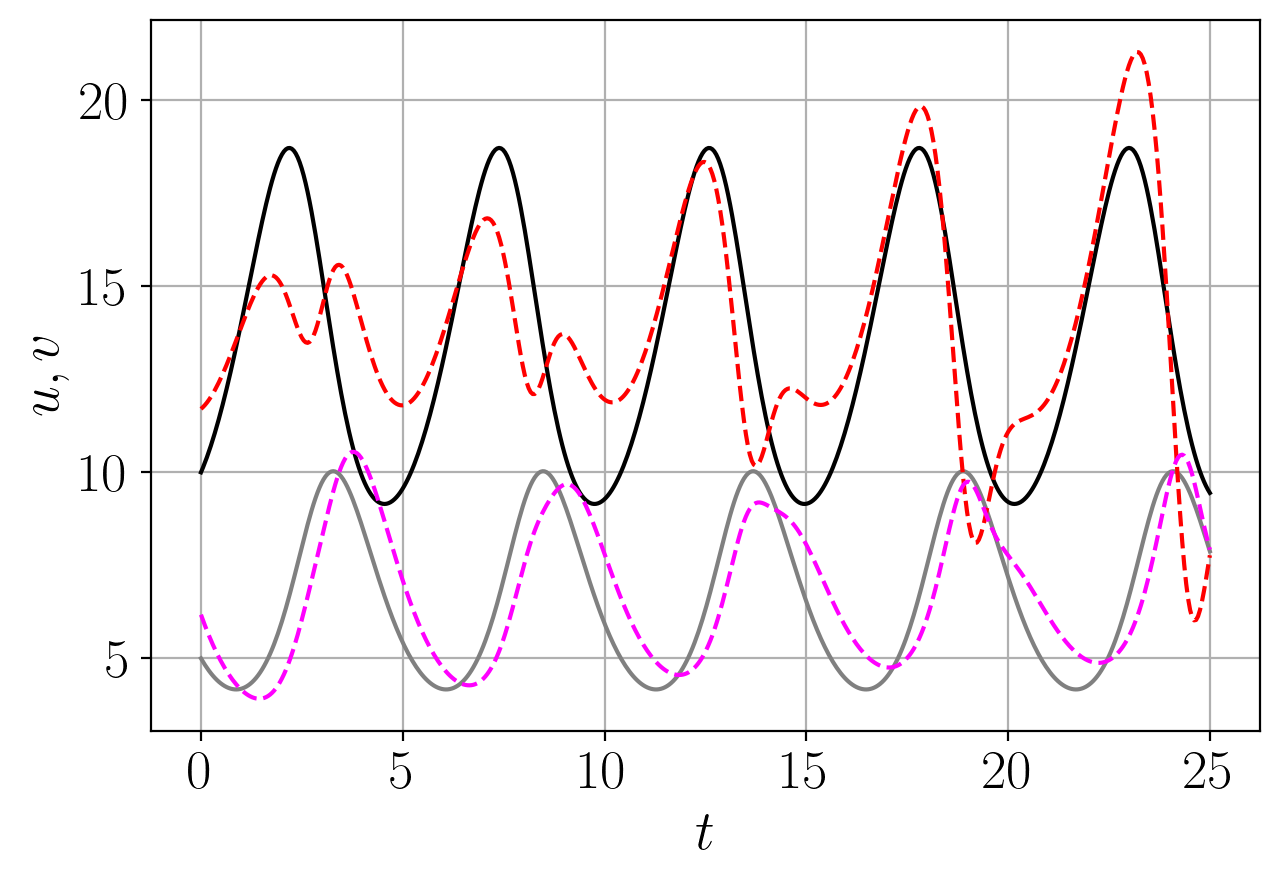}
 \caption{$\mathcal{N}=20$}
 \label{fig:worst_case_n10_p1}
\end{subfigure}
\hfill
\begin{subfigure}[b]{0.49\textwidth}
 \centering
 \includegraphics[width=\textwidth]{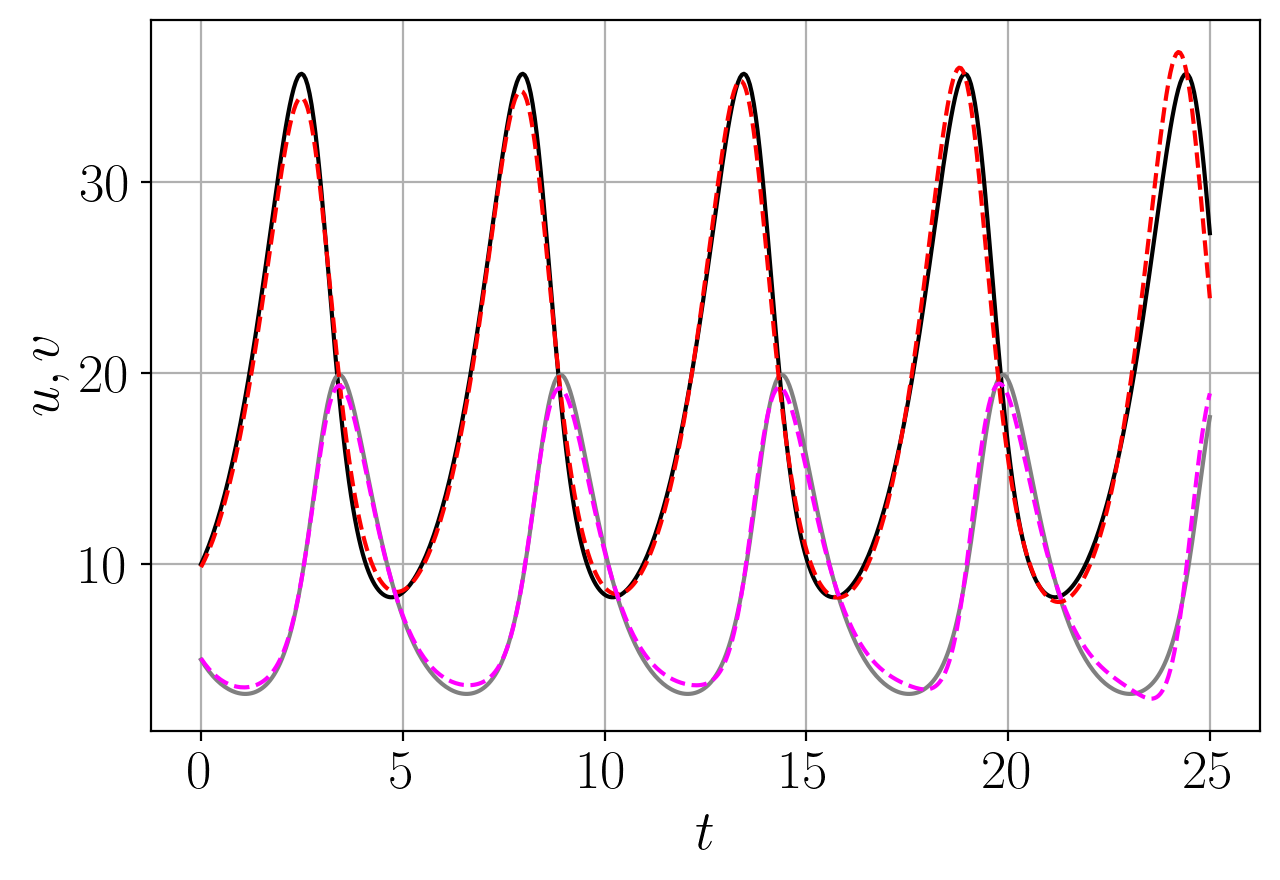}
 \caption{$\mathcal{N}=50$}
 \label{fig:worst_case_n10_p2}
\end{subfigure}
%\\
\begin{subfigure}[b]{0.49\textwidth}
 \centering
 \includegraphics[width=\textwidth]{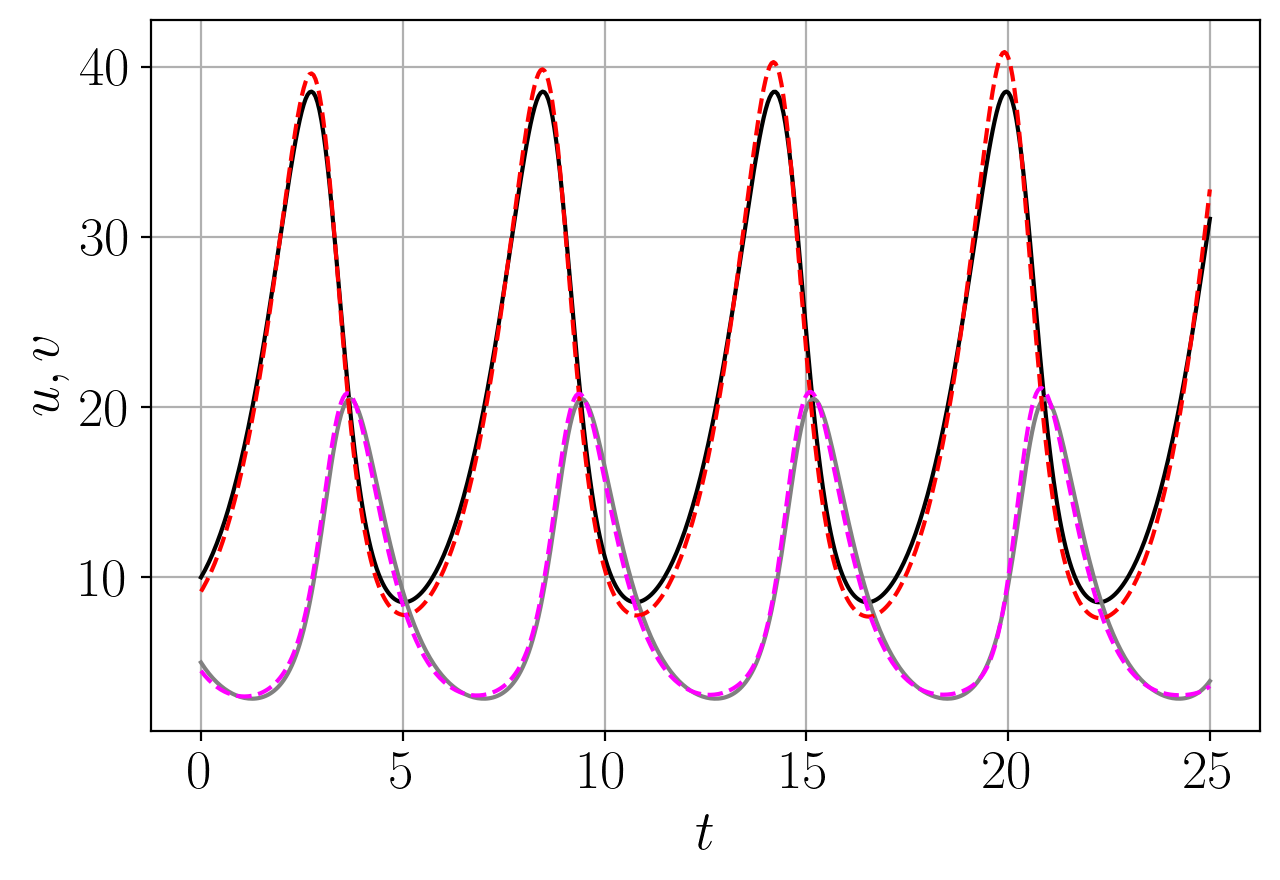}
 \caption{$\mathcal{N}=100$}
 \label{fig:worst_case_n20_p1}
\end{subfigure}
\hfill
\begin{subfigure}[b]{0.49\textwidth}
 \centering
 \includegraphics[width=\textwidth]{Figures/Results/Lotka-Volterra/worst_trajectory_Ntrain100p2.png}
 \caption{$\mathcal{N}=150$}
 \label{fig:worst_case_n20_p2}
\end{subfigure}
%\\
%\begin{subfigure}[b]{0.49\textwidth}
% \centering
% \includegraphics[width=\textwidth]{Figures/Results/Lotka-Volterra/worst_trajectory_Ntrain50p1.png}
% \caption{$\mathcal{N}=50$, $p_{\max}=1$.}
% \label{fig:worst_case_n50_p1}
%\end{subfigure}
%\hfill
%\begin{subfigure}[b]{0.49\textwidth}
% \centering
% \includegraphics[width=\textwidth]{Figures/Results/Lotka-Volterra/worst_trajectory_Ntrain50p2.png}
% \caption{$\mathcal{N}=50$, $p_{\max}=2$.}
% \label{fig:worst_case_n50_p2}
%\end{subfigure}
\caption{Lotka-Volterra: Worst-case surrogate model predictions for the trajectories of the prey ($u$) and predator ($v$) species for increasing number of training samples and $p_{\text{max}}=2$. The true trajectories are given with solid lines ($u$: black, $v$: gray), while the surrogate's predictions are given with dashed lines ($u$: red, $v$: magenta).}
\label{fig:worst_case_trajectories_lv}
\end{figure}

As further verification with respect to the accuracy of the surrogate model, particularly considering data sets of limited size, Figure~\ref{fig:worst_case_trajectories_lv} presents the worst-case predictions of the surrogate model regarding the evolution of the prey and predator species. 
The worst-case approximations correspond to the maximum $L_2$ error computed out of the $\mathcal{N}_*=5000$ validation data. It is obvious that $\mathcal{N}=20$ training data points are insufficient for an acceptable worst-case prediction, which however is not surprising. 
A significant improvement is observed already for $\mathcal{N}=50$ training data where even the worst-case prediction of the surrogate matches the true trajectories very closely.%, where the PCEs with $p_{\max}=2$ are comparatively more accurate, as also previously shown in Figure~\ref{fig:l2_lv}.

\begin{figure}[t!]
\centering
\begin{subfigure}[b]{0.49\textwidth}
 \centering
 \includegraphics[width=\textwidth]{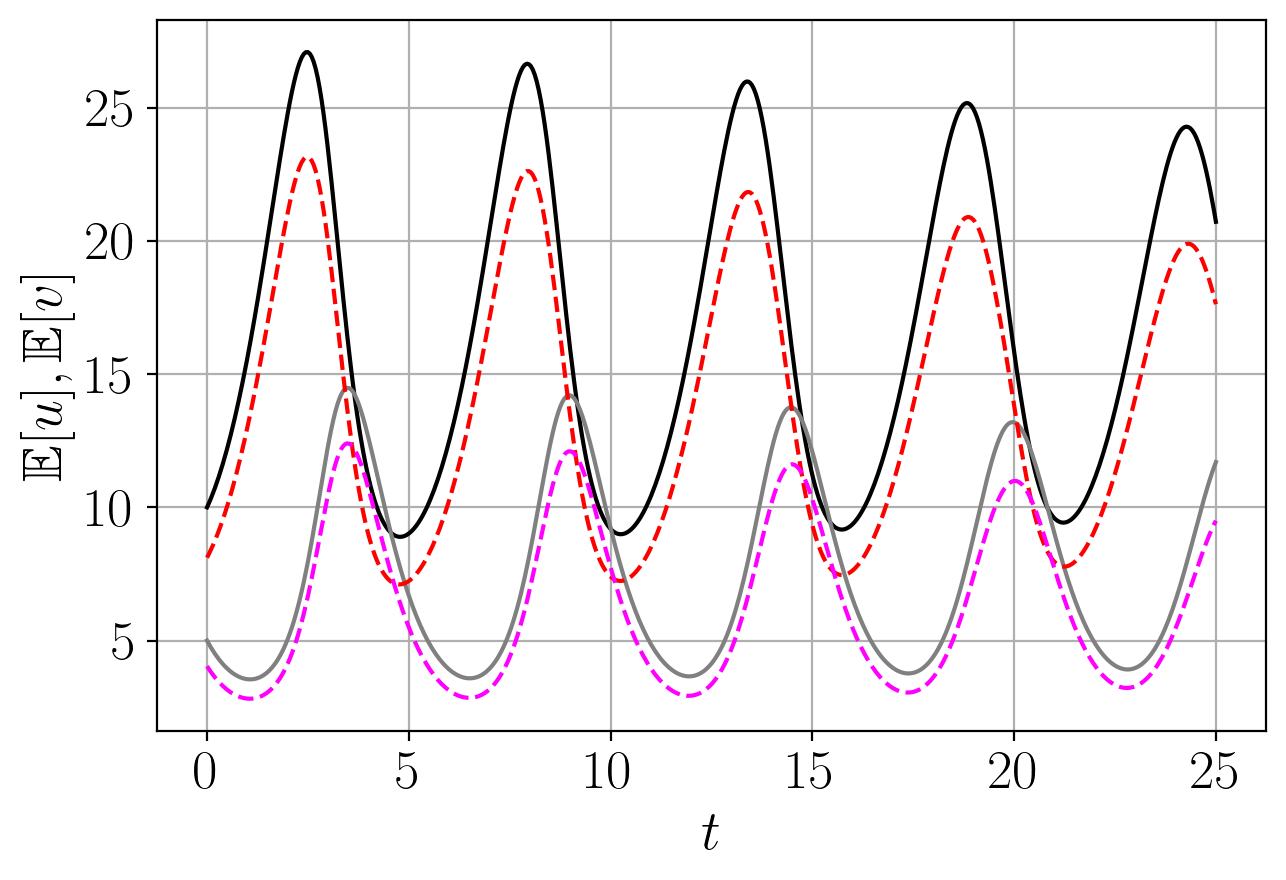}
 \caption{Mean, $\mathcal{N}=20$.}
 \label{fig:mean_n10_p2}
\end{subfigure}
\hfill
\begin{subfigure}[b]{0.49\textwidth}
 \centering
 \includegraphics[width=\textwidth]{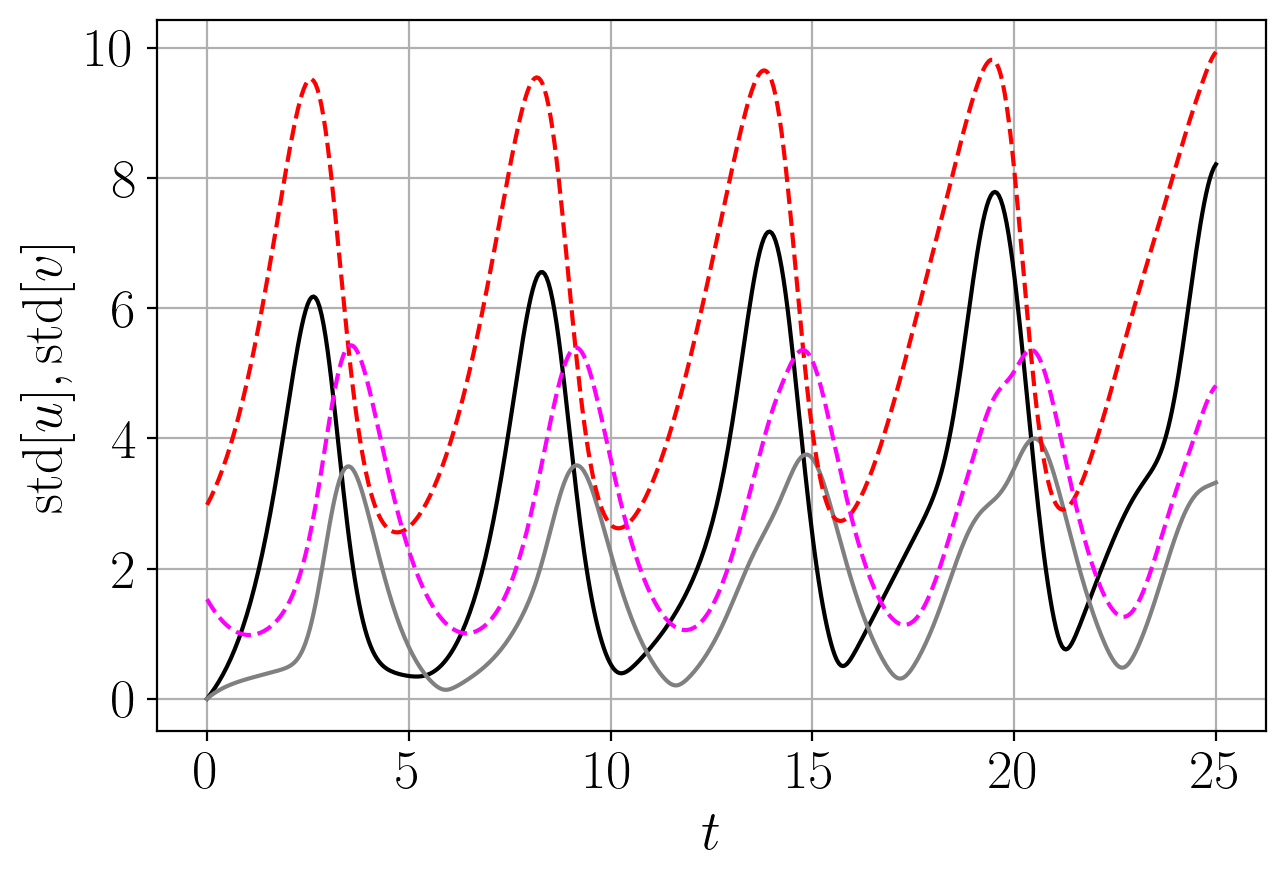}
 \caption{Standard deviation, $\mathcal{N}=20$.}
 \label{fig:std_n10_p2}
\end{subfigure}
\\
\begin{subfigure}[b]{0.49\textwidth}
 \centering
 \includegraphics[width=\textwidth]{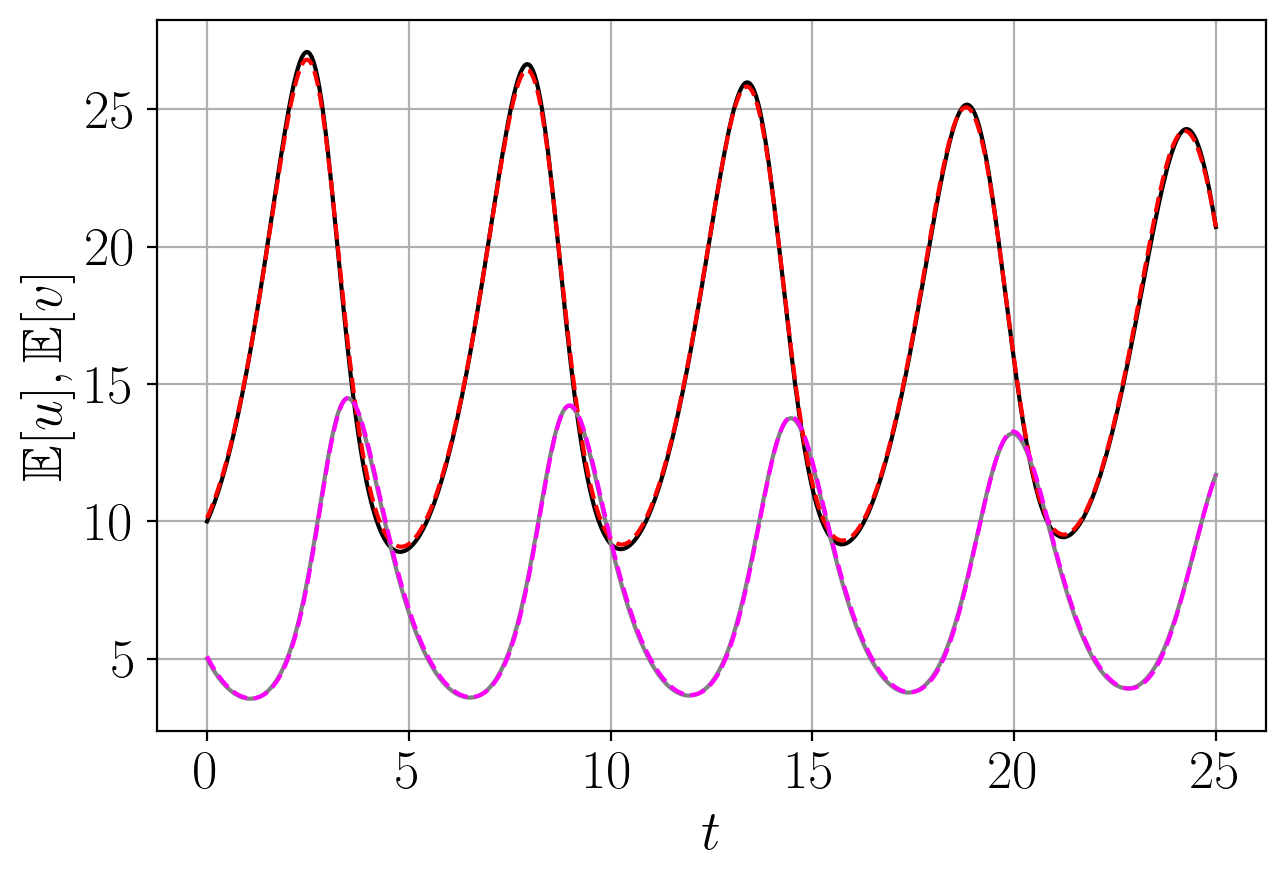}
 \caption{Mean, $\mathcal{N}=50$.}
 \label{fig:mean_n20_p2}
\end{subfigure}
\hfill
\begin{subfigure}[b]{0.49\textwidth}
 \centering
 \includegraphics[width=\textwidth]{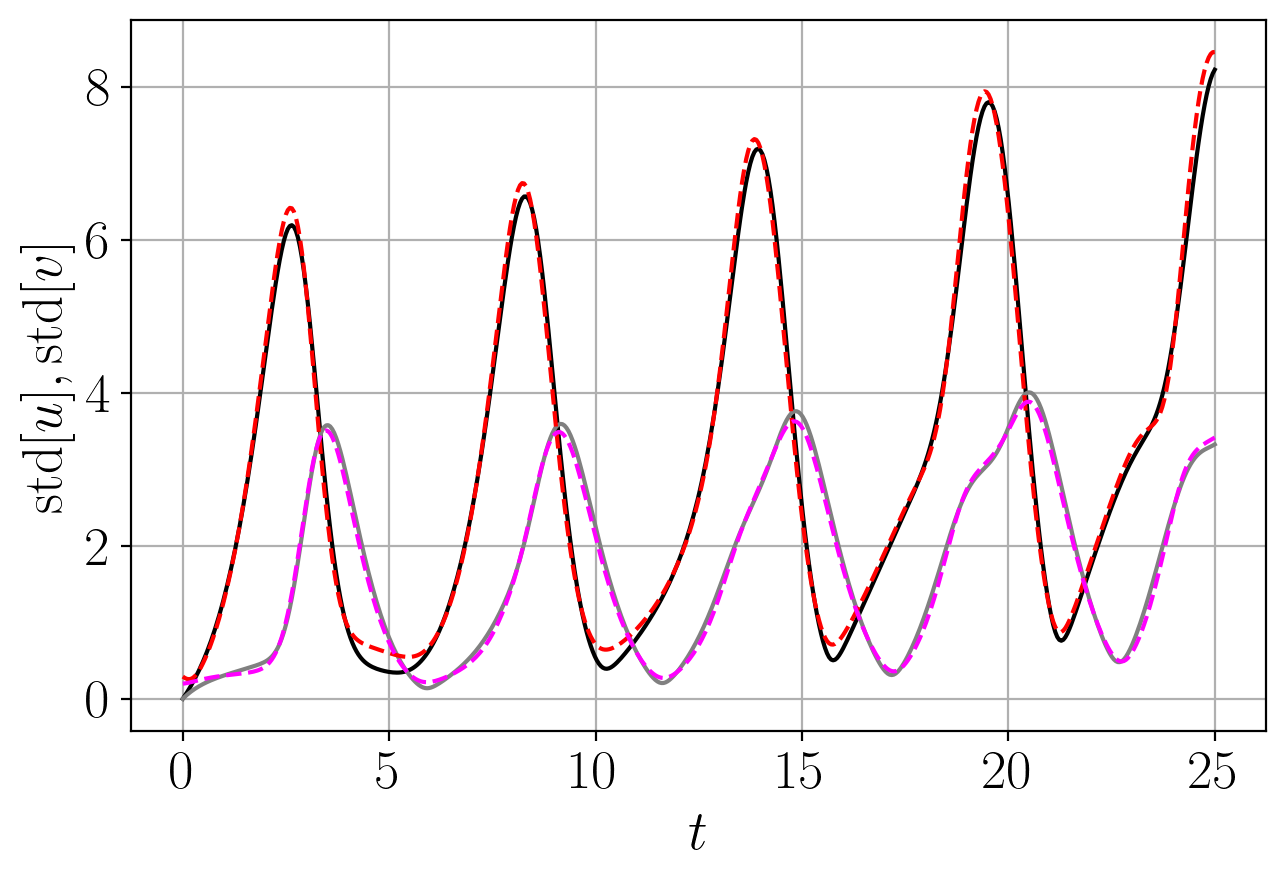}
 \caption{Standard deviation, $\mathcal{N}=50$.}
 \label{fig:std_n20_p2}
\end{subfigure}
\\
\begin{subfigure}[b]{0.49\textwidth}
 \centering
 \includegraphics[width=\textwidth]{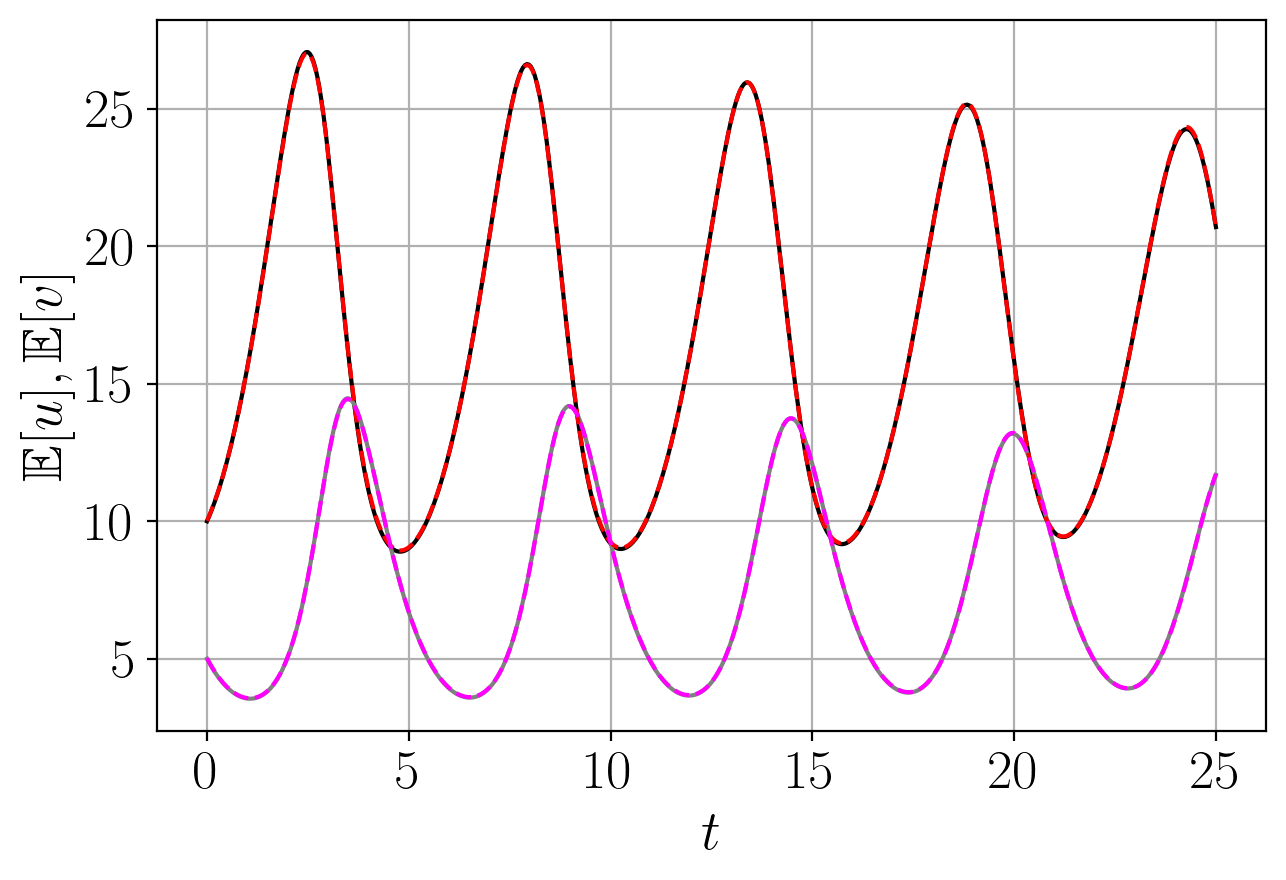}
 \caption{Mean, $\mathcal{N}=150$.}
 \label{fig:mean_n20_p2}
\end{subfigure}
\hfill
\begin{subfigure}[b]{0.49\textwidth}
 \centering
 \includegraphics[width=1.05\textwidth]{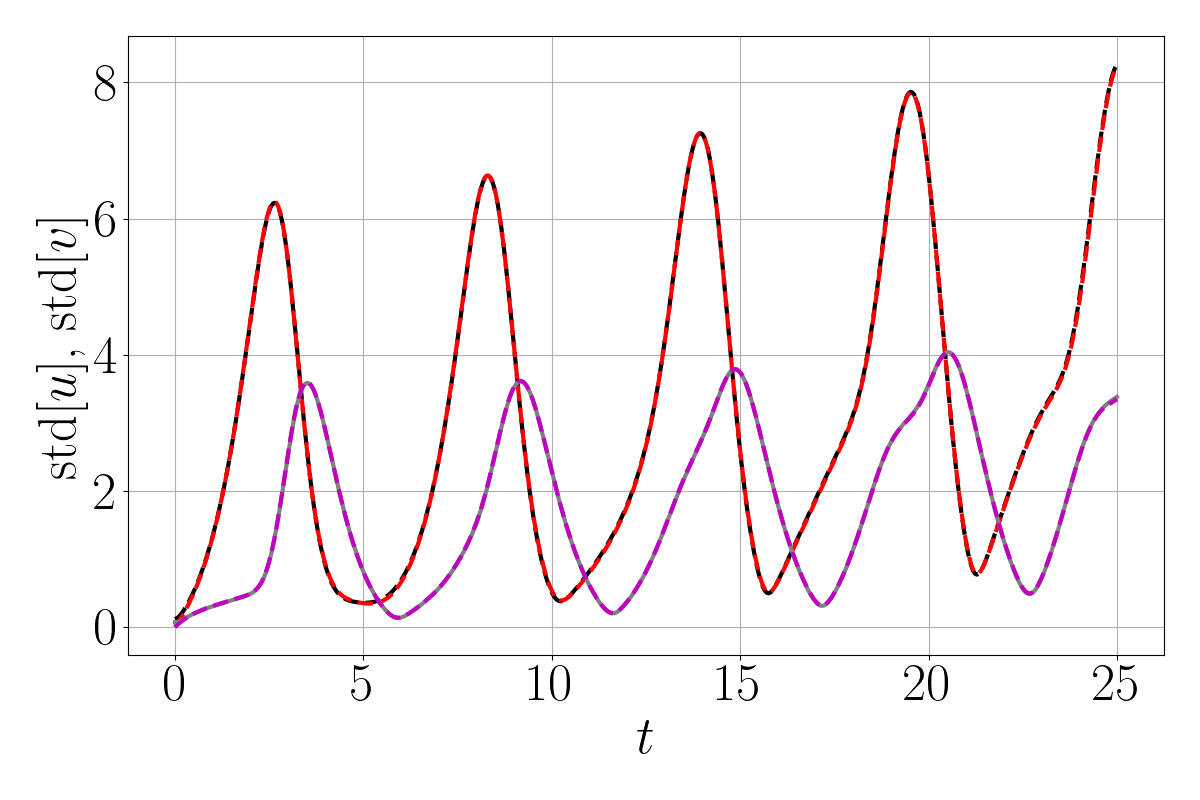}
 \caption{Standard deviation, $\mathcal{N}=150$.}
 \label{fig:std_n20_p2}
\end{subfigure}
\caption{Lotka-Volterra: Moment estimation for the trajectories of the prey ($u$) and predator ($v$) species. The means and standard deviations computed with the true model are given with solid lines ($u$: black, $v$: gray), while the surrogate-based estimations (for $p_{\max}=2$) are given with dashed lines ($u$: red, $v$: magenta).}
\label{fig:uq_lv}
\end{figure}

Last, using the same $N_*=5000$ test realizations, we perform moment estimation via Monte Carlso sampling. 
Figure~\ref{fig:uq_lv} shows the mean and the standard deviation of both predator and prey species over time, computed using both the original model and the proposed surrogate.  
In the latter case, we choose for $p_{\max}=2$.
Looking at the mean trajectories and the corresponding standard deviations, it can be seen that the surrogate-based moment estimation is very accurate with as little as $\mathcal{N}=50$ training data points, where only minor discrepancies can be observed in comparison to the moments computed using the true model. 
For $\mathcal{N}=150$ training data points, surrogate-based estimation of the mean and the standard deviation is almost indistinguishable to the ones computed with the original model.

\begin{comment}
% UQ plots
\begin{figure}
\centering
\begin{subfigure}[b]{0.49\textwidth}
 \centering
 \includegraphics[width=\textwidth]{PGA-PCE/Figures/Results/Lotka-Volterra/LV_UQ_p_2_ntrain_10.png}
 \caption{$\mathcal{N}=10$, $p_{\max}=2$}
 \label{fig:uq_lv_n10}
\end{subfigure}
\hfill
\begin{subfigure}[b]{0.49\textwidth}
 \centering
 \includegraphics[width=\textwidth]{PGA-PCE/Figures/Results/Lotka-Volterra/LV_std_UQ_p_2_ntrain_10.png}
 \caption{$\mathcal{N}=10$, $p_{\max}=2$}
 \label{fig:uq_std_lv_n10}
\end{subfigure}
\\
\begin{subfigure}[b]{0.49\textwidth}
 \centering
 \includegraphics[width=\textwidth]{PGA-PCE/Figures/Results/Lotka-Volterra/LV_UQ_p_2_ntrain_20.png}
 \caption{$\mathcal{N}=20$, $p_{\max}=2$}
 \label{fig:uq_lv_n20}
\end{subfigure}
\hfill
\begin{subfigure}[b]{0.49\textwidth}
 \centering
 \includegraphics[width=\textwidth]{PGA-PCE/Figures/Results/Lotka-Volterra/LV_std_UQ_p_2_ntrain_20.png}
 \caption{$\mathcal{N}=20$, $p_{\max}=2$}
 \label{fig:uq_std_lv_n20}
\end{subfigure}
\caption{Lotka-Volterra: Mean field and standard deviation field for Monte Carlo simulation with $\mathcal{N}=10, 20$ training data and $p_{\max}=2$.}
\label{fig:uq_lv}
\end{figure}
\end{comment}

\subsection{Continuous stirred-tank reactor}
\label{subsec:CSTR}
The continuous stirred-tank reactor (CSTR) is a chemical reactor model that is commonly used in chemical and process engineering modeling and design applications \cite{seborg2016process}. 
We consider an ideal CSTR with first-order reaction kinetics and a single reagent. 
The corresponding concentration and energy balances with respect to the contents of the reactor are given by the system of ODEs
\begin{equation} 
\label{eq:cstr}
\begin{split}
    V\frac{dc}{dt} & = q(c_{f} - c) - V k(T) c, \\
    V \rho C_p \frac{dT}{dt} & = w C_p (T_f - T) + (-\Delta H_R) V k(T) c + UA(T_c - T),
\end{split}
\end{equation}
where $k(T) = k_0 \exp\left(-E_a/RT\right)$ is the rate constant according to Arrhenius law. 
All other model parameters are listed in Table~\ref{table:cstr}.
As uncertain parameter, we consider the cooling temperature $T_c$, which is chosen to follow a uniform distribution in the range of $\left[305, 310\right]$~K.
The system is solved with an explicit Runge-Kutta method of fourth order, with time step equal to $0.01$ minutes and for a total duration of $5$ minutes. 
The initial conditions are $c_0=0.5$~gmol/lt and $T_0=350$~K.
The QoIs are the concentration $c$ and the temperature $T$ over time, both of which are extremely sensitive to changes in the cooling temperature.
With the given time discretization, the full response of the CSTR model $\mathscr{Y}_i=[\mathbf{c}_i, \mathbf{T}_i]$ is a point lying in $\mathbb{R}^{2\times 500}$.

\begin{table}[b!]
\small
\caption{Parameters of the CSTR model.}
\vspace{-7pt}
\centering
\begin{tabular}{l c c l}
\toprule
Parameters & \hspace{15pt} & Notation & \hspace{2pt} Uncertainty/value \\ [0.5ex]
\toprule
Reagent concentration (gmol/lt)  &  & $c$ &  $c_0 = 0.5$ \\
Feed concentration (gmol/lt) &  &  $c_f$ & $1.0$\\
Temperature (K)  &  & $T$ &  $T_0 = 350$ \\
Feed temperature (K) &  &  $T_f$ & $350$ \\
Cooling temperature (K)  &  & $T_c$ &  $\it{U}(305, 310)$ \\
Activation energy (J/gmol) & & $E_a$ & $72750$ \\
Pre-exponential factor (1/min) & & $k_0$ & $7.2 \cdot 10^7$ \\ 
Gas constant (J/kmol/K) & & $R$ & $8.314$ \\
Reactor volume (lt) & & $V$ & $100$ \\
Density (g/lt) & & $\rho$ &  $1000$\\
Heat capacity (J/g/K) & & $C_p$ & $0.239$ \\
Reaction enthalpy (J/gmol) & & $\Delta H_R$ & $-5 \cdot 10^4$ \\
Heat transfer coefficient (J/min/K) &  &  $UA$ & $5 \cdot 10^4$\\
Feed flowrate (lt/min) &  &  $q$ & $100$ \\
\bottomrule
\end{tabular}
\label{table:cstr}
\end{table}

Surrogate models with respect to the matrices $\textbf{U}$, $\textbf{V}$, and $\boldsymbol{\Sigma}$, as well as for the QoI $\mathscr{Y}$ are constructed using training data sets of size $\mathcal{N} \in \left\{20, 50, 100\right\}$ and $p_{\max}=2$.
Otherwise, the surrogate modeling procedure is similar to the one described in Section~\ref{subsec:LV}.
In this case, the objective is to learn the mapping  $T_c \rightarrow \mathscr{Y}$ using a small number of training data $\{T_{c,i}, \mathscr{Y}_i\}$. 
Despite considering only a single random parameter, the surrogate modeling task remains challenging due to the sensitivity of both temperature and reagent concentration to the cooling temperature.

\begin{figure}[!ht]
\centering
\includegraphics[width=0.6\textwidth]{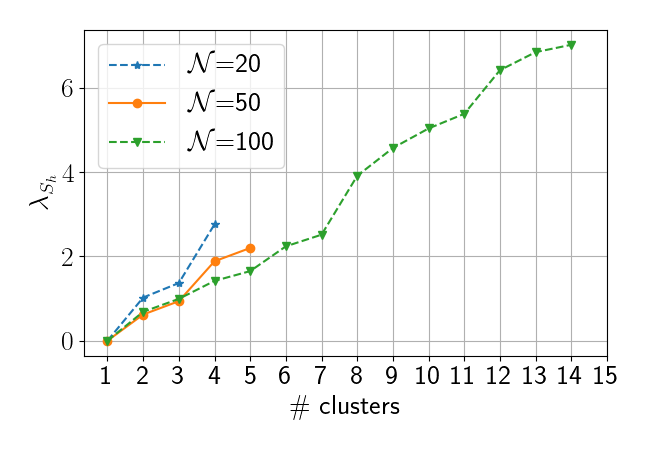}
\caption{Convergence of the Silhouette coefficient for an increasing number of clusters and for training data sets of increasing size $\mathcal{N}$. The optimum number of clusters is the one corresponding to the maximum peak of the each plot since the minimum number of points ($N_h=5$) allowed within a cluster was reached.}
\label{fig:frechet_cstr}
\end{figure}

% Frechet variances figure
%\begin{figure}
%\centering
%\begin{subfigure}[b]{0.49\textwidth}
% \centering
% \includegraphics[width=\textwidth]{Figures/Results/CSTR/CSTR_FrechetVarianceConvergence_Ntrain_50.png}
% \caption{$\mathcal{N}=50$.}
%\end{subfigure}
%\hfill
%\begin{subfigure}[b]{0.49\textwidth}
% \centering
% \includegraphics[width=\textwidth]{Figures/Results/CSTR/CSTR_FrechetVarianceConvergence_Ntrain_100.png}
% \caption{$\mathcal{N}=100$.}
% \label{fig:frechet_lv_n100}
%\end{subfigure}
%\caption{CSTR: Convergence of the minimum mean Fr\'{e}chet variance (solid blue line) and the corresponding 96\% confidence interval (light blue area) for an increasing number of clusters and for training data sets of increasing size $\mathcal{N}$.}
%\label{fig:frechet_cstr}
%\end{figure}

Figure~\ref{fig:frechet_cstr} shows the convergence of the Silhouette coefficient for an increasing number of clusters and  for training data sets of size $\mathcal{N} = 50$ and $\mathcal{N} = 100$. 
For $\mathcal{N} = 100$, 17 clusters are required.

\begin{figure}[t!]
\centering
\begin{subfigure}[b]{0.49\textwidth}
 \centering
 \includegraphics[width=\textwidth]{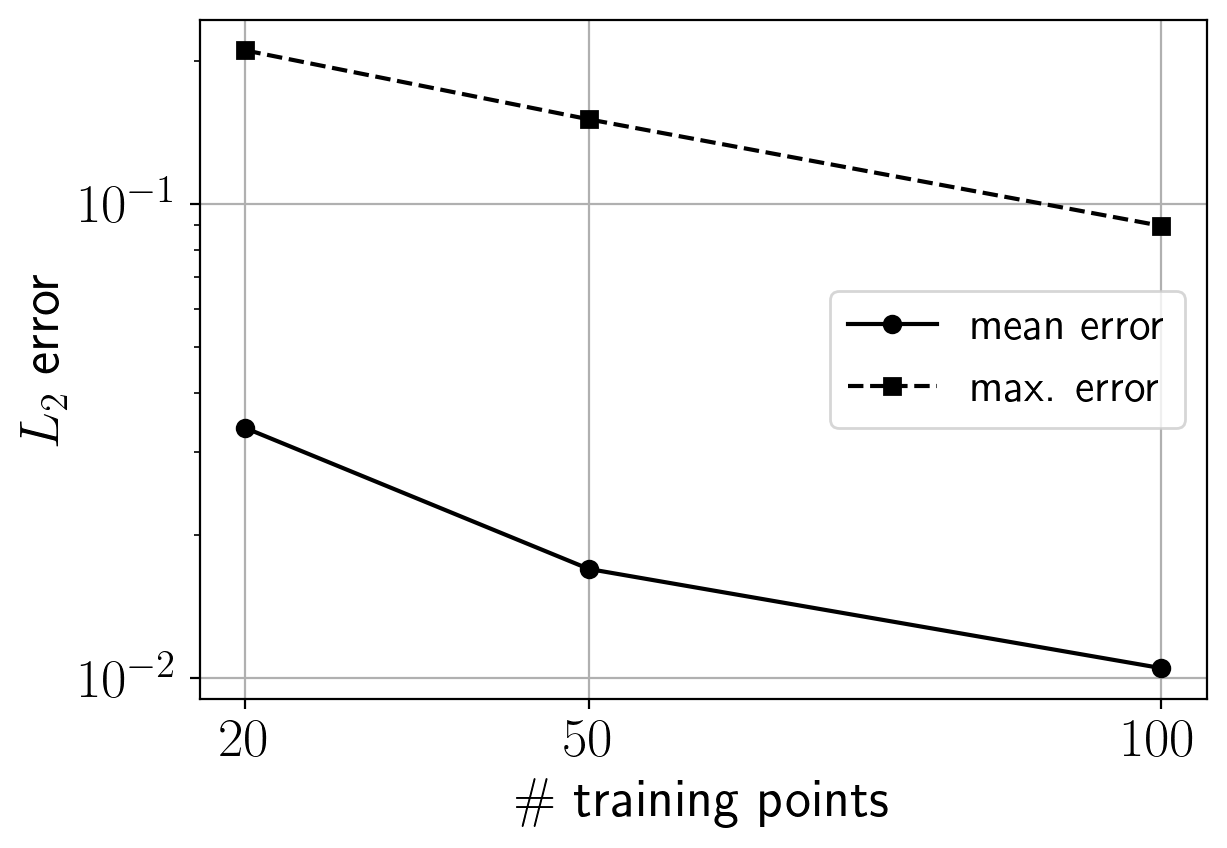}
 \caption{$U$ prediction.}
\end{subfigure}
\hfill
\begin{subfigure}[b]{0.49\textwidth}
 \centering
 \includegraphics[width=\textwidth]{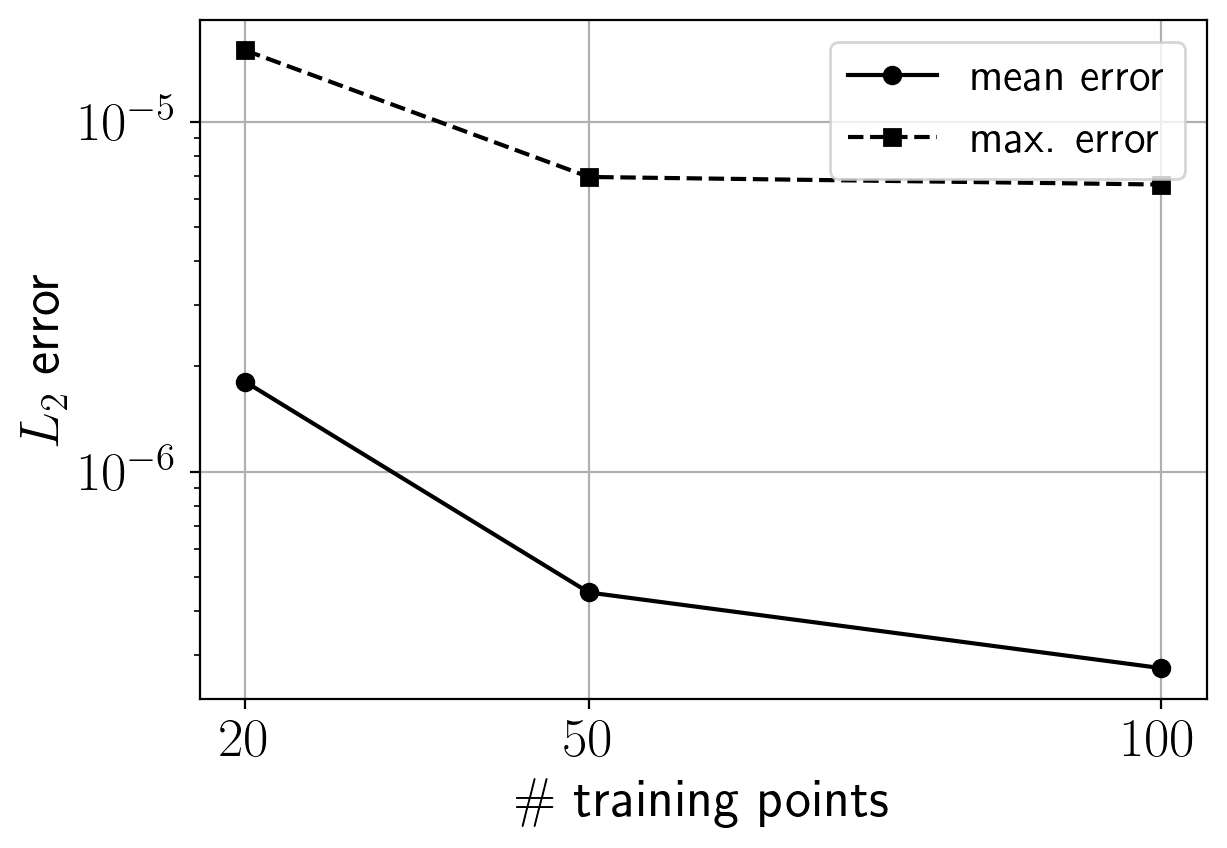}
 \caption{$V$ prediction.}
\end{subfigure}
\\
\begin{subfigure}[b]{0.49\textwidth}
 \centering
 \includegraphics[width=\textwidth]{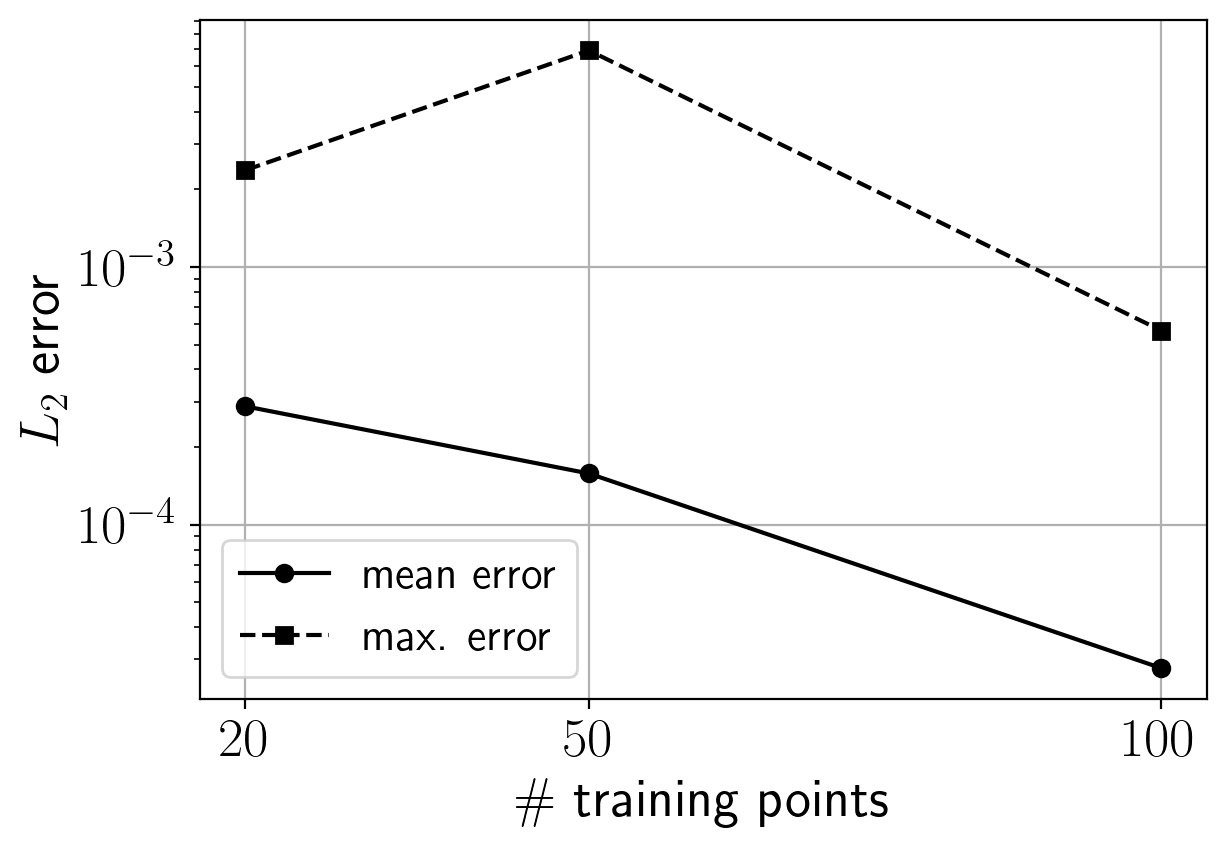}
 \caption{$\Sigma$ prediction.}
\end{subfigure}
\hfill
\begin{subfigure}[b]{0.49\textwidth}
 \centering
 \includegraphics[width=\textwidth]{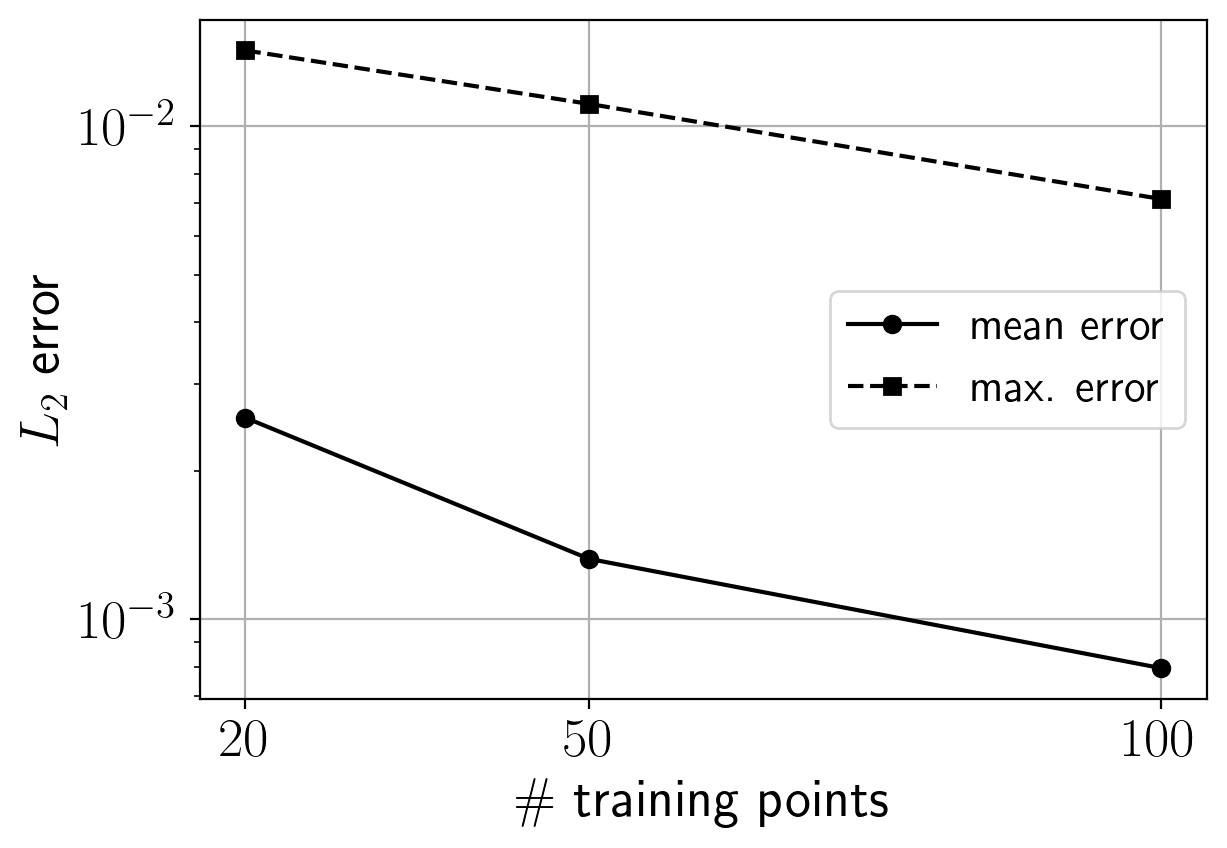}
 \caption{$\mathscr{Y}$ prediction.}
\end{subfigure}
\caption{CSTR: $L_2$ errors of the surrogate model's predictions with respect to the matrices $U$, $V$, and $\Sigma$, as well as for the QoI $\mathscr{Y}$, for training data sets of increasing size. The $L_2$ errors have been computed using a validation data set with $\mathcal{N}_*=5000$ data points. The PCE surrogates are constructed with $p_{\max}=2$.}
\label{fig:l2_cstr}
\end{figure}

% worst trajectories
\begin{figure}[t!]
\centering
\begin{subfigure}[b]{0.49\textwidth}
 \centering
 \includegraphics[width=\textwidth]{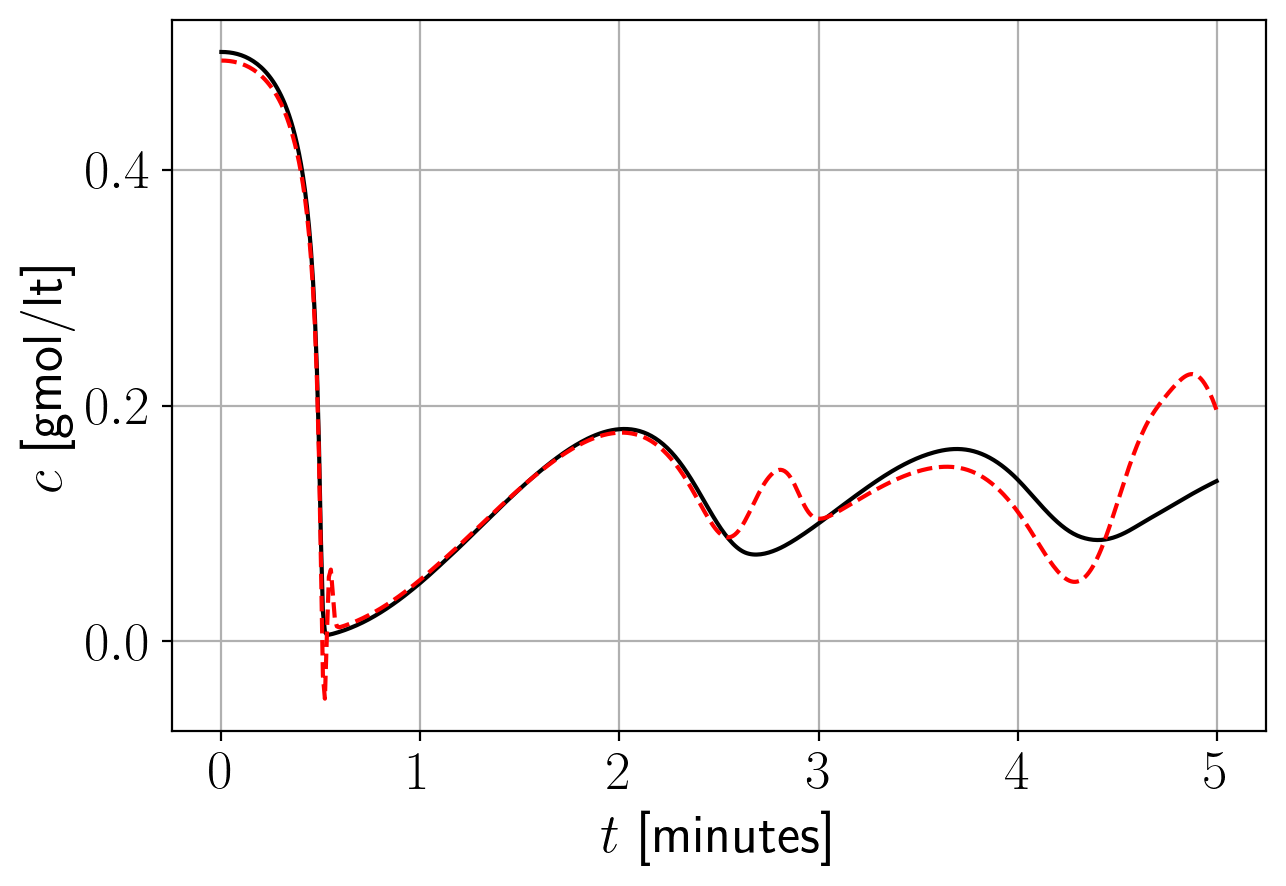}
 \caption{Worst-case concentration, $\mathcal{N}=20$.}
\end{subfigure}
\hfill
\begin{subfigure}[b]{0.49\textwidth}
 \centering
 \includegraphics[width=\textwidth]{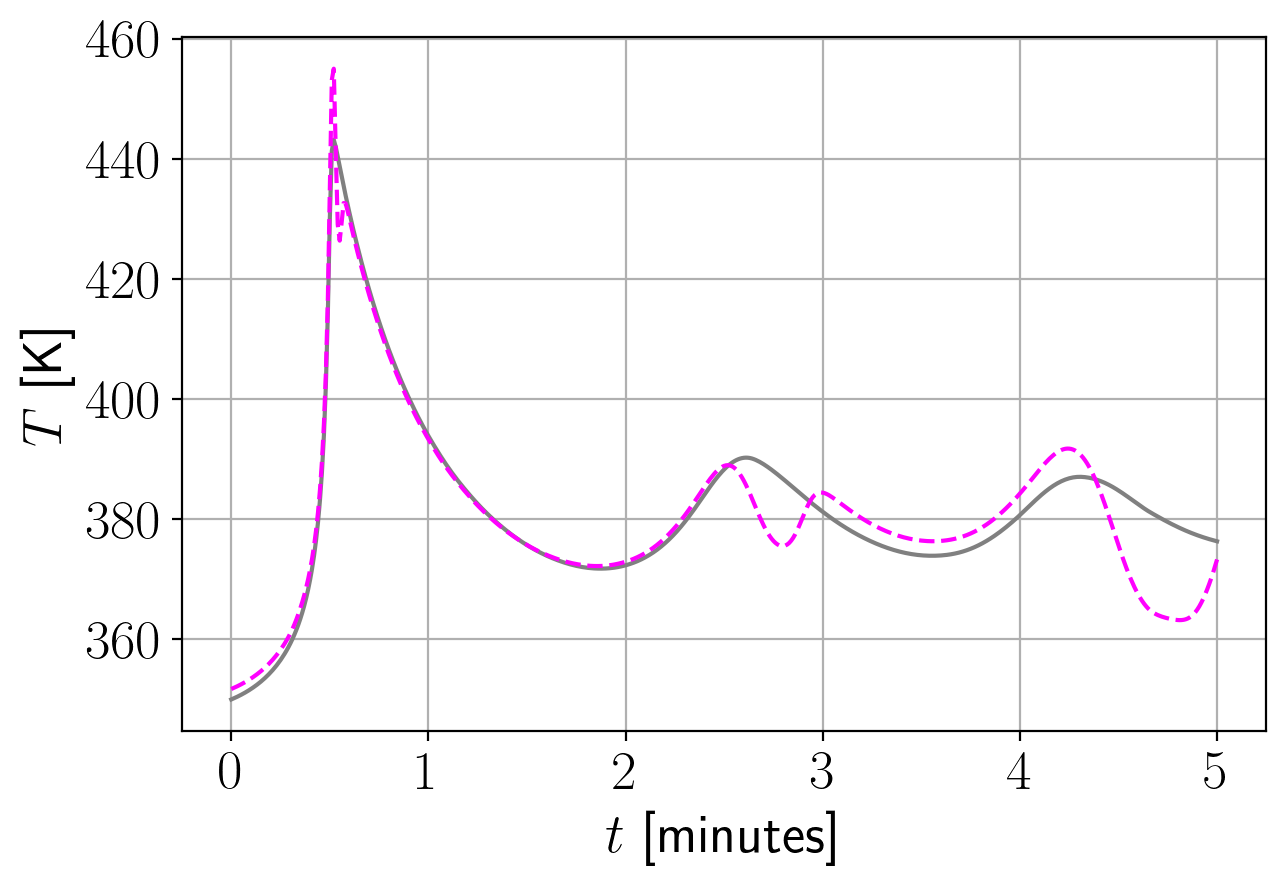}
 \caption{Worst-case temperature, $\mathcal{N}=20$.}
\end{subfigure}
\\
\begin{subfigure}[b]{0.49\textwidth}
 \centering
 \includegraphics[width=\textwidth]{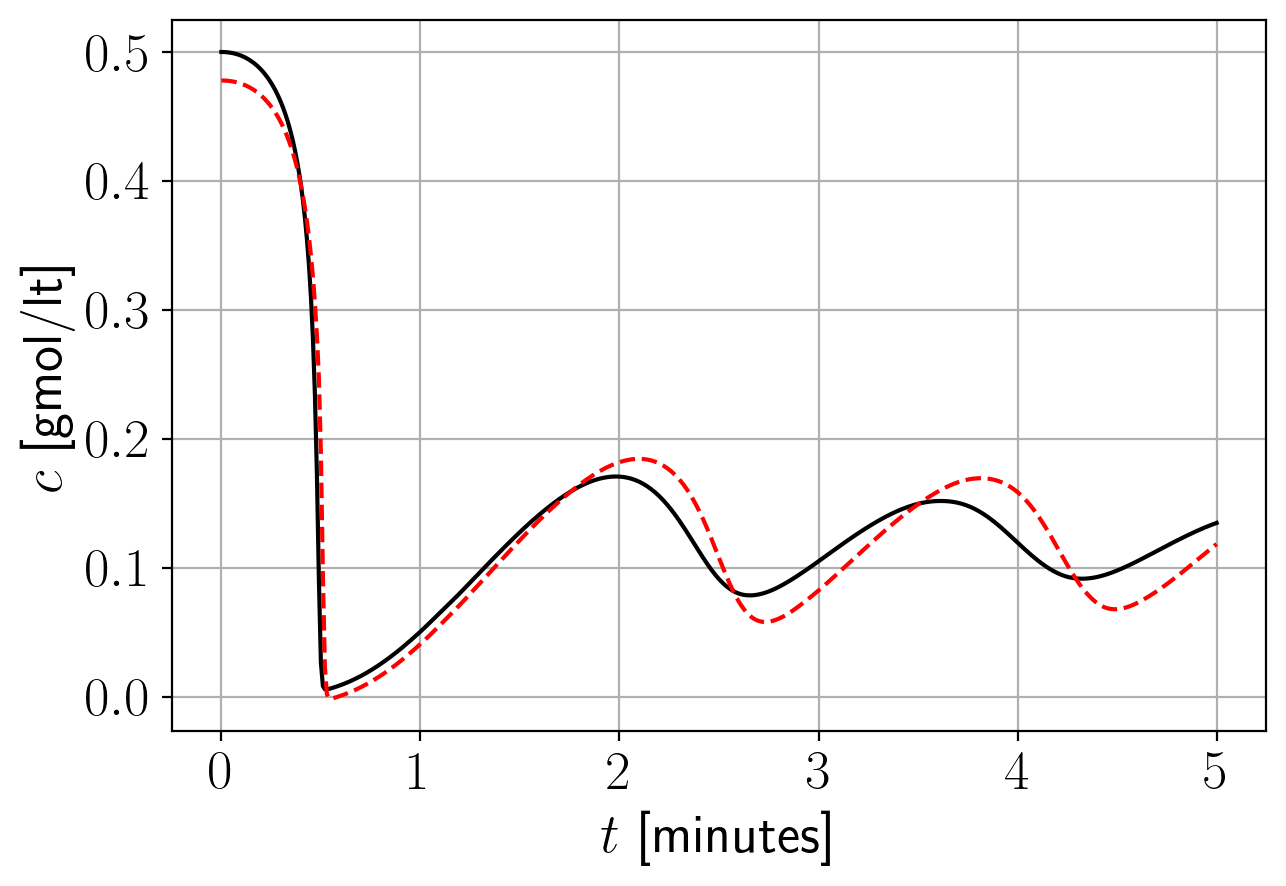}
 \caption{Worst-case concentration, $\mathcal{N}=50$.}
\end{subfigure}
\hfill
\begin{subfigure}[b]{0.49\textwidth}
 \centering
 \includegraphics[width=\textwidth]{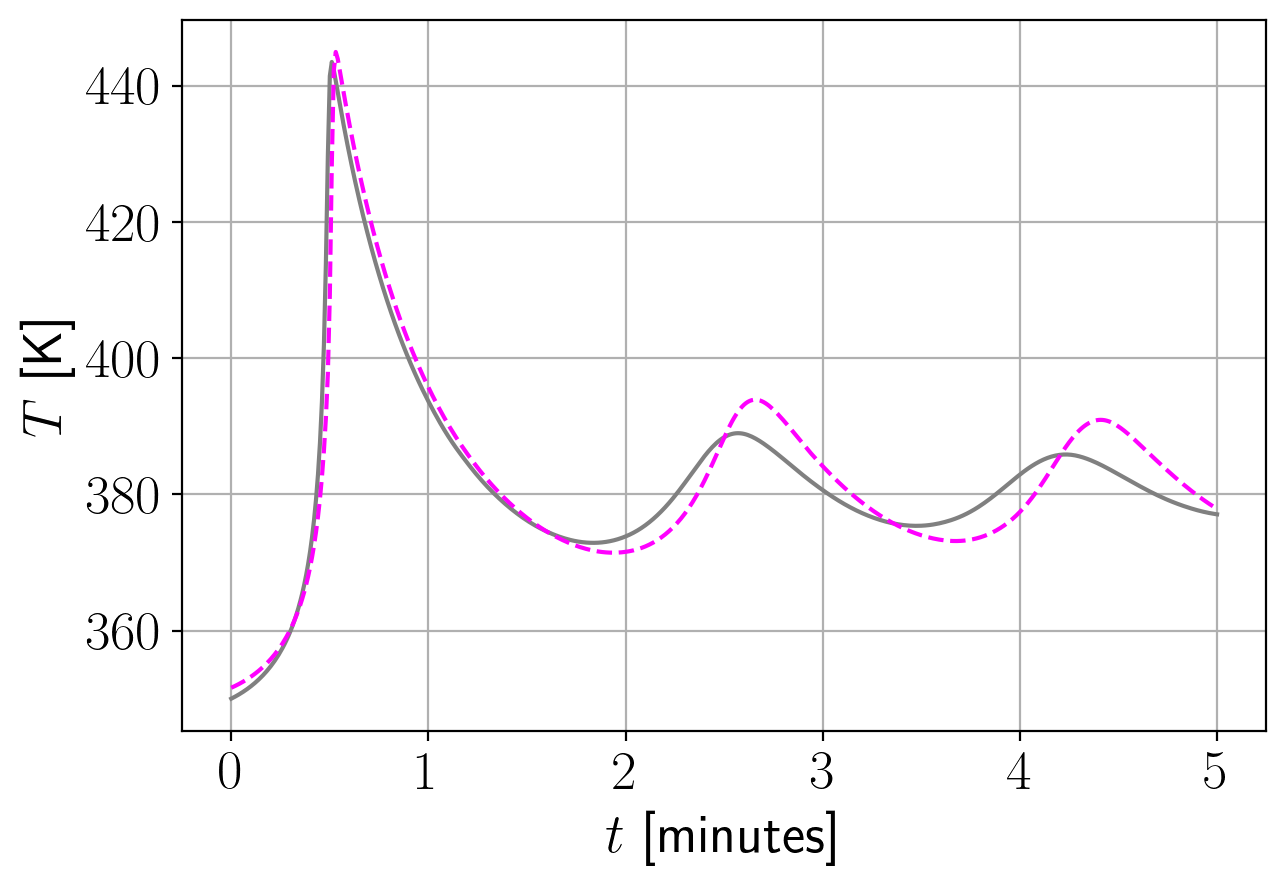}
 \caption{Worst-case temperature, $\mathcal{N}=50$.}
\end{subfigure}
\\
\begin{subfigure}[b]{0.49\textwidth}
 \centering
 \includegraphics[width=\textwidth]{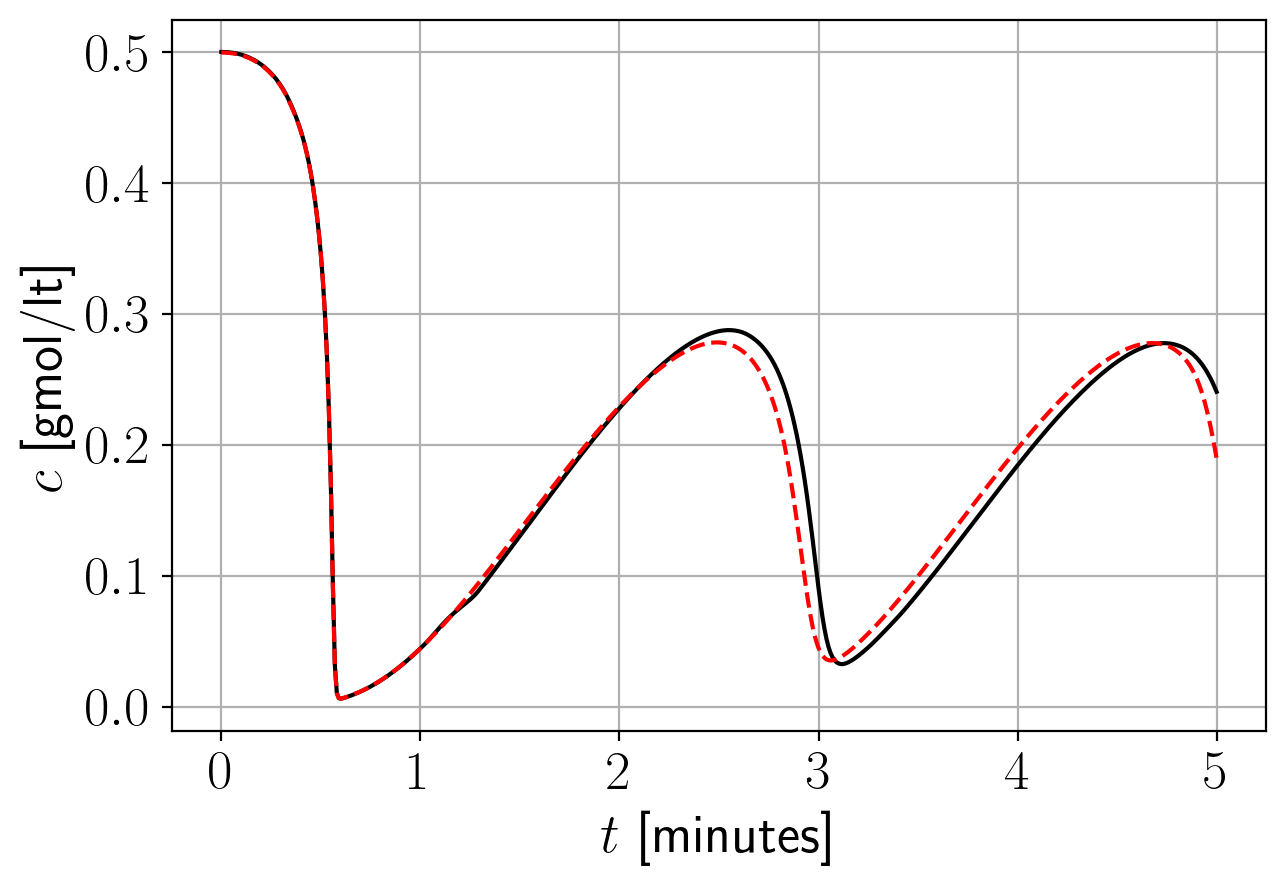}
 \caption{Worst-case concentration, $\mathcal{N}=100$.}
\end{subfigure}
\hfill
\begin{subfigure}[b]{0.49\textwidth}
 \centering
 \includegraphics[width=\textwidth]{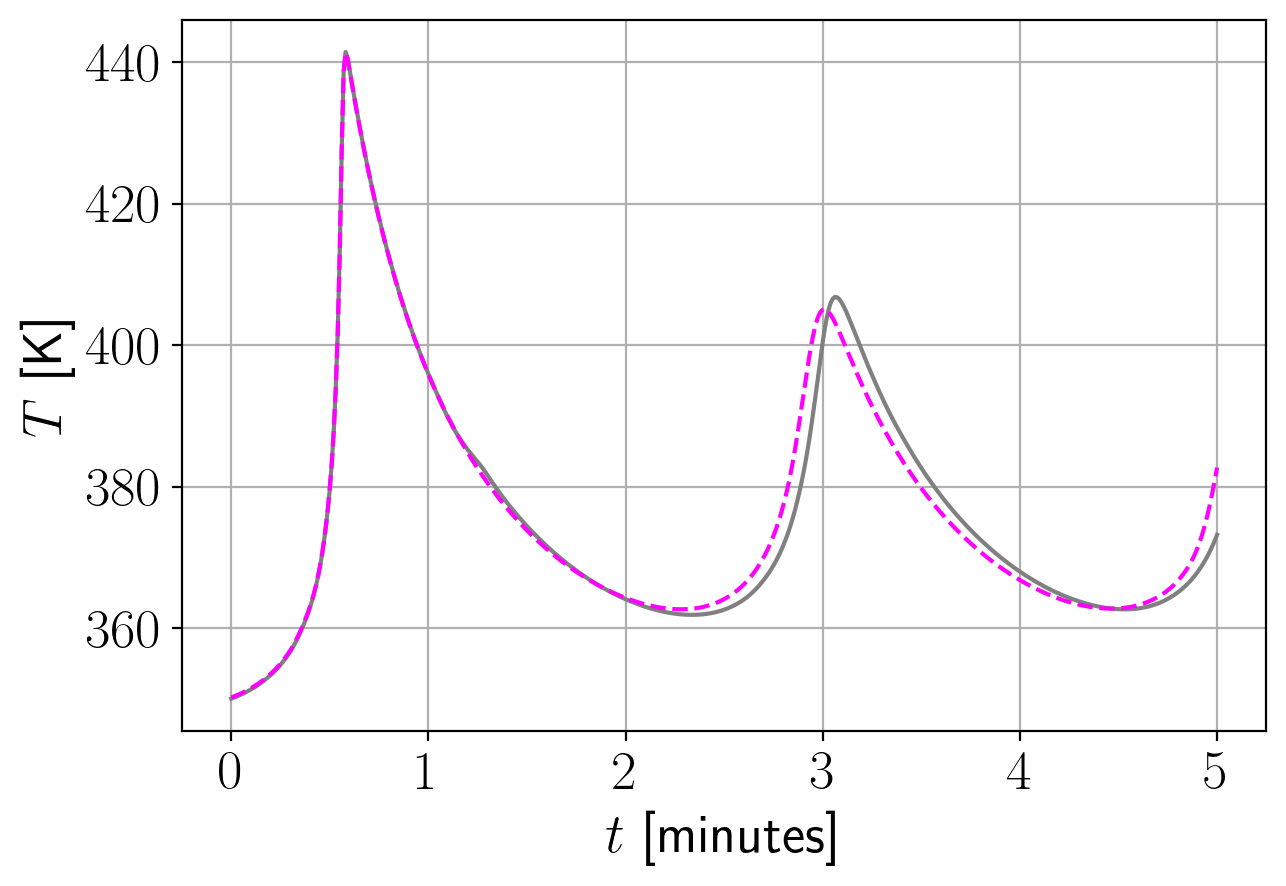}
 \caption{Worst-case temperature, $\mathcal{N}=100$.}
\end{subfigure}
\caption{CSTR: Worst-case surrogate model predictions for the concentration and temperature over time. The true trajectories are given with solid lines ($c$: black, $T$: gray), while the surrogate's predictions are given with dashed lines ($c$: red, $T$: magenta).}
\label{fig:worst_case_trajectories_cstr}
\end{figure}

Figure~\ref{fig:l2_cstr} shows the surrogate model's $L_2$ error with respect to $\textbf{U}$, $\textbf{V}$, $\boldsymbol{\Sigma}$, and $\mathscr{Y}$, computed using a test data set with $\mathcal{N}_*=5000$ data points.
Interestingly, while the $L_2$ error regarding the QoI $\mathscr{Y}$ decreases monotonically for increasing training data, in two cases, namely, for the matrices $\textbf{U}$ and $\boldsymbol{\Sigma}$, the maximum error stagnates or even deteriorates, e.g., for the surrogate's predictions of $\boldsymbol{\Sigma}$ with $\mathcal{N}=50$ training data points.
Figure~\ref{fig:worst_case_trajectories_cstr} shows the worst-case surrogate model predictions with respect to the concentration and the temperature, out of the aforementioned $\mathcal{N}_*=5000$ test model evaluations. 
As can be observed, with $\mathcal{N}=50$ training data points, the surrogate's prediction captures the correct physical behavior, however, without sufficient accuracy. 
For $\mathcal{N}=100$ training data points, the accuracy of the surrogate increases significantly and only minor differences to the true model can be observed, even in the worst case.

% UQ
\begin{figure}[t!]
\centering
\begin{subfigure}[b]{0.49\textwidth}
 \centering
 \includegraphics[width=\textwidth]{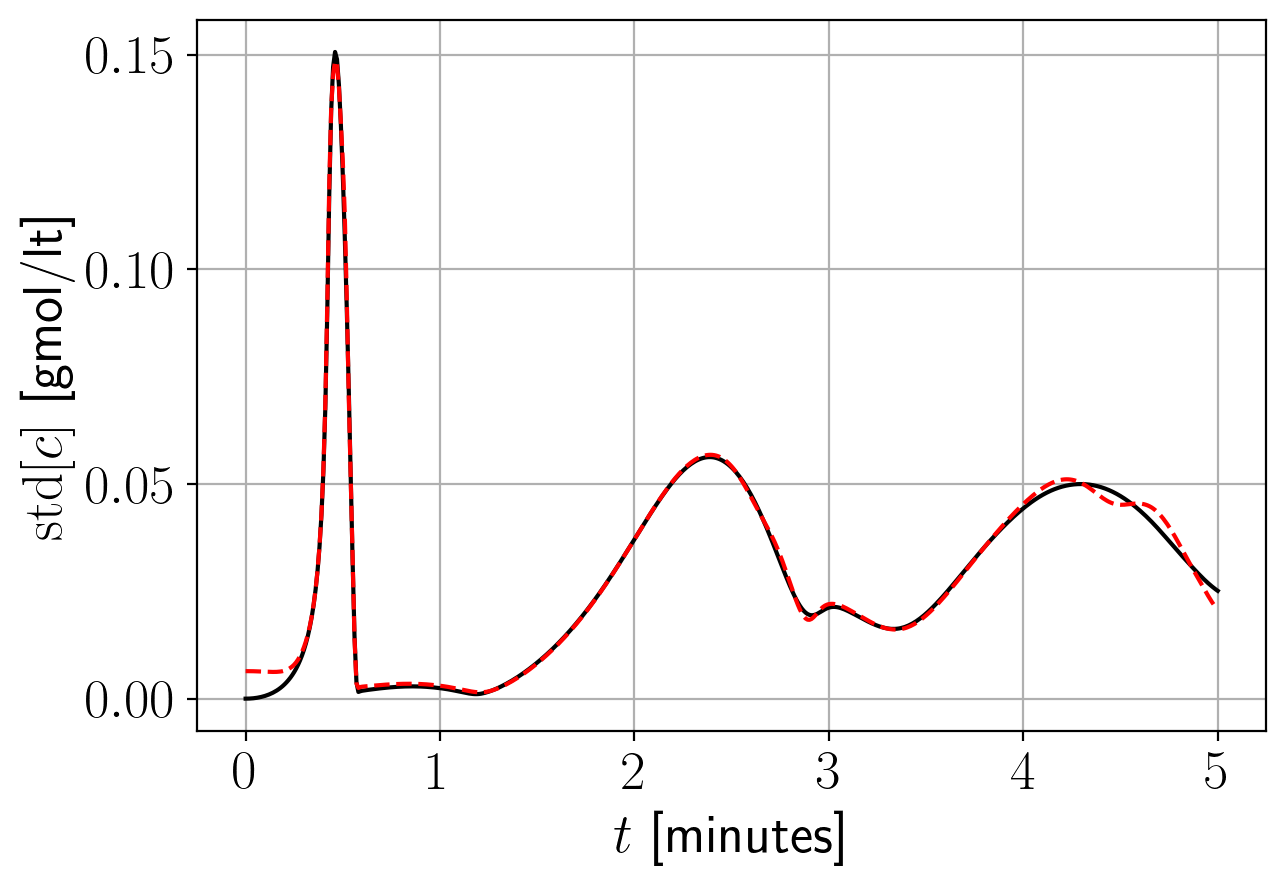}
 \caption{Standard deviation of concentration, $\mathcal{N}=20$.}
 \label{fig:std_c_n20_p2}
\end{subfigure}
\hfill
\begin{subfigure}[b]{0.49\textwidth}
 \centering
 \includegraphics[width=\textwidth]{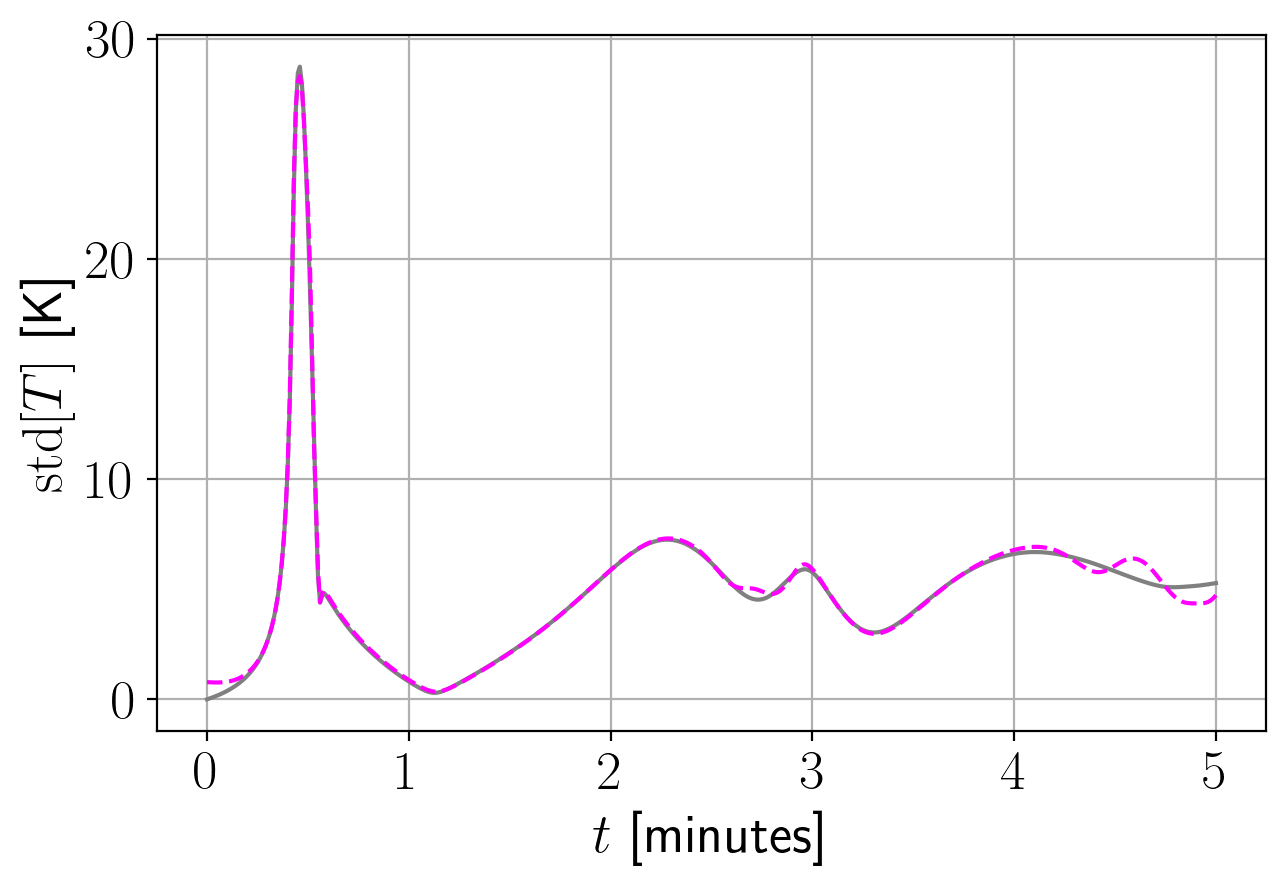}
 \caption{Standard deviation of temperature, $\mathcal{N}=20$.}
 \label{fig:std_T_n20_p2}
\end{subfigure}
\\
\begin{subfigure}[b]{0.49\textwidth}
\centering
\includegraphics[width=\textwidth]{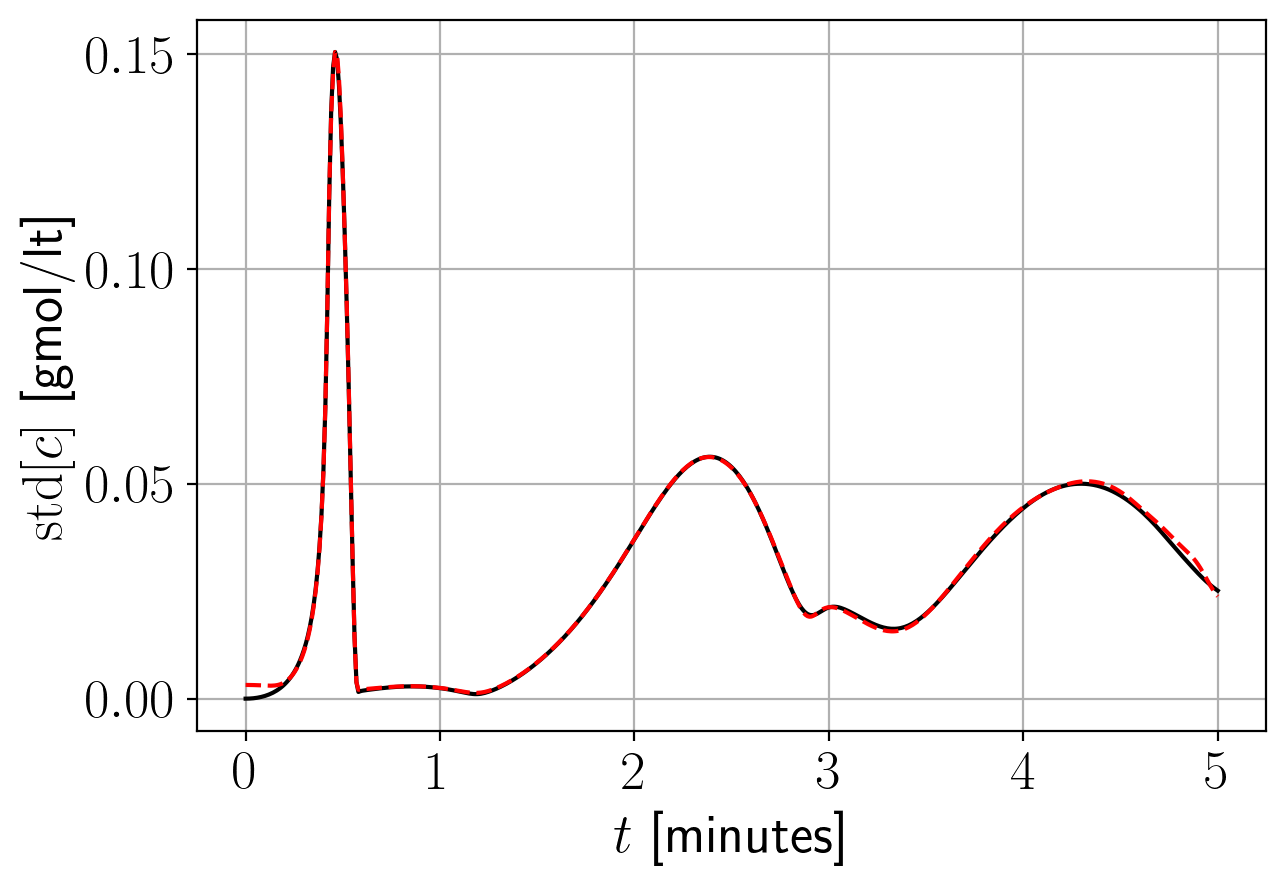}
\caption{Standard deviation of concentration, $\mathcal{N}=50$.}
\label{fig:std_c_n50_p2}
\end{subfigure}
\hfill
\begin{subfigure}[b]{0.49\textwidth}
\centering
\includegraphics[width=\textwidth]{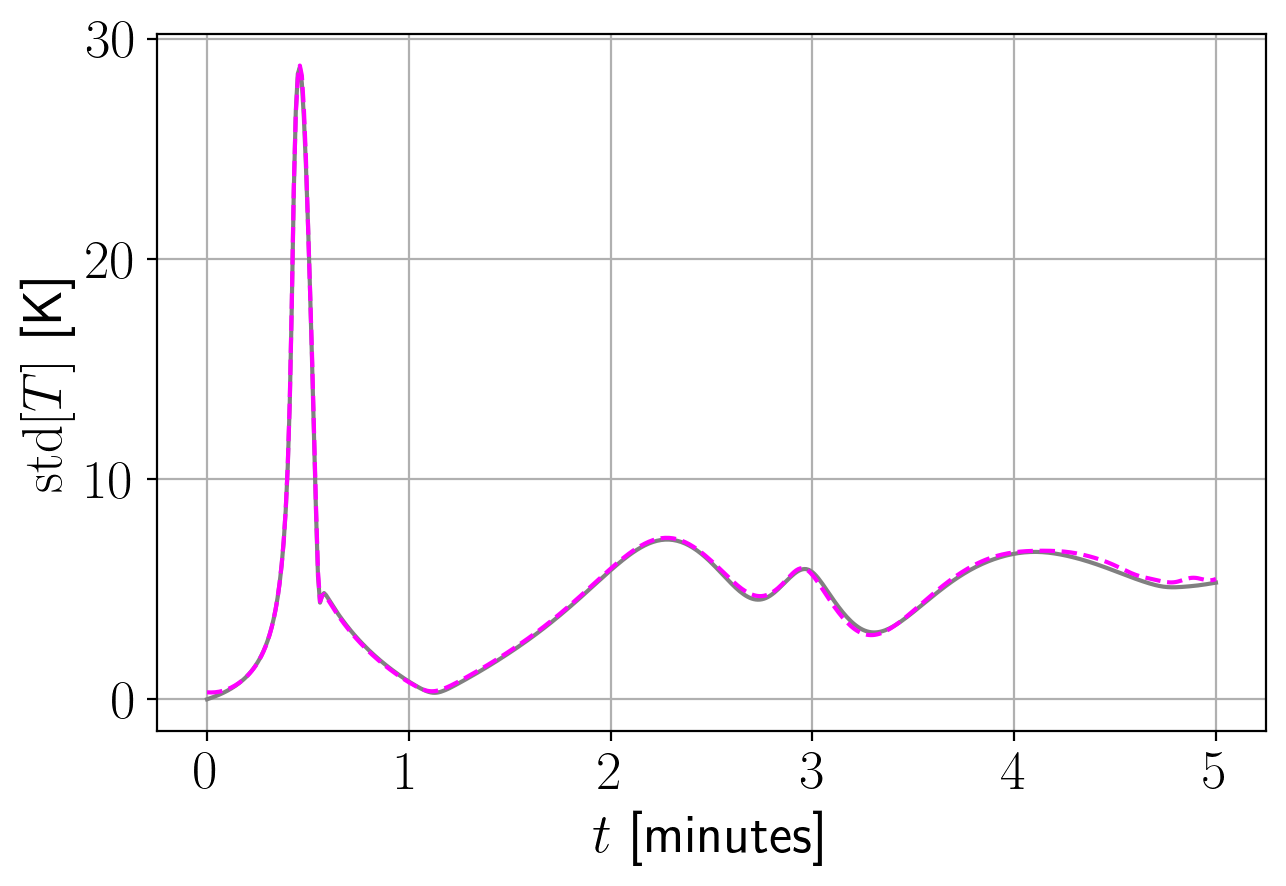}
\caption{Standard deviation of temperature, $\mathcal{N}=50$.}
\label{fig:std_T_n50_p2}
\end{subfigure}
\\
\begin{subfigure}[b]{0.49\textwidth}
\centering
\includegraphics[width=\textwidth]{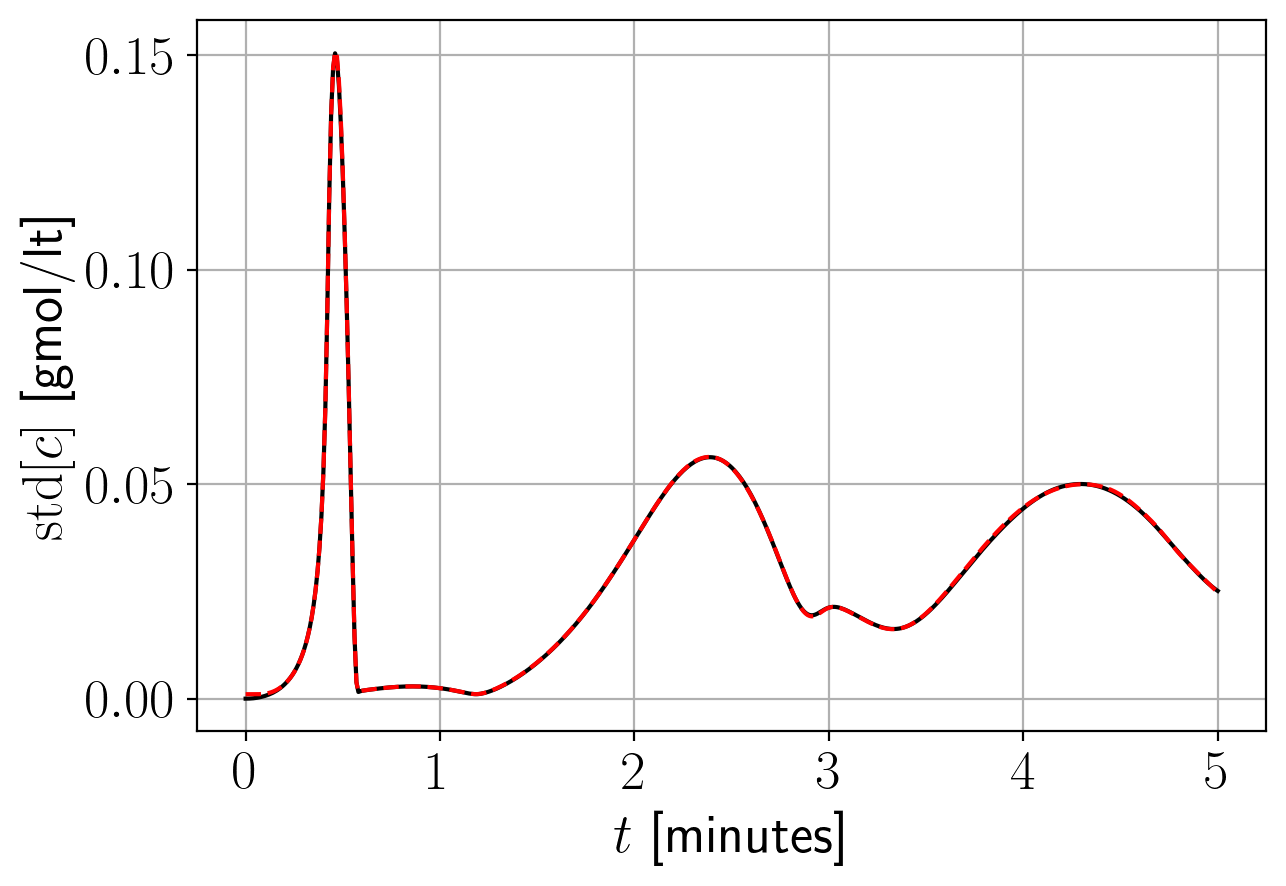}
\caption{Standard deviation of concentration, $\mathcal{N}=100$.}
\label{fig:std_c_n100_p2}
\end{subfigure}
\hfill
\begin{subfigure}[b]{0.49\textwidth}
\centering
\includegraphics[width=\textwidth]{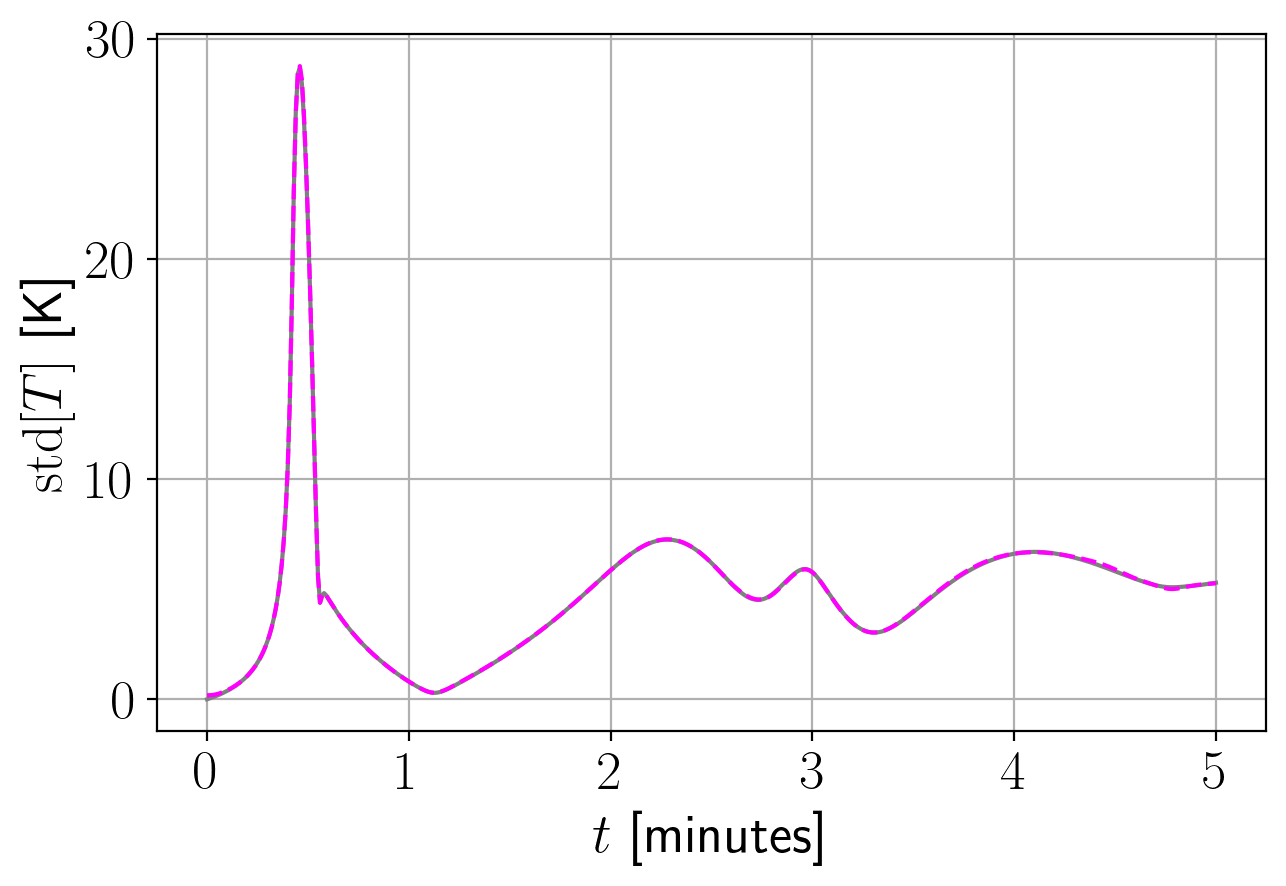}
\caption{Standard deviation of temperature, $\mathcal{N}=100$.}
\label{fig:std_T_n100_p2}
\end{subfigure}
\caption{CSTR: Standard deviation estimation for the concentration and the temperature over time. The standard deviations computed with the true model are given with solid lines ($c$: black, $T$: gray), while the surrogate-based estimations are given with dashed lines ($c$: red, $T$: magenta).}
\label{fig:uq_cstr}
\end{figure}

Last, Figure~\ref{fig:uq_cstr} shows the surrogate-based estimations for standard deviation of $c$ and $T$ over time, in comparison to the estimations based on the true CSTR model. 
Similar to \ref{subsec:LV}, the estimations for the standard deviations is based on Monte Carlo sampling, using the $\mathcal{N}_*=5000$ original and surrogate model evaluations that were previously employed to compute the $L_2$ error metrics and the worst-case predictions of the surrogate.
The estimations for the mean trajectories of $c$ and $T$ have been omitted, due to the fact that the surrogate's estimations match those of the true model even for $\mathcal{N}=20$ training data points.
Contrarily, in the case of the standard deviation, differences between the true and the surrogate model can be observed for  $\mathcal{N}=20$ and  $\mathcal{N}=50$ training data points, in particular at the start and end times of the simulation. 
These differences become negligible if $\mathcal{N}=100$ data points are used to train the surrogate model.

\subsection{Rayleigh-B\'enard convection}

Rayleigh-B\'enard convection is a natural phenomenon that occurs in a thin layer of fluid when it is heated from below \cite{chilla2012new}. It is characterized by the formation of fluid motion patterns driven by buoyancy forces. The convection arises due to a temperature gradient ($\Delta T$) across the fluid layer, with warmer fluid rising and cooler fluid sinking, creating a circulation pattern. The instability in Rayleigh-B\'enard convection is determined by the non-dimensional Rayleigh number 
\begin{equation}
    Ra = \frac{\alpha \Delta Tgh^3}{\nu \kappa}
\end{equation}

where $\alpha$ is the thermal expansion coefficient, $g$ is the gravitational acceleration, $h$ is the thickness of the fluid layer, $\nu$ is the kinematic viscosity and $\kappa$ is the thermal diffusivity. %The latter two parameters are related to the Prandtl ($Pr$)  and Rayleigh numbers as
%\begin{eqnarray}
%    \kappa= (Ra \cdot Pr)^{-1/2} \\ \nonumber
%    \nu = (\frac{Ra}{Pr})^{-1/2} \nonumber
%\end{eqnarray}
This number represents the ratio between the buoyancy forces and the viscous forces within the fluid. When the temperature gradient $\Delta T$ exceeds a certain threshold, the Rayleigh number surpasses a critical value, leading to the onset of convection and the formation of characteristic patterns, such as thermal plumes or rolls.

The dimensional form of the Rayleigh-B\'enard equations for an incompressible fluid defined on a domain $\Omega$ can be derived by considering the governing conservation laws for mass, momentum, and energy, along with the Boussinesq approximation, which assumes that density perturbations solely impact the gravitational force.
 
\[ 
\left\{
\begin{array}{ll}
      \frac{d\textbf{u}}{dt}=-\frac{1}{\rho_0}\nabla p + \frac{\rho}{\rho_0}g + \nu \nabla^2 \textbf{u} & \textbf{x} \in \Omega, t>0 \\
      \frac{dT}{dt}=  \kappa \nabla^2 T & \textbf{x} \in \Omega, t>0 \\
      \nabla \textbf{u} = 0   \\
      \rho = \rho_0(1-\alpha(T-T_0))
\end{array} 
\right. 
\]
where $ \frac{d}{dt}$ denotes material derivative, $\textbf{u}$, $p, T$ are the fluid velocity, pressure and temperature respectively, $T_0$ is the temperature at the lower plate, and $\textbf{x} = (x, y)$ are the spatial coordinates. The corresponding boundary and initial conditions are defined, respectively, as

\[ 
\left\{
\begin{array}{ll}
      T(\textbf{x}, t)|_{y=0}=T_0 & \textbf{x} \in \Omega, t>0 \\
      T(\textbf{x}, h)|_{y=0}=T_1 & \textbf{x} \in \Omega, t>0 \\
      \textbf{u}(\textbf{x}, h)|_{y=0}=\textbf{u}(\textbf{x}, h)|_{y=h}=0 & \textbf{x} \in \Omega, t>0  \\
      T(y, t)|_{t=0}=T_0+\frac{y}{h}(T_1-T_0) + 0.1 \nu(\textbf{x})  & \textbf{x} \in \Omega \\
      \textbf{u}(\textbf{x}, t)|_{t=0}=0 & \textbf{x} \in \Omega
\end{array} 
\right. 
\]
where $T_0$, and $T_1$ are the fixed temperatures of the lower and upper plates, respectively. The simulation takes place in a domain $\Omega = [0, 4] \times [0, 1]$, discretized with $n_x \times n_y = 128 \times 64$ mesh points. We consider the dimensionless Rayleigh number to be uniformly distributed in the range $[1\cdot 10^6, 5\cdot 10^6]$, while the  the Prandtl number is equal to 1.  For each realization, the partial differential equations is solved in the time interval $t = [0, 20]$ for $\delta t = 0.25$. We want to train our model to predict the response (buoyancy forces) at time $t = 18$. We generated 200 data from which $\mathcal{N} = 50$ where used for training and $\mathcal{N}^* = 150$ where used for testing.  Datasets were generated using the Dedalus Project that can be found in \url{https://github.com/ DedalusProject/dedalus}.

%\begin{figure}[!ht]
%\centering
%\includegraphics[width=0.5\textwidth]{PGA-PCE/Figures/Results/Rayleigh-%Bernard/RBe_meanSillouette_Ntrain.png}
%\caption{Convergence of the Silhouette coefficient for an increasing number of %clusters. The optimum number of clusters is 6.}
%\label{fig:frechet_RB2D}
%\end{figure}

\begin{figure}[H]
\centering
\begin{subfigure}[b]{0.78\textwidth}
 \centering
 \includegraphics[clip=true, trim=0cm 5cm 0cm 6cm, width=\textwidth]{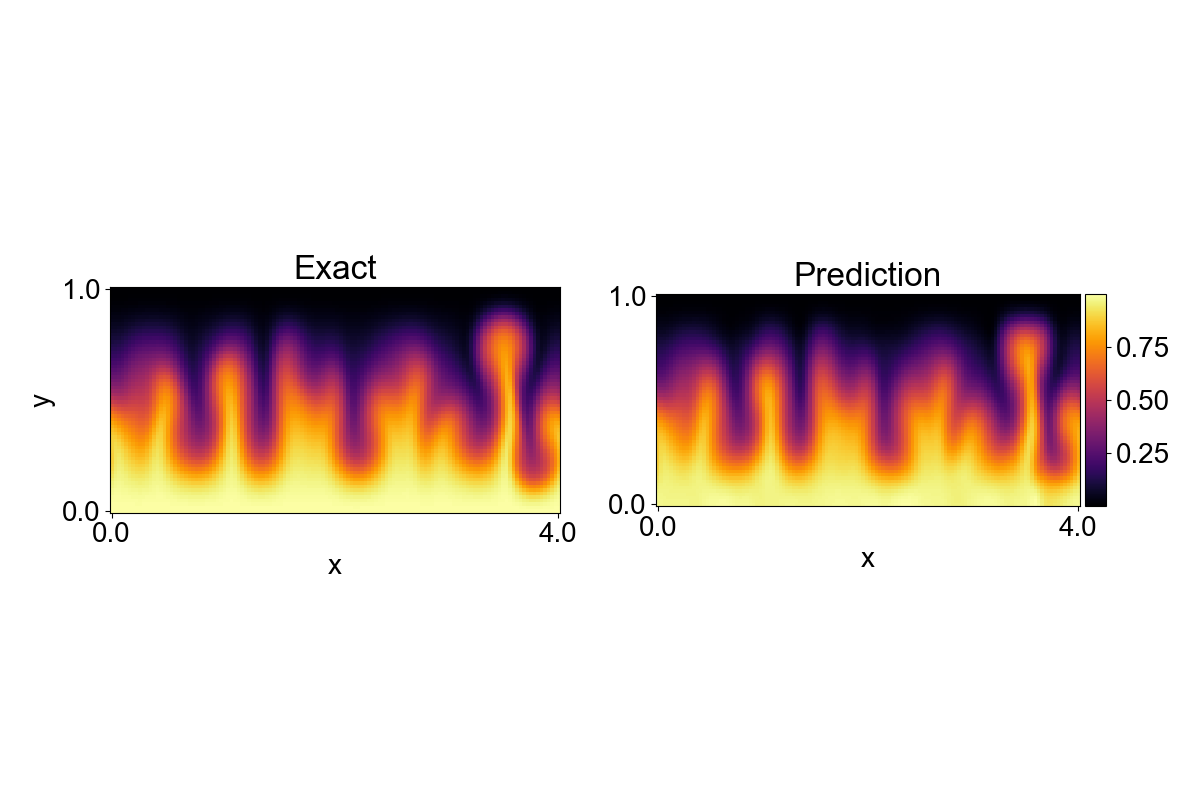}
 \caption{Best prediction}
\end{subfigure}
\hfill
\begin{subfigure}[b]{0.83\textwidth}
 \centering
 \includegraphics[clip=true, trim=0cm 5cm 0cm 6cm, width=\textwidth]{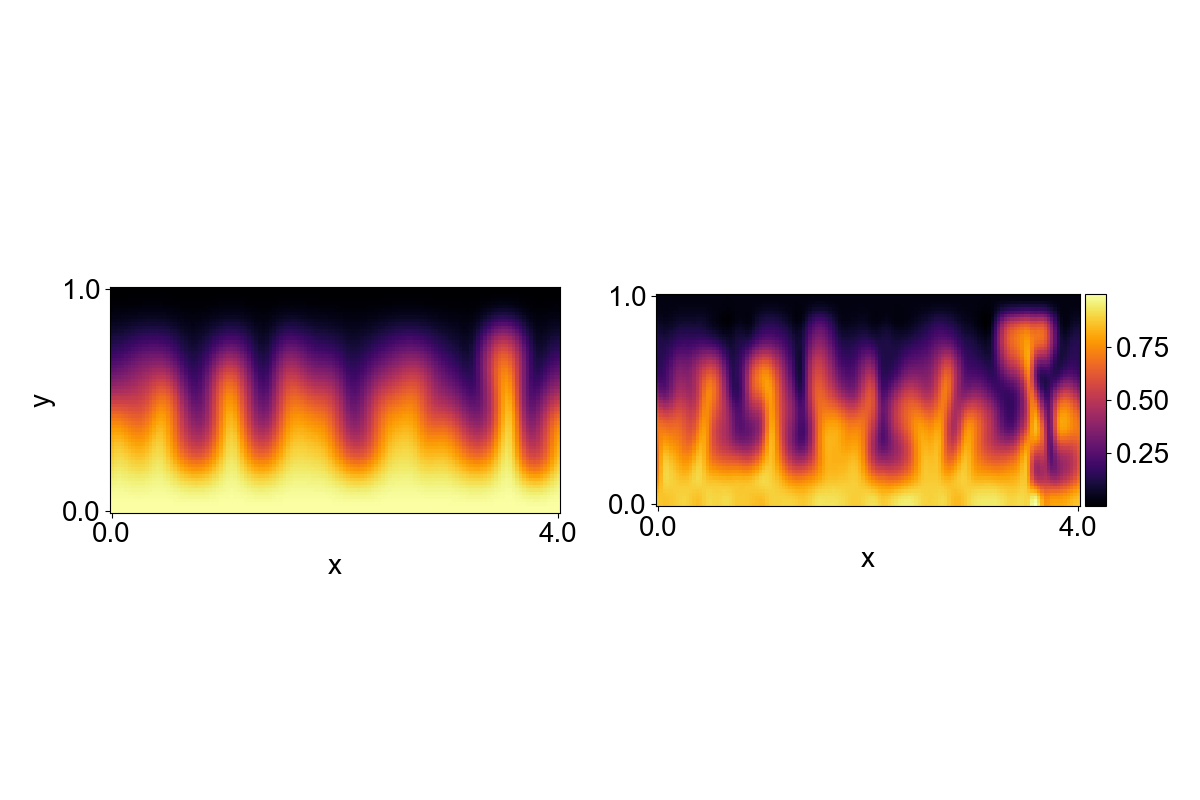}
 \caption{Worst prediction}
\end{subfigure}
\caption{Rayleigh-B\'enard: Best-case (first row) and worst-case (second row) surrogate model predictions for the (normalized, non-dimensional) buoyancy vs the reference solution. }
\label{fig:RB2D_best_worst}
\end{figure}

%Figure~\ref{fig:frechet_cstr} shows the convergence of the Silhouette coefficient for an increasing number of clusters from which we observe that  6 clusters are required. 
We ran our algorithm and the optimal number of clusters was found to be 7. Figure \ref{fig:RB2D_best_worst} depicts  the best- and worst-case surrogate model predictions with respect to the buoyancy, out of the 150 test model evaluations. As can be observed, even with $\mathcal{N} = 50$ training
data points, the surrogate’s prediction captures the correct physical behavior, with only minor differences to the true model can be observed, even in the worst-case.

% Frechet variances figure
\begin{figure}[H]
\centering
\begin{subfigure}[b]{0.8\textwidth}
 \centering
 \includegraphics[clip=true, trim=0cm 5cm 0cm 6cm, width=\textwidth]{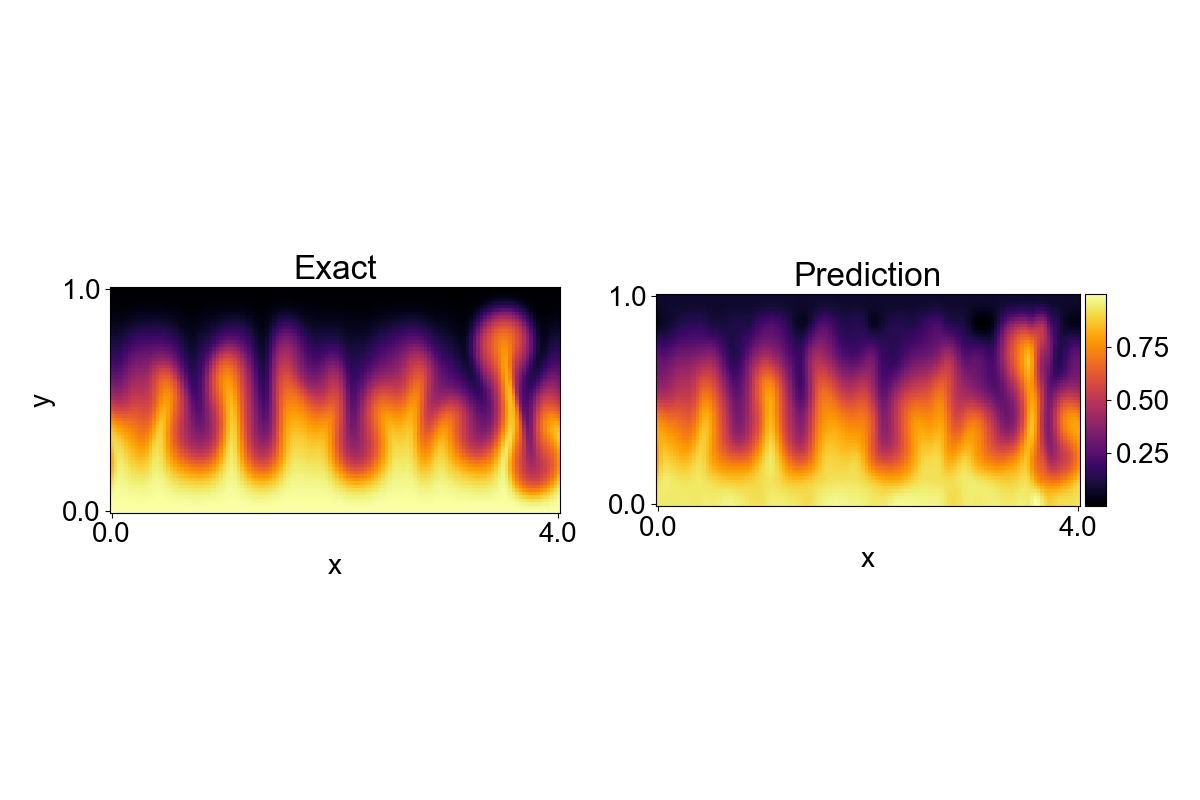}
 \caption{Mean}
\end{subfigure}
\hfill
\begin{subfigure}[b]{0.8\textwidth}
 \centering
 \includegraphics[clip=true, trim=0cm 5cm 0cm 6cm, width=\textwidth]{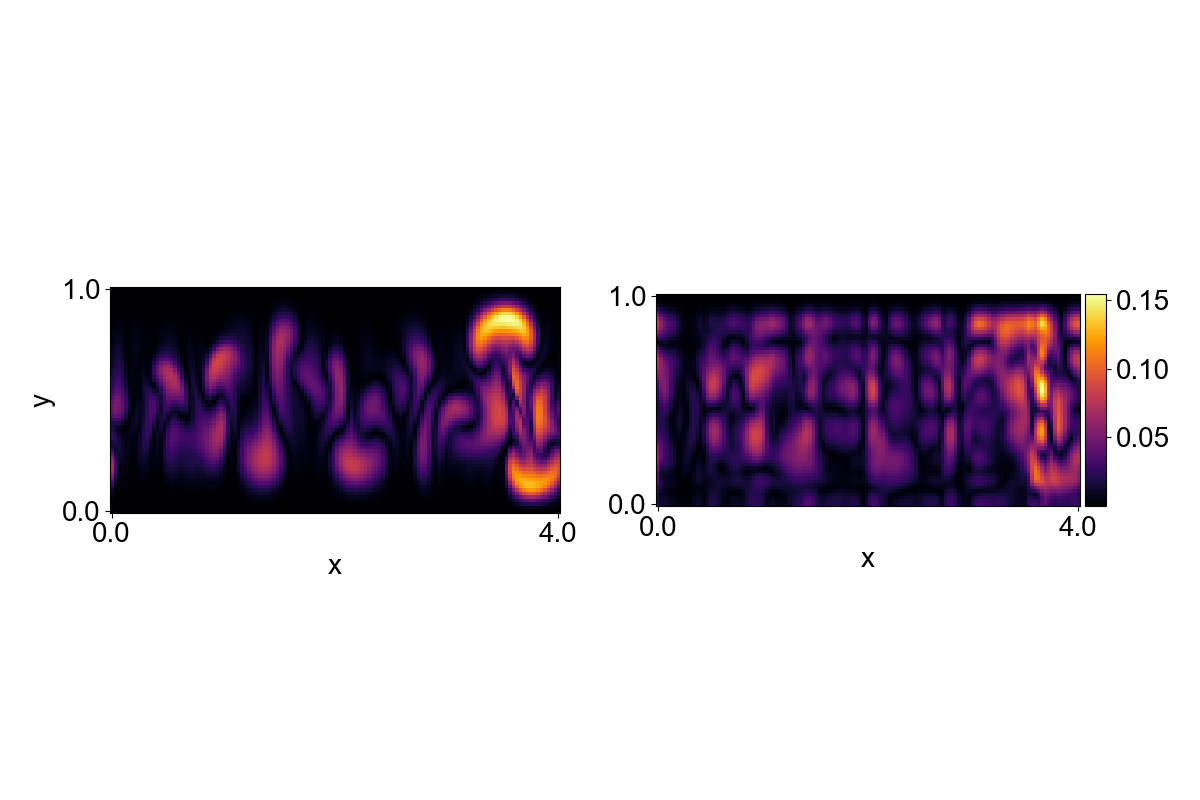}
 \caption{Standard deviation}
\end{subfigure}
\caption{Rayleigh-B\'enard: Relative errors for the mean field and standard deviation surrogate predictions. }
\label{fig:RB2D_mean_field}
\end{figure}

\begin{figure}[H]
\centering
 \includegraphics[clip=true, trim=0cm 5cm 0cm 6cm, width=\textwidth]{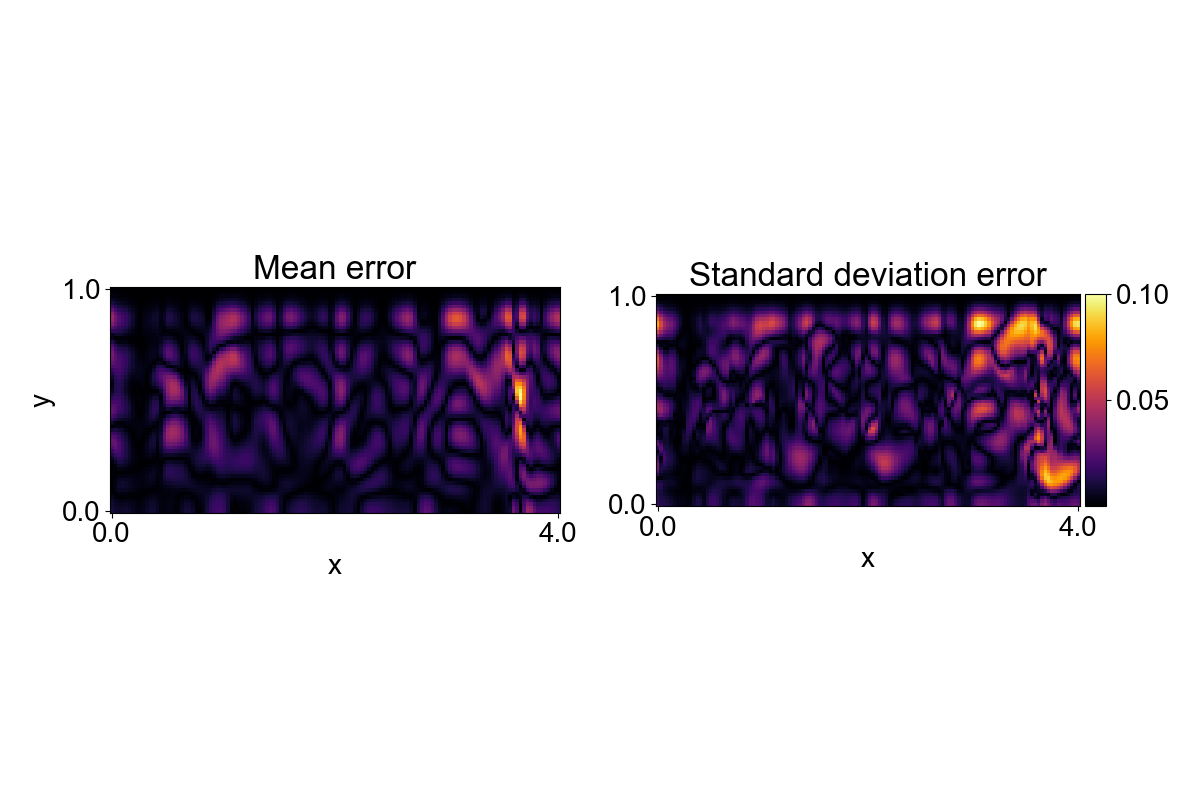}
\caption{Rayleigh-B\'enard: Mean field  and standard deviation  errors. }
\label{fig:RB2D_errors}
\end{figure}

Figure \ref{fig:RB2D_mean_field} shows the surrogate's estimations for the mean buoyoncy field and the standard deviation (based on the $\mathcal{N}=150$ test points), in comparison to the estimations based on the true  model. As we can observe from this figure, the surrogate's
estimations match those of the true model  with sufficient accuracy. Similarly, the differences between the true model and the surrogate model in the case of the standard deviation are negligible. These differences are quantified in terms of the $L_2$ error, depicted in Fig. \ref{fig:RB2D_errors}.

\section{Discussion and Conclusions}
\label{discussion}

\noindent
The paper presents a new methodology for creating surrogate models in high-dimensional systems for uncertainty quantification. The method  combines Principal Geodesic Analysis on the Grassmann manifold of
the response with Polynomial Chaos expansion to construct a mapping between the random input parameters
and the projection of the response on the local principal geodesic submanifolds.  The resulting surrogate model uses an adaptive algorithm based on unsupervised learning and
the minimization of the sample Frèchet variance on the Grassmann manifold. The method is tested on four applications from different fields and shows promising results for accurately predicting new solutions. Additionally, the method is computationally efficient and reduces the cost associated with high-fidelity simulations. However, the method's computational cost becomes intractable for cases with high-dimensional input parameter spaces, and the use of standard random or quasi-random sampling techniques might not be ideal for all cases. Future research should focus on addressing these limitations and challenges.

%\section{Acknowledgements}

\biboptions{sort&compress}
\bibliography{manuscript}

\appendix

\section{Projection onto the tangent space}
\label{appendix_1}

\noindent
In the neighborhood of $\textbf{X}_1 \in \mathcal{G}_{p, n}$, a point $\textbf{X}_2$ can be mapped onto the tangent plane $\mathcal{T}_{\textbf{X}_1}$ using the logarithmic mapping $\log_{\textbf{X}_1}(\textbf{X}_2)=\boldsymbol{\Gamma}$. The inverse mapping onto the manifold can be performed using the exponential mapping $\exp_{\textbf{X}_1}(\boldsymbol{\Gamma})=\textbf{X}_2$ which is derived by taking the thin SVD of $\boldsymbol{\Gamma}_1$, $\boldsymbol{\Gamma}_1 = \textbf{U}\boldsymbol{\Sigma}\textbf{V}^\intercal$, i.e.

\begin{equation}\label{eq:4}
\textbf{X}_2 = \exp_{\textbf{X}_1}(\textbf{U}\boldsymbol{\Sigma}\textbf{V}^\intercal)=\textbf{X}_1\textbf{V}\cos(\boldsymbol{\Sigma})\textbf{Q}^\intercal + \textbf{U}\sin(\boldsymbol{\Sigma})\textbf{Q}^\intercal.
\end{equation}
where $\textbf{U}$ and $\textbf{V}$ are orthonormal matrices, $\boldsymbol{\Sigma}$ is diagonal matrix with positive real entries, and $\textbf{Q}$ is an orthogonal $n\times n$ matrix. Requiring that $\boldsymbol{\Gamma}$ lies on $\mathcal{T}_{\textbf{X}_1}$, defines the following set of equations
\begin{subequations}
    \begin{equation}\label{eq:6a}
    \textbf{V}\cos(\boldsymbol{\Sigma})\textbf{Q}^\intercal = \textbf{X}_1^\intercal \textbf{X}_2.
    \end{equation}
    \begin{equation}\label{eq:6b}
    \textbf{U}\sin(\boldsymbol{\Sigma})\textbf{Q}^\intercal =\textbf{X}_2- \textbf{X}_1\textbf{X}_1^\intercal\textbf{X}_2. 
    \end{equation}
\end{subequations}
Multiplying \eqref{eq:6a} by the inverse of \eqref{eq:6b} yields
\begin{equation}\label{eq:7}
    \textbf{U}\tan(\boldsymbol{\Sigma})\textbf{V}^\intercal = (\textbf{X}_2- \textbf{X}_1\textbf{X}_1^\intercal\textbf{X}_2) (\textbf{X}_1^\intercal \textbf{X}_2)^{-1}.
\end{equation}
The exponential mapping can be performed by taking the thin SVD of the matrix  $\textbf{M} = (\textbf{X}_2- \textbf{X}_1\textbf{X}_1^\intercal\textbf{X}_2) (\textbf{X}_1^\intercal \textbf{X}_2)^{-1}=\textbf{U}\boldsymbol{\Sigma}\textbf{V}^\intercal$. The logarithmic map can be defined by inversion of Eq.\ \eqref{eq:7} as \cite{Begelfor2006}:
\begin{equation}\label{eq:8}
    \log_{\textbf{X}_1}(\textbf{X}_2)=\boldsymbol{\Gamma} =\textbf{U}\tan^{-1}(\boldsymbol{\Sigma})\textbf{V}^\intercal.
\end{equation}

\section{Steps of the proposed methodology}
\label{appendix_2}

\noindent
A summary of the steps is given below. 

\begin{algorithm}[H]
\caption{}\label{alg:cap2}
\begin{algorithmic}
\Require A dataset $\{\boldsymbol{\theta}_i, \textbf{y}_i\}_{i=1}^{\mathcal{N}}$
%\Ensure $y = x^n$
\State Project $\textbf{y}_i$ on the Grassmann:  $\textbf{y}_i = \textbf{U}_i \boldsymbol{\Sigma}_i\textbf{V}_i^\intercal$

%\Comment{iterations number}
\State Perform Riemmannian K-means on $\{\textbf{U}_i\}_{i=1}^N$ on the Grassmann using Algorithm \ref{alg:cap}
\State For each  cluster $C_h$ perform Principal Component Analysis using Algorithm \ref{alg:cap1}:
\begin{itemize}
\item Find the Karcher means of points $\textbf{U}_j$ and  $\textbf{V}_j \in C_h$ using Eq.(\ref{eq:KarcherMean}) 
\item Project the points $\textbf{U}_j/\textbf{V}_j$ onto the tangent spaces with origin the corresponding Karcher means using the logarithmic mapping (see \ref{appendix_1})
\item On each tangent space perform Principal Component Analysis
\end{itemize}
\State For each  cluster $C_h$ train three PCE surrogates according to Eq.(\ref{eq:PCEB}).

\State For a new realization in the parameter space $\boldsymbol{\theta}^\star$:
\begin{itemize}
\item Find the nearest neighbour of $\boldsymbol{\theta}^\star$ in the parameter space using Eq.(\ref{eq.nn}) 
\item Use the corresponding trained PCE to predict:
\begin{itemize}
    \item the tangent vectors according to Eqs.(\ref{eq.pce_predict1}a) and (\ref{eq.pce_predict1}b)
    \item the singular values $\boldsymbol{\Sigma}_j$ according to Eq.(\ref{eq.predict_sigma}) 
    \end{itemize}
\item Map the predicted tangent vectors onto the Grassmannian with the exponential mapping (see \ref{appendix_1}) 
\item Invert the SVD to predict the full-field solution according to Eq.(\ref{eq.invert_svd})
\end{itemize}
\end{algorithmic}
\end{algorithm}

\end{document}